\documentclass[final]{cvpr}

\usepackage{times}
\usepackage{epsfig}
\usepackage{graphicx}
\usepackage{amsmath}
\usepackage{amssymb}

\usepackage[ruled]{algorithm2e}

\makeatletter
\newcommand{\thickhline}{%
    \noalign {\ifnum 0=`}\fi \hrule height 1pt
    \futurelet \reserved@a \@xhline
}
\makeatother

\usepackage[pagebackref=true,breaklinks=true,colorlinks,bookmarks=false]{hyperref}
\usepackage{tablefootnote}
\usepackage{array}

\def\cvprPaperID{5594} 
\def\confYear{CVPR 2021}

\begin{document}

\title{GMOT-40: A Benchmark for Generic Multiple Object Tracking}

\author{Hexin Bai$^1$\\
Temple University\\
Philadelphia, USA\\
{\tt\small hexin.bai@temple.edu}
\and
Wensheng Cheng$^1$\\
Stony Brook University\\
Stony Brook, USA\\
{\tt\small wenscheng@cs.stonybrook.edu}
\and
Peng Chu$^1$\\
Microsoft\\
Redmond, USA\\
{\tt\small pengchu@microsoft.com}
\and
Juehuan Liu\\
Temple University\\
Philadelphia, USA\\
{\tt\small juehuan.liu@temple.edu}
\and
Kai Zhang\\
Temple University\\
Philadelphia, USA\\
{\tt\small zhang.kai@temple.edu}
\and
Haibin Ling$^2$\\
Stony Brook University\\
Stony Brook, USA\\
{\tt\small hling@cs.stonybrook.edu}
}

\maketitle
\pagestyle{empty}  
\thispagestyle{empty} 


\footnotetext[1]{Equal contribution}
\footnotetext[2]{Corresponding author}
\begin{abstract}
Multiple Object Tracking (MOT) has witnessed remarkable advances in recent years. However, existing studies dominantly request prior knowledge of the tracking target (eg, pedestrians), and hence may not generalize well to unseen categories. In contrast, Generic Multiple Object Tracking (GMOT), which requires little prior information about the target, is largely under-explored. In this paper, we make contributions to boost the study of GMOT in three aspects. First, we construct the first publicly available dense GMOT dataset, dubbed GMOT-40, which contains 40 carefully annotated sequences evenly distributed among 10 object categories. In addition, two tracking protocols are adopted to evaluate different characteristics of tracking algorithms. Second, by noting the lack of devoted tracking algorithms, we have designed a series of baseline GMOT algorithms. Third, we perform a thorough evaluations on GMOT-40, involving popular MOT algorithms (with necessary modifications) and the proposed baselines. The GMOT-40 benchmark is publicly available at https://github.com/Spritea/GMOT40.

\end{abstract}

\section{Introduction}

Multiple Object Tracking (MOT) has long been studied in the computer vision community~\cite{ciaparrone2020deep,luo2014multiple}, due to its wide range of applications such as in robotics, surveillance, autonomous driving, cell tracking, \etc\ 
Remarkable advances have been made recently in MOT, partly due to the progress of major components such as detection, single object tracking, association, \etc  Another driving force comes from the popularization of MOT benchmarks (\eg, \cite{geiger2012we, MOTChallenge2015,MOT16, wen2020ua, zhu2020vision}). 
Despite the achievement, previous studies in MOT mostly focus on a specific object category of interest (pedestrian, car, cell, \etc) and rely on models of such objects. For example, detectors of such objects are often pre-trained offline, and motion patterns for specific objects are sometimes utilized as well. It remains unclear how well existing MOT algorithms generalize to unseen objects and hence constrains the expansion of MOT to new applications, especially those with limited data for training object detectors.  

By contrast, Generic Multiple Object Tracking (GMOT), which requests no prior knowledge of the objects to be tracked, aims to deal with these issues.
Hence GMOT could be applied in video editing, animal behaviour analysis, and vision based object counting. Despite its wide applications, it is however seriously under-explored, except for some early investigations~\cite{luo2013generic,luo2014bi}. Comparing the progress in GMOT with that in MOT, we see a clear lack of GMOT benchmark, and the absence of GMOT baselines with effective deep learning ingredients. Note that we follow the definition of GMOT in ~\cite{luo2014bi}, \textit{i.e.}, tracking multiple objects of a generic object class.


\begin{figure}[!t]
    \begin{center}
        \includegraphics[width=1\linewidth]{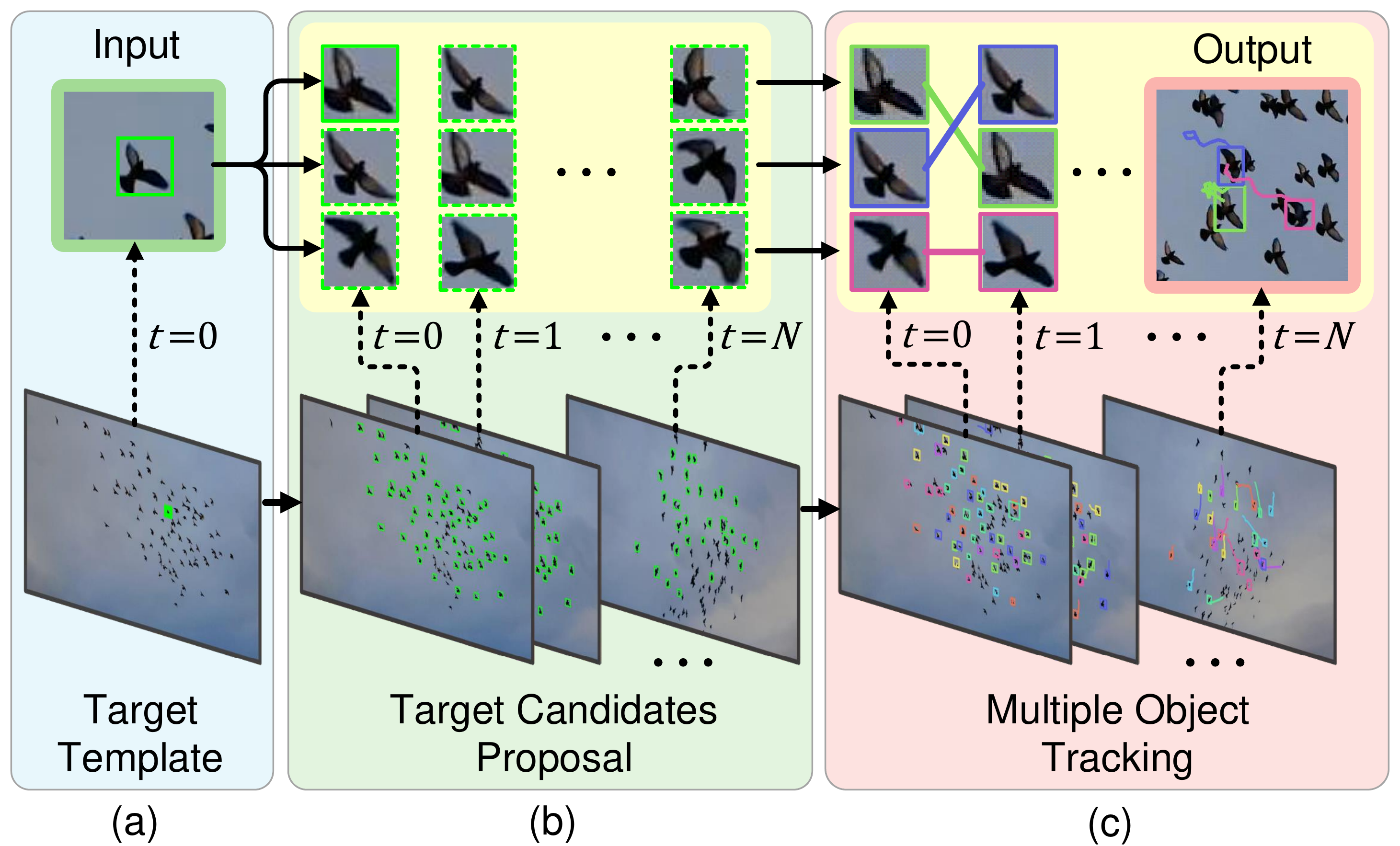}\\
    \end{center}
    \caption{One-shot generic multiple object tracking (GMOT). (a): The input of one-shot generic MOT is a single bounding box to indicate a target template in the first frame. (b): The target template is used to discover and propose all other target candidates of same category, which is different than model-based MOT where a pre-trained detector (typically class-specific) is required. (c): MOT then can be performed on the proposed candidates in either an online or offline manner. Yellow rectangles are zoomed-in local views of targets.}
    \label{fig:pipeline}
\end{figure}
Addressing the above issues, in this paper, we contribute to the study of GMOT in three aspects: dataset, baseline, and evaluation. First, we construct the first publicly available dense GMOT dataset, dubbed \textit{GMOT-40}, for systematical study of GMOT. GMOT-40 contains 40 carefully selected sequences, which cover ten categories (\eg, \textit{insect} and \textit{balloon}) with four sequences per category. Each sequence contains multiple objects of same category, and the average number of objects per frame is around 22. All sequences are manually annotated with careful validation/correction. The sequences involve many challenging factors such as heavy blur, occlusion, \etc\  
A tracking Protocol is adopted to evaluate different characteristics of tracking algorithms. The \textit{one-shot GMOT} \cite{luo2013generic, luo2014bi}, takes as input the bounding box of \textit{one} target object in the first frame, and aims to detect and track \textit{all} objects of the same category. Figure~\ref{fig:pipeline} illustrates the one-shot GMOT Protocol. 

Second, we design a series of baseline tracking algorithms dedicated to one-shot GMOT. These baselines consist of a one-shot detection stage and a target association stage. The one-shot detection stage is adapted from the recently proposed GlobalTrack algorithm~\cite{Huang_Zhao_Huang_2020}. The target association stage comes from several typical MOT algorithms. For each baseline, the one-shot detection algorithm plays the role of public detector. 

Third, we conduct thorough evaluations on GMOT-40. The evaluation involves both classic tracking algorithms (\eg, \cite{bochinski2017high,wojke2017simple, xiang2015learning}) and recently proposed one (\eg, \cite{chu2019famnet}), with necessary modifications. The results show that, as an important tracking problem, GMOT has a large room for improvement.

To summarize, we make three contributions in this paper:
\begin{itemize}
    \item the first publicly available dense GMOT dataset, GMOT-40, which is carefully designed and annotated, along with evaluation Protocol, 
    \item a series of GMOT baselines adapted from modern deep-learning enhanced MOT algorithm, and \item thorough evaluations and analysis on GMOT-40. 
\end{itemize}


\section{Related Work}



\begin{table}[t]
\centering
\caption{Comparison of densely annotated data used in GMOT studies. \# seq: number of sequence, \# cat: number of categories, \# tgt: average number of targets per frame. $^\star$: Estimated from samples in the paper.}
\label{tab: model-free trackers}
\begin{tabular}{p{0.3\linewidth}p{0.1\linewidth}p{0.1\linewidth}p{0.1\linewidth}p{0.15\linewidth}}
\hline\thickhline
Publication & Year & \# seq. & \# cat. &  \# tgt. \\ 
\hline
Luo \etal \cite{luo2013generic} &  2013 & 4 & 4 & $\approx$15$^\star$ \\
Zhang \etal \cite{zhang2013preserving} & 2014 & 9 & 9 & $\approx$3$^\star$  \\
Luo \etal \cite{luo2014bi} & 2014 & 8 & 8 & $\approx$15$^\star$  \\
Zhu \etal \cite{zhu2016model}  & 2017 & 3 & 1 & 13.13 \\
Liu \etal \cite{Ionreid2020} & 2020 & 24 & 9 & 3.375 \\
\hline
GMOT-40 & 2021 & \textbf{40} & \textbf{10} & \textbf{26.58}  \\
\hline\thickhline
\end{tabular}
\end{table}


\subsection{MOT Algorithms}

Multiple object tracking (MOT) has been an active research area for decades~\cite{ciaparrone2020deep,luo2014multiple}. Based on whether the target priors are presumed to the tracker, MOT approaches can be roughly categorized as model-based and model-free methods. In the context of model-based methods, the most popular framework is the tracking-by-detection one where a category-aware detector is employed for generating candidate proposals, and the tracker itself primarily focuses on solving the data association problem. Many methods have been investigated under this framework, such as Hungarian algorithm~\cite{bewley2016simple,fang2018recurrent,huang2008robust}, network flow~\cite{dehghan2015target,zamir2012gmcp,zhang2008global}, graph multicut~\cite{lifted_disjoint_paths_2020_ICML,keuper2015efficient,tang2017multiple}, multiple hypotheses tracking~\cite{chen2017enhancing,kim2015multiple} and multi-dimensional assignment~\cite{collins2012multitarget,shi2019rank} using a variety of affinity estimation schemes. With recent advances in deep learning, deep neural networks are also learned to solve the data association problem~\cite{braso2020learning,chu2019famnet, milan2017online}. 

Model-based MOT methods can automatically handle the entering and exiting events of targets. However, it heavily depends on using target priors by employing a category detector or the Re-identification~(ReID) based affinity estimator. Therefore, most recent MOT methods in this category focus on pedestrian and vehicle tracking. For example, there is an increasing popularity in the community to leverage ReID dataset~\cite{li2014deepreid,ristani2016performance, zheng2015scalable} or pose estimation dataset~\cite{andriluka2018posetrack} to improve association robustness during tracking~\cite{braso2020learning,henschel2019multiple,karthik2020simple, zhan2020simple}, while others adopt the state-of-the-art person detection techniques, such as~\cite{bergmann2019tracking, han2020mat,pang2020tubetk,peng2020chained,shan2020fgagt}. These detection and ReID networks are trained and hence limited by the available datasets, therefore, the generic targets will not be handled and tracked successfully by methods in this category.

Despite the dominant effort on the person and vehicle tracking, there are a number of works that have focused on other target categories. Cell tracking~\cite{bise2011reliable,mavska2014benchmark,ulman2017objective,yang2018multiple} is a popular topic in this section. Detecting and tracking multiple objects, such as ants~\cite{khan2004mcmc}, bats~\cite{betke2007tracking}, birds~\cite{luo2014bi}, bees~\cite{bozek2018towards} and fish~\cite{fontaine2007model,spampinato2008detecting,spampinato2012covariance} are also investigated. Methods proposed in those works also need special modeling of target appearance or motion pattern thus cannot be applied generally in generic targets either.

Model-free methods contribute another category of solutions to MOT. Tracking without target prior is primarily proposed for solving \textit{Single Object Tracking} (SOT) where only one bounding box of target is given at the first frame and no category prior is known to the tracker. It is an emerging topic to extend the model-free idea to the context of MOT. However there is no unified framework so far. In~\cite{zhang2013preserving}, structure information is used to help the tracking of multiple appearance-wise similar objects. Appearance and motion models are learned in~\cite{Ionreid2020} to tackle sudden appearance change and occlusion. Both the two methods need the manual initialization of all targets. In~\cite{zhu2016model}, a generic category independent object proposal module is used to generate target candidates. Luo \etal.~\cite{luo2014bi} proposed to use clustered Multiple Task Learning for generic object detection. All these works are evaluated on datasets that either have limited number of sequences or limited number of target categories.



\subsection{MOT Benchmarks}

There are multiple benchmark datasets for model-based MOT. One of the oldest benchmarks is the PETS benchmark~\cite{ferryman2009pets2009} which contains three sequences for single camera MOT while all of them are on pedestrians. Later on, a benchmark mainly for autonomous driving is KITTI \cite{geiger2012we} which contains two categories of pedestrian and vehicle. After that, a benchmark dataset solely on pedestrian tracking was proposed by Alahi \etal~\cite{alahi2014socially}. Although this benchmark contains 42 million pedestrian trajectories, yet its annotation is not high-quality (\ie, not annotated by human). Then a MOT benchmark dataset on vehicle tracking was released with the name UA-DETRAC~\cite{wen2020ua} which contains 100 sequences. In the same year MOT15 was released~\cite{MOTChallenge2015} which organized the publicly available MOT data by then and became one of the most popular MOT benchmarks. Yet it is worth noting that there are just two categories: people and vehicle in this benchmark, and only 22 sequences are included. Later, MOT16 \cite{MOT16} was published with 14 sequences,  devoted to people and vehicle tracking. VisDrone ~\cite{zhu2020vision} was released with 96 sequences focused on vehicle and people. 

In addition to the popular MOT benchmark dataset mentioned above on people and vehicle tracking, there are some other benchmark datasets on special classes such as honey bees and cells. For example, the multiple cell tracking dataset~\cite{ulman2017objective} has 52 sequences with a focus on cell, the honey-bee tracking dataset~\cite{bozek2018towards} has 60 sequences of the honey bee. 

As shown in Table~\ref{tab: model-free trackers}, high quality datasets dedicated for model-free MOT are rare. In~\cite{zhang2013preserving}, Zhang~\etal collect a dataset with nine video sequences, each for a different type of target. Among the videos, three are adapted from a SOT dataset, while the rest videos are collected from YouTube. The dataset contains average of 3 targets per frame. Each video here has average of 842 frames in length. Targets in the dataset are present all-time in the video, which relieves the tracker of handling the entering and exiting event of targets. Luo \etal collected datasets with four and eight videos in \cite{luo2013generic} and \cite{luo2014bi} respectively for an early study of GMOT. Recent works~\cite{zhu2016model, Ionreid2020} tend to use mixed sequences picked from other SOT or multiple pedestrian tracking datasets. Recently, a large-scale benchmark for tracking any object (TAO) is proposed~\cite{dave2020tao}. However, TAO is not densely annotated and has low annotation quality. Only one out of every 30 frames is annotated by hand, and the average trajectories of TAO in each sequence is only $5.9$. Besides, the task of TAO is to track multiple objects of different classes, which differs with the GMOT concept in this paper. Hence we do not include TAO in comparison Table~\ref{tab: model-free trackers}.

Compared with the data used in previous studies, our proposed GMOT-40 dataset provides the the first publicly available dense dataset on GMOT. GMOT-40 contains more sequences and categories than previous GMOT datasets. Moreover, the target density in GMOT-40 is much higher than existing datasets, \eg, 26.58 per sequence \textit{vs} 5.9 per sequence in TAO, and the sequences involve many real-world challenges such as entering and exiting events, fast motion, occlusion, \etc. As a result, the release of GMOT-40 is expected to largely facilitate future research in GMOT.

\section{The Generic MOT Dataset GMOT-40}

\begin{figure*}[!h]
 	\small
	\centering
	\begingroup
	\tabcolsep=0.5mm
    \def\arraystretch{0.10}
	\begin{tabular}
		{ccccc}%
		\includegraphics[width=0.195\linewidth,height=0.12\linewidth]{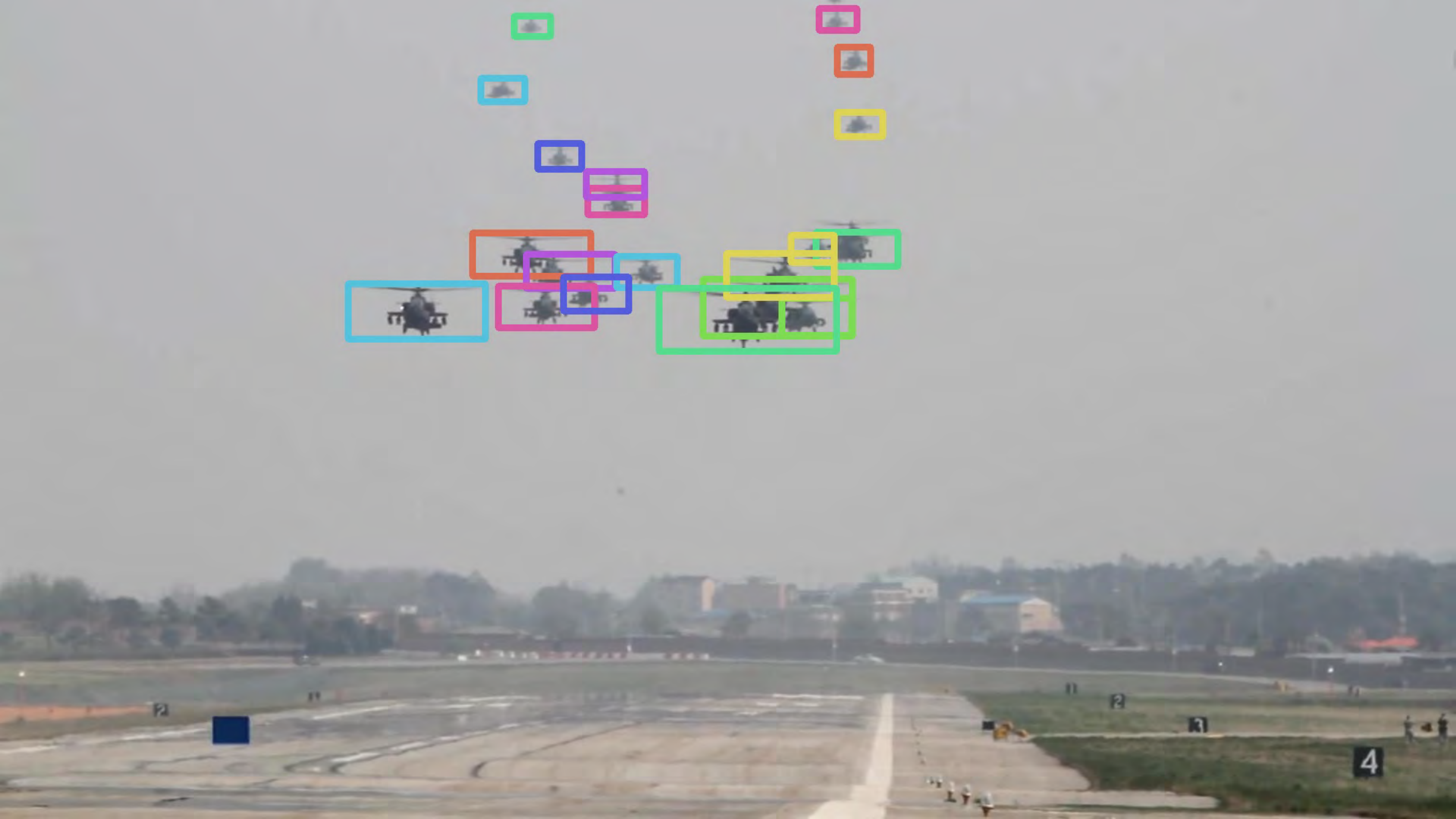}    & 
		\includegraphics[width=0.195\linewidth,height=0.12\linewidth]{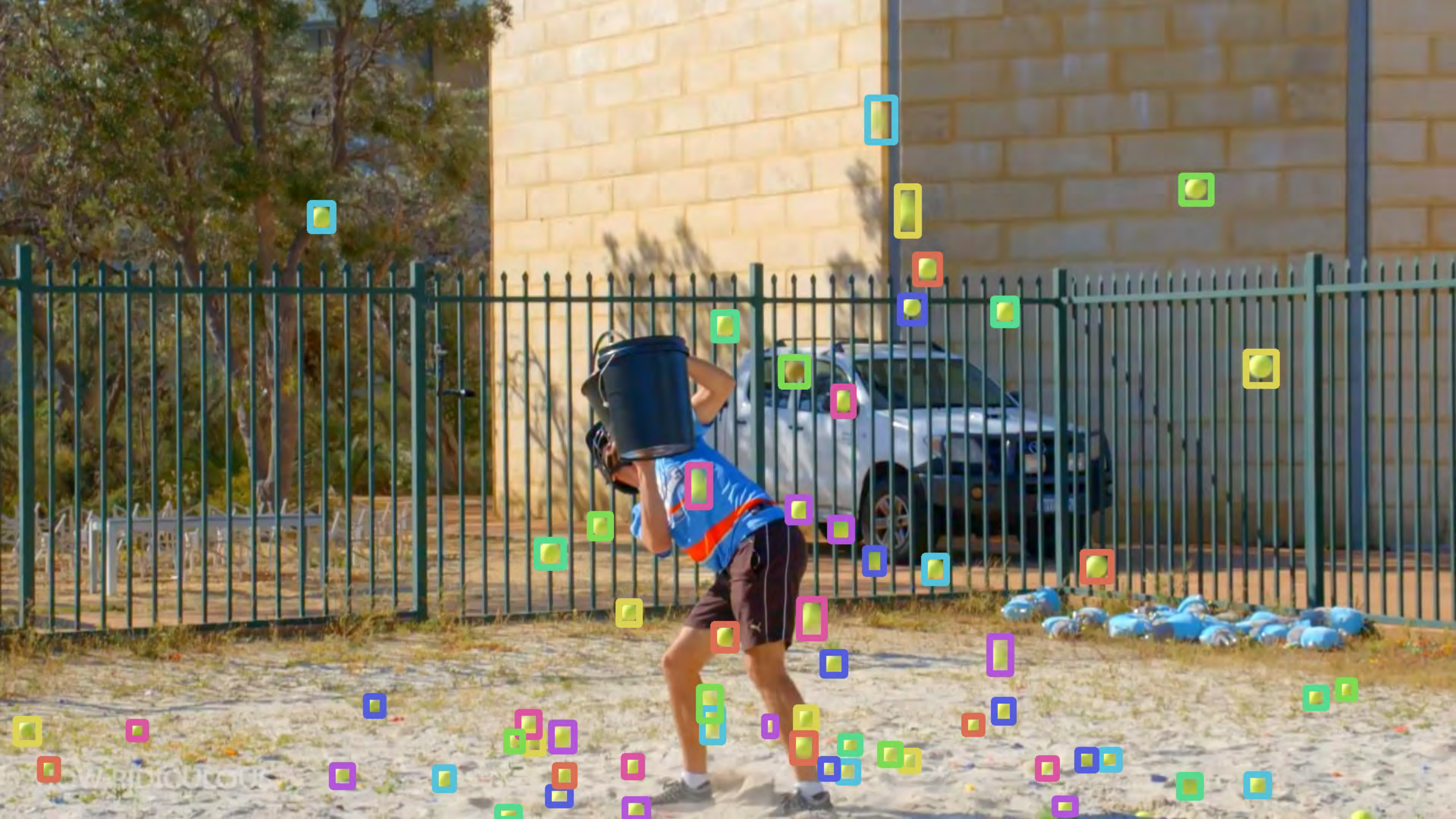}    & 
		\includegraphics[width=0.195\linewidth,height=0.12\linewidth]{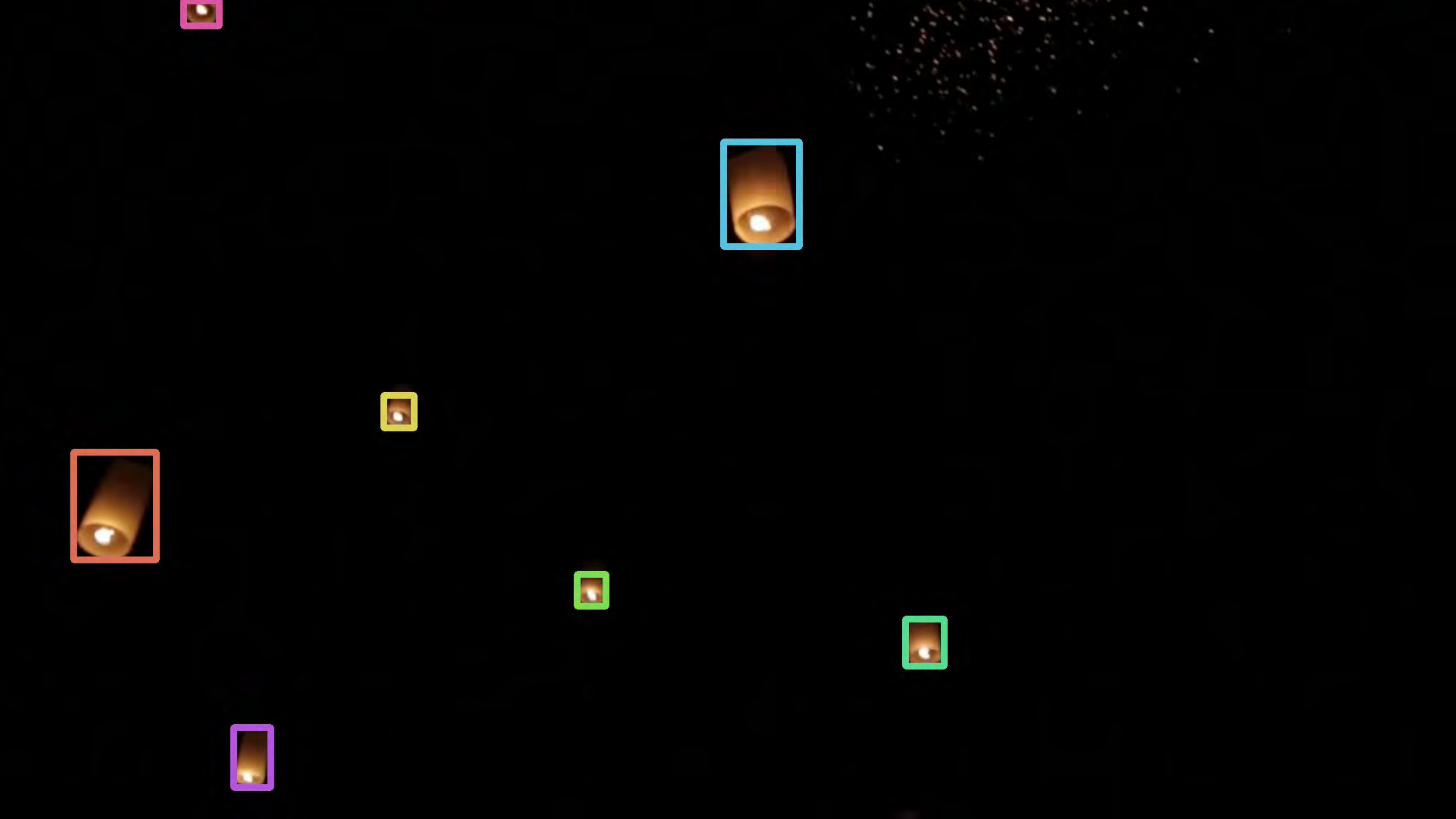}    &
		\includegraphics[width=0.195\linewidth,height=0.12\linewidth]{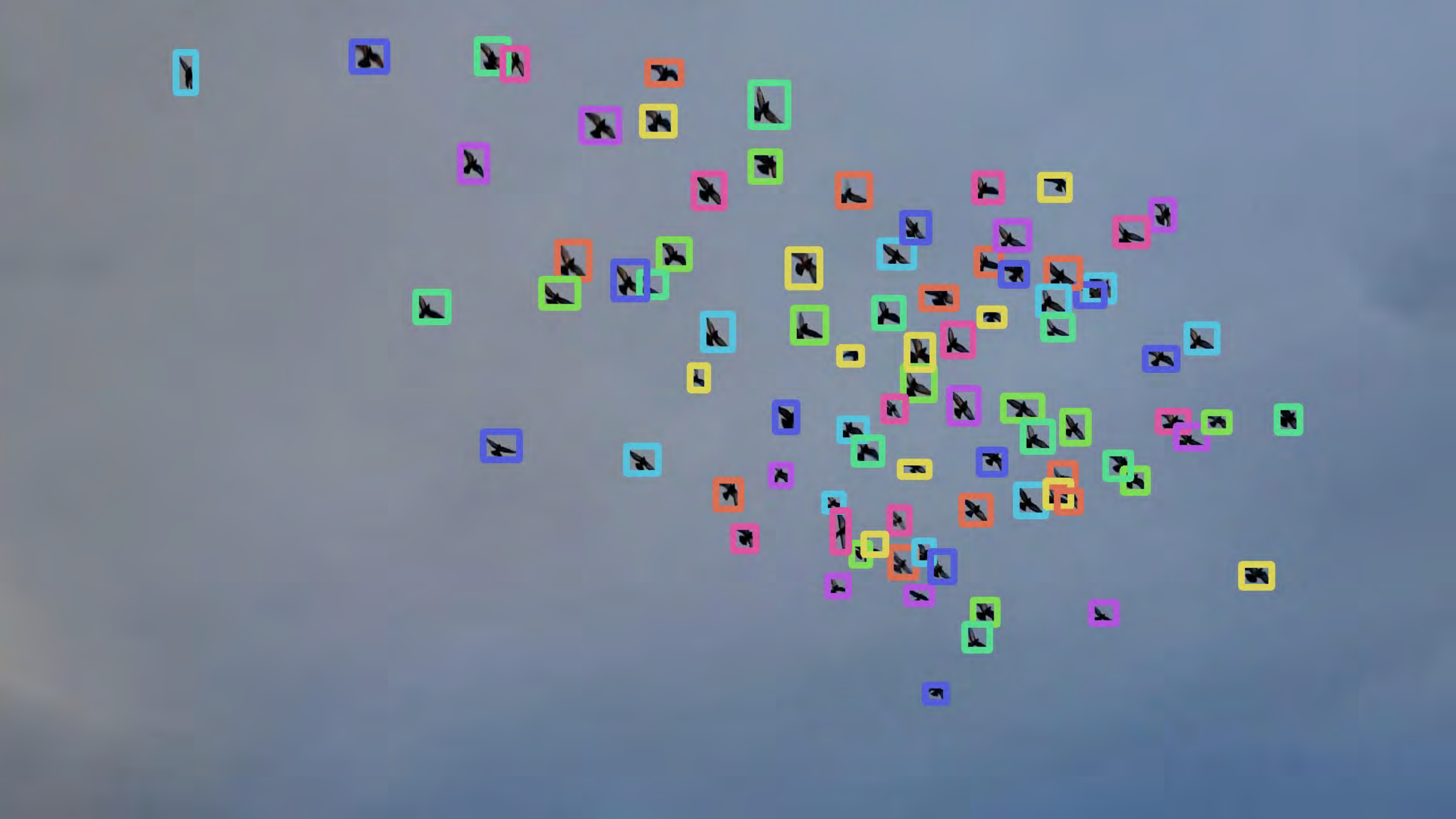}    &
		\includegraphics[width=0.195\linewidth,height=0.12\linewidth]{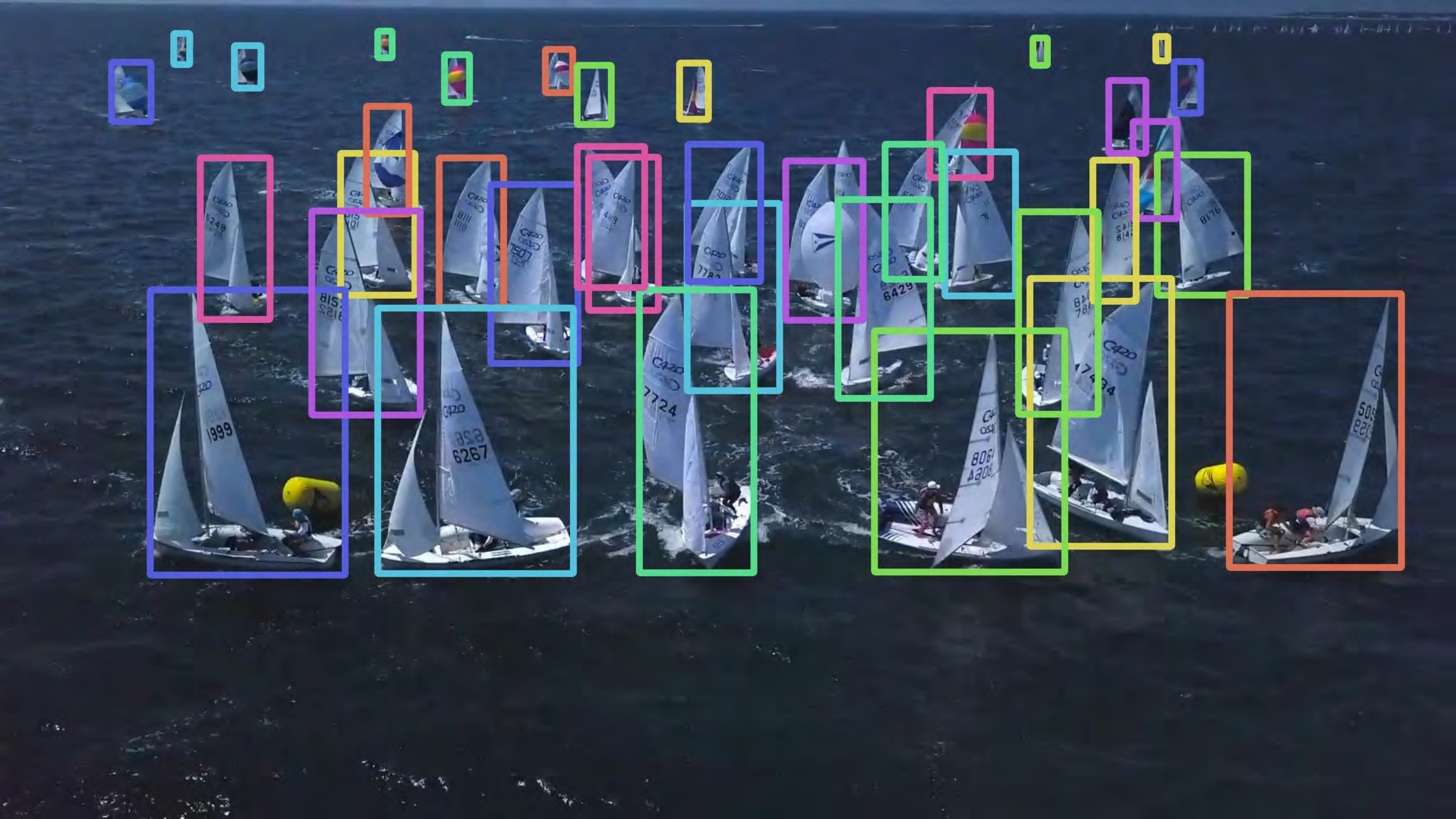} 
		\\
		\vspace{0.1mm}\\
		\\
		\includegraphics[width=0.195\linewidth,height=0.12\linewidth]{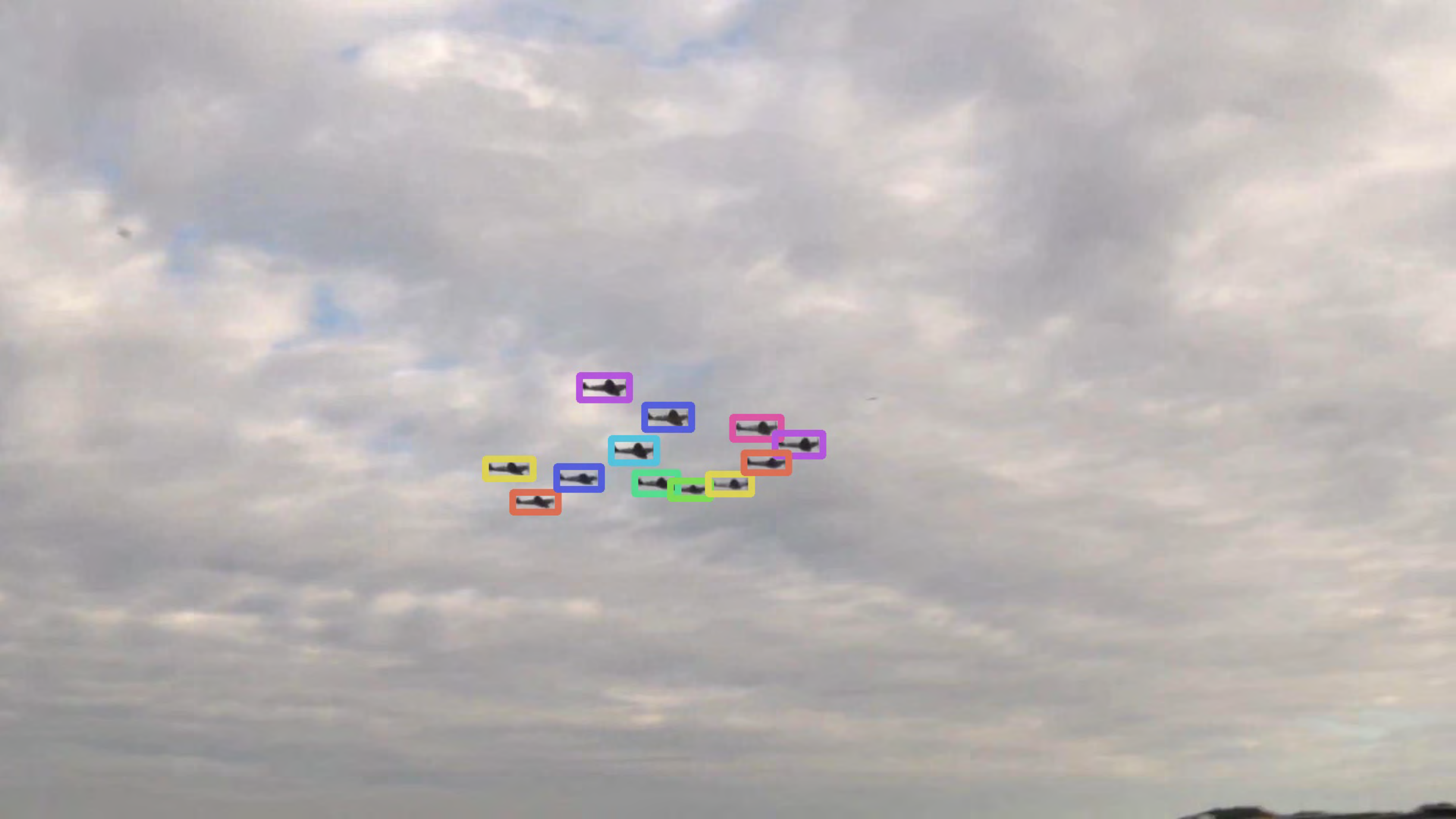}    & 
		\includegraphics[width=0.195\linewidth,height=0.12\linewidth]{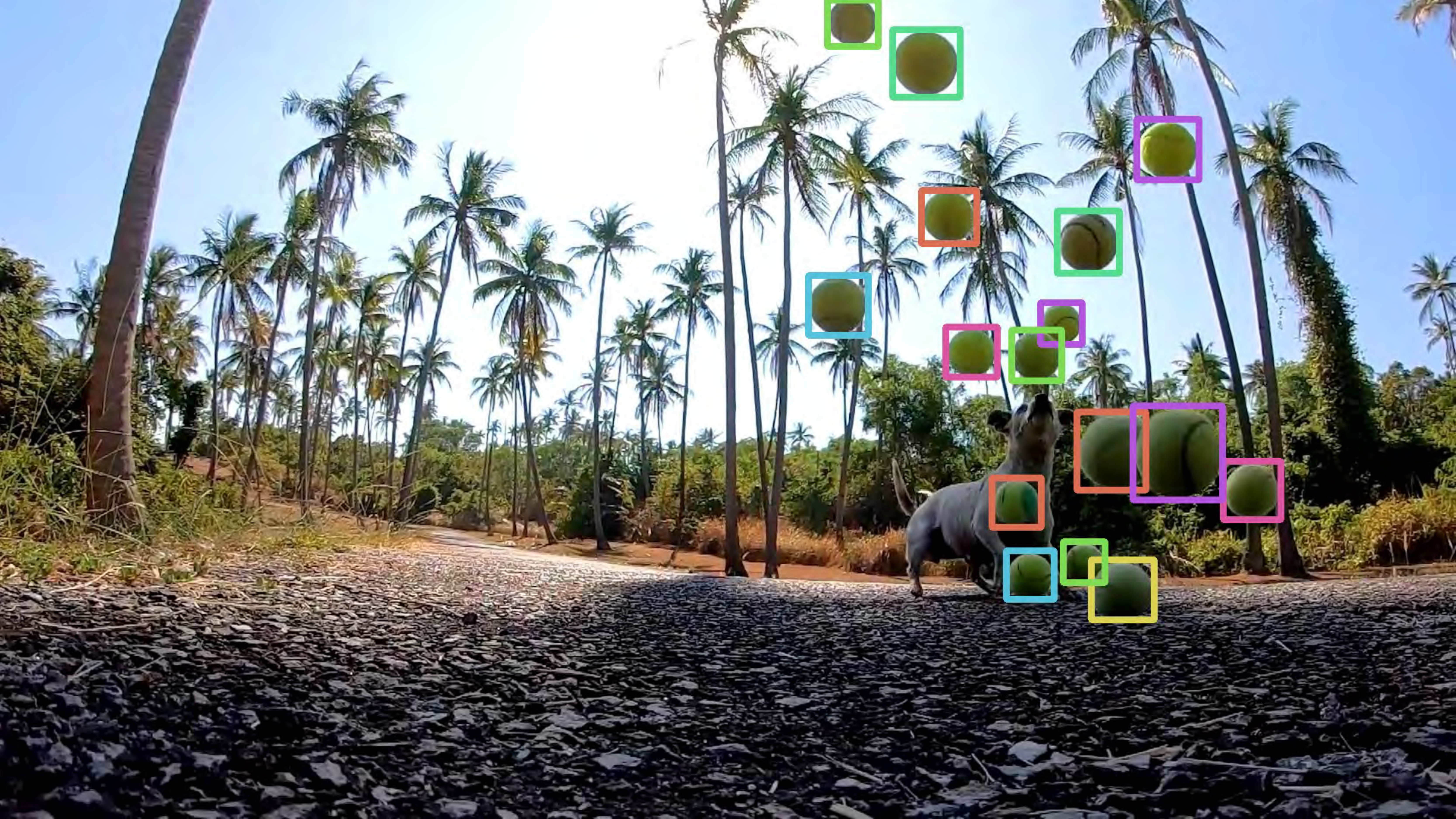}    & 
		\includegraphics[width=0.195\linewidth,height=0.12\linewidth]{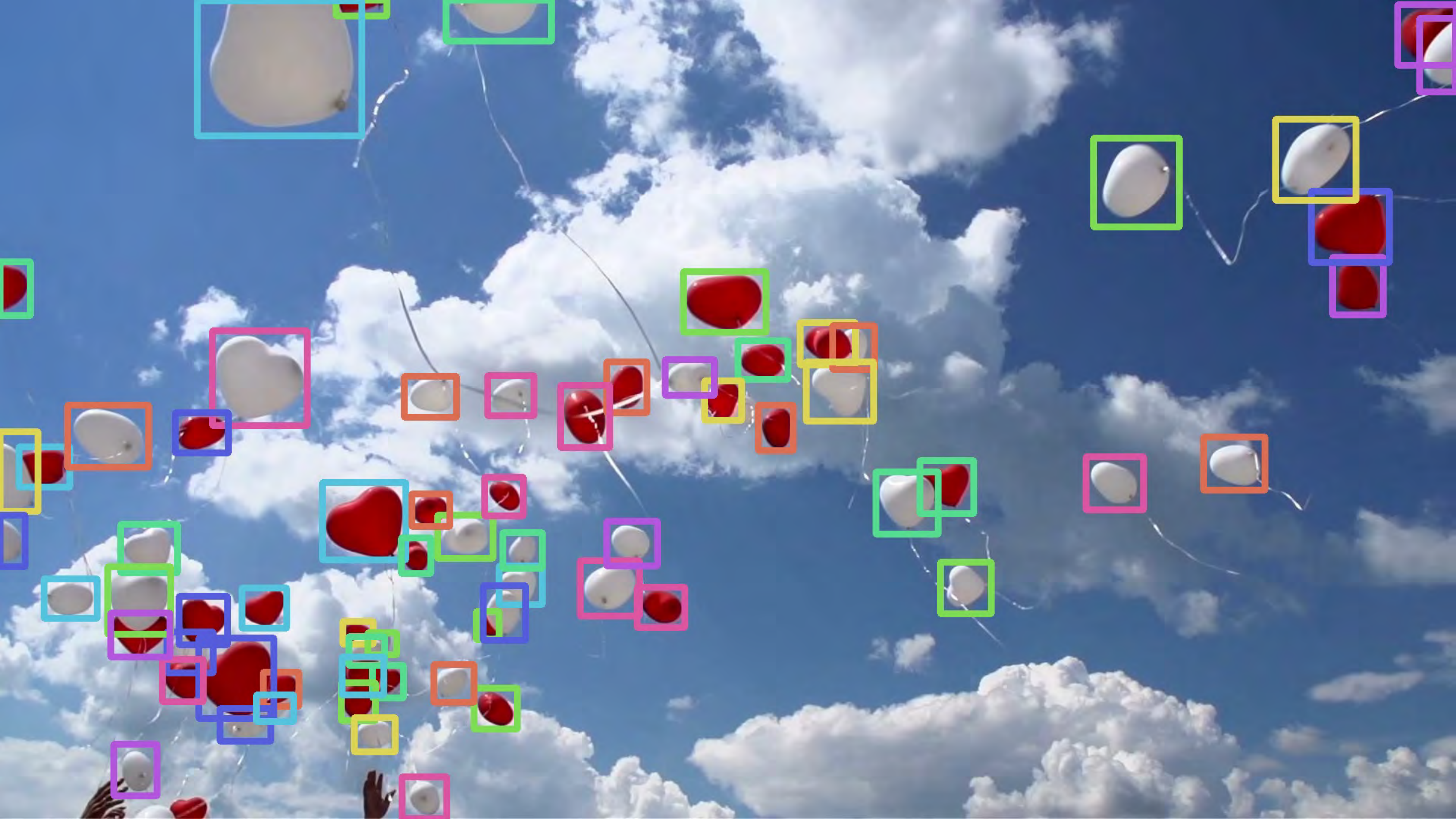}    &
		\includegraphics[width=0.195\linewidth,height=0.12\linewidth]{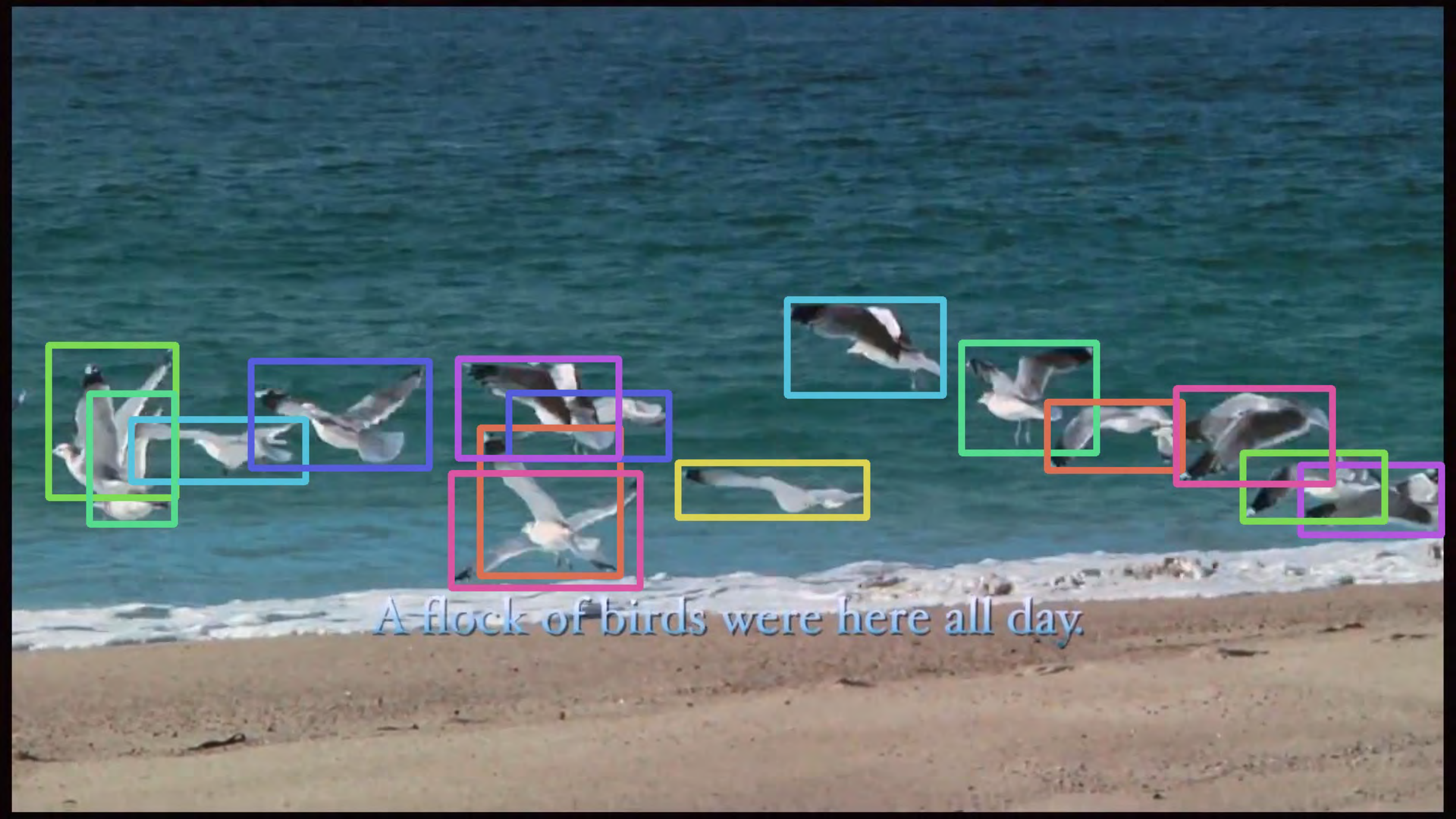}    &
		\includegraphics[width=0.195\linewidth,height=0.12\linewidth]{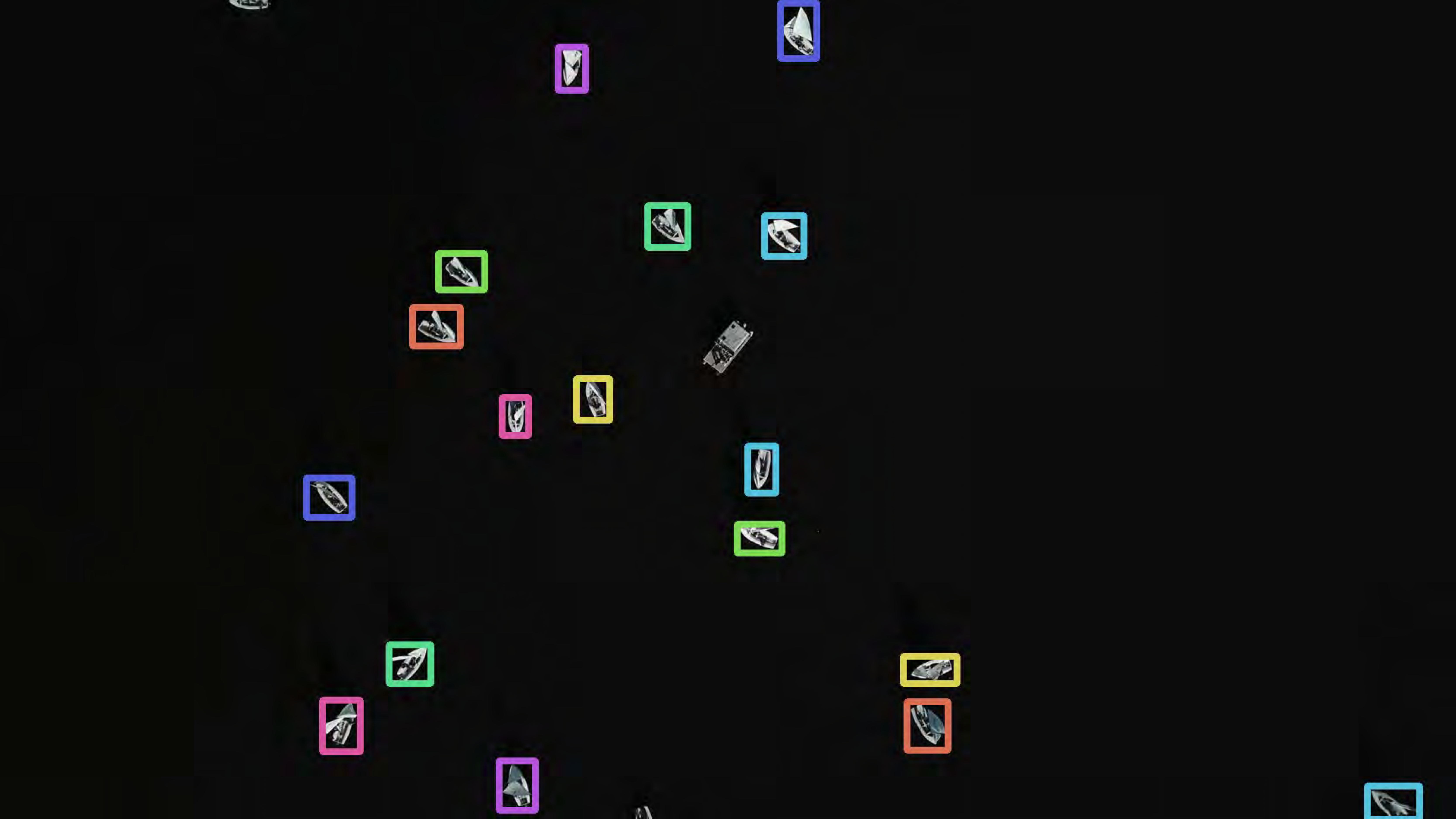} \\
		\rule{0pt}{0.75ex}\\
		airplane & ball & balloon & bird & boat \vspace{1mm}\\
		\rule{0pt}{0.75ex}\\
		\includegraphics[width=0.195\linewidth,height=0.12\linewidth]{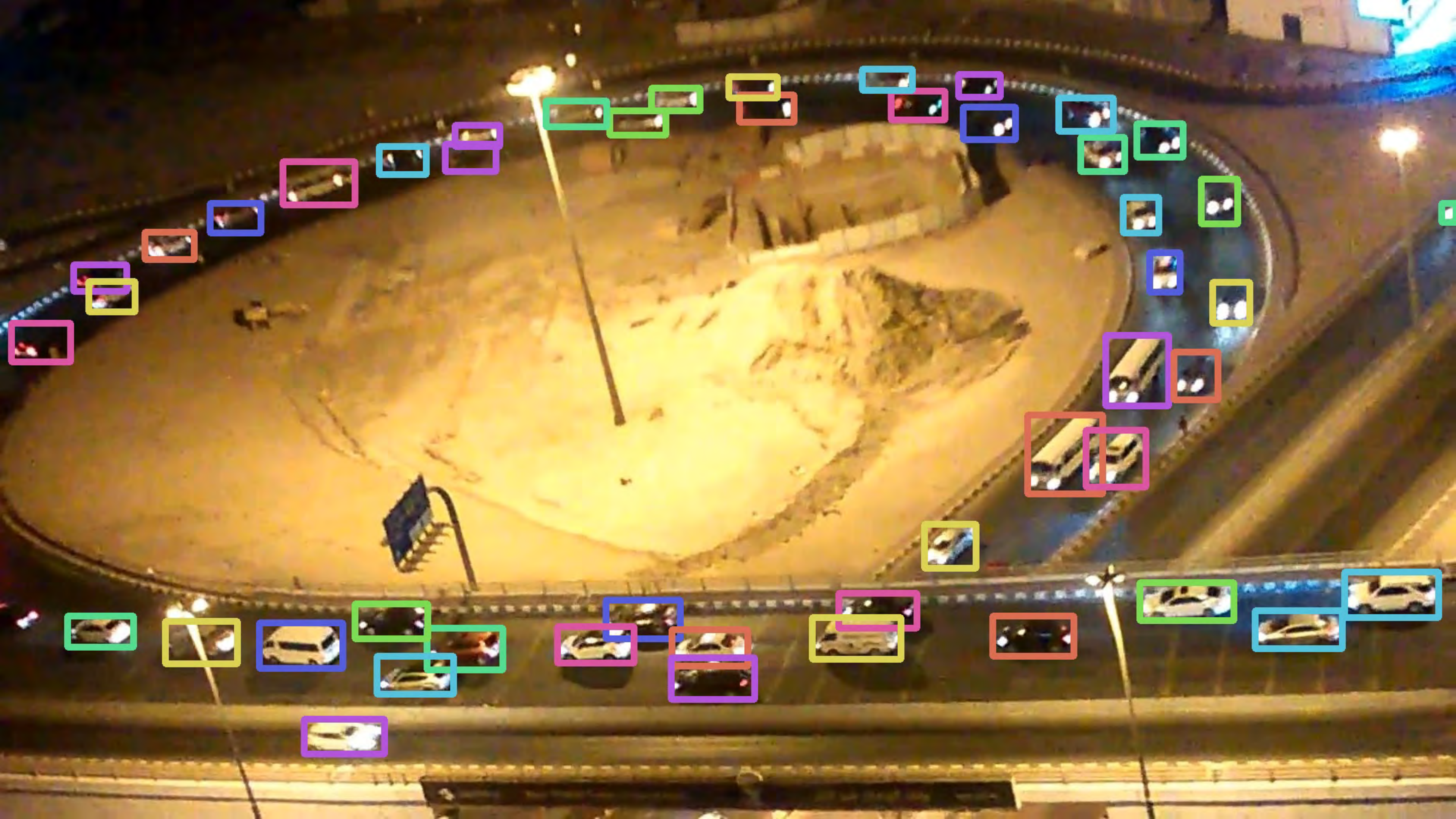}    & 
		\includegraphics[width=0.195\linewidth,height=0.12\linewidth]{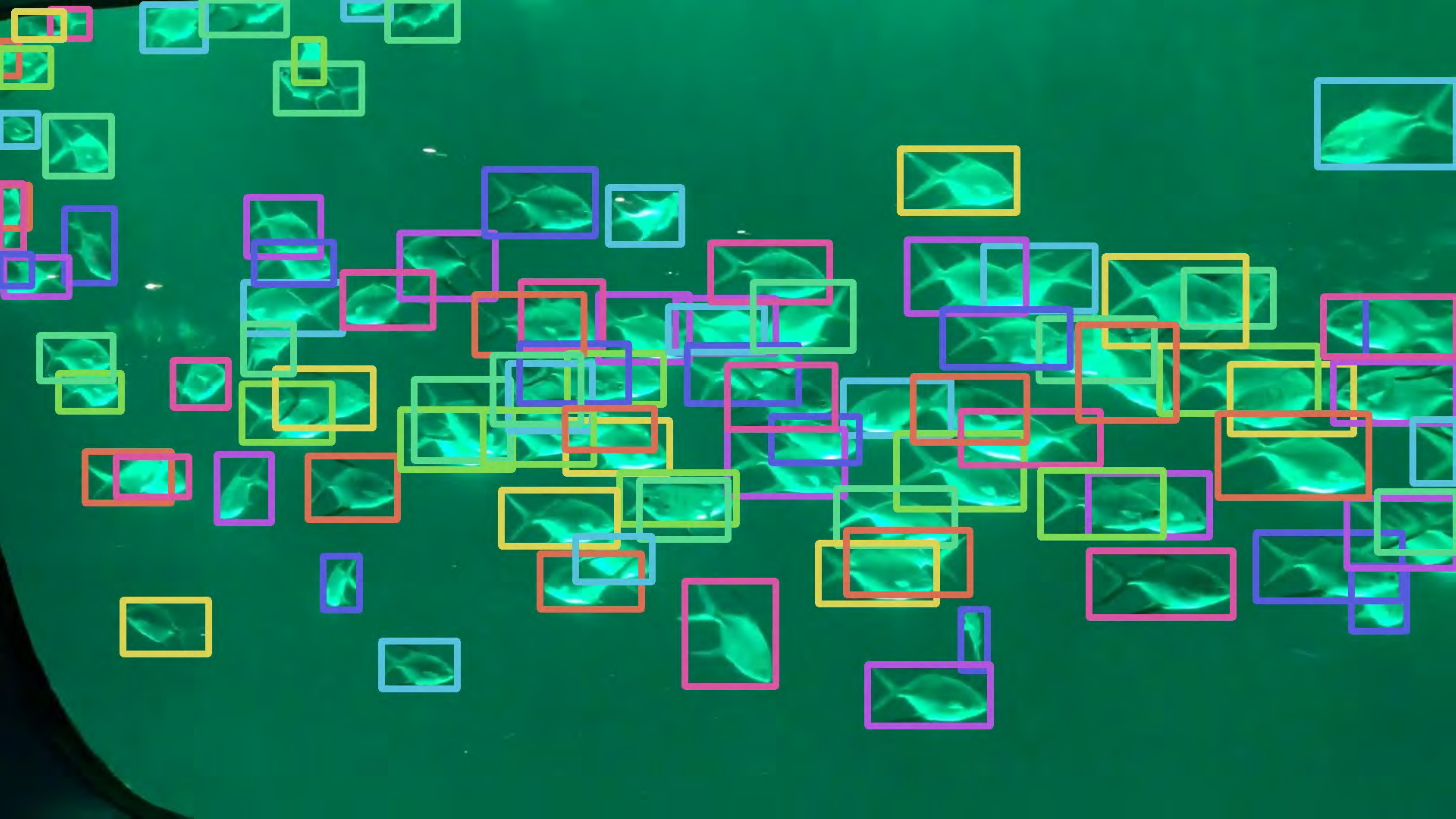}    & 
		\includegraphics[width=0.195\linewidth,height=0.12\linewidth]{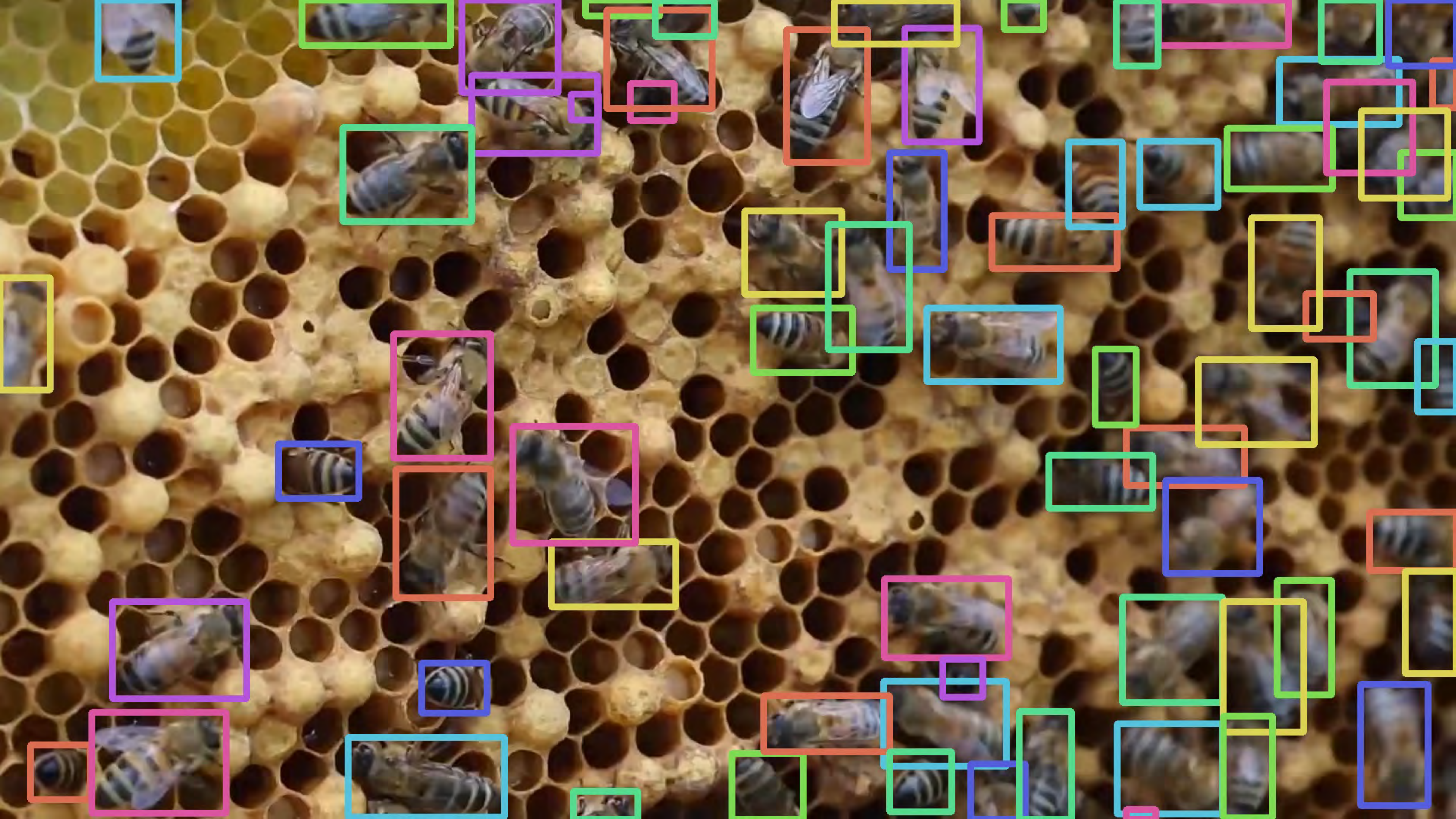}    &
		\includegraphics[width=0.195\linewidth,height=0.12\linewidth]{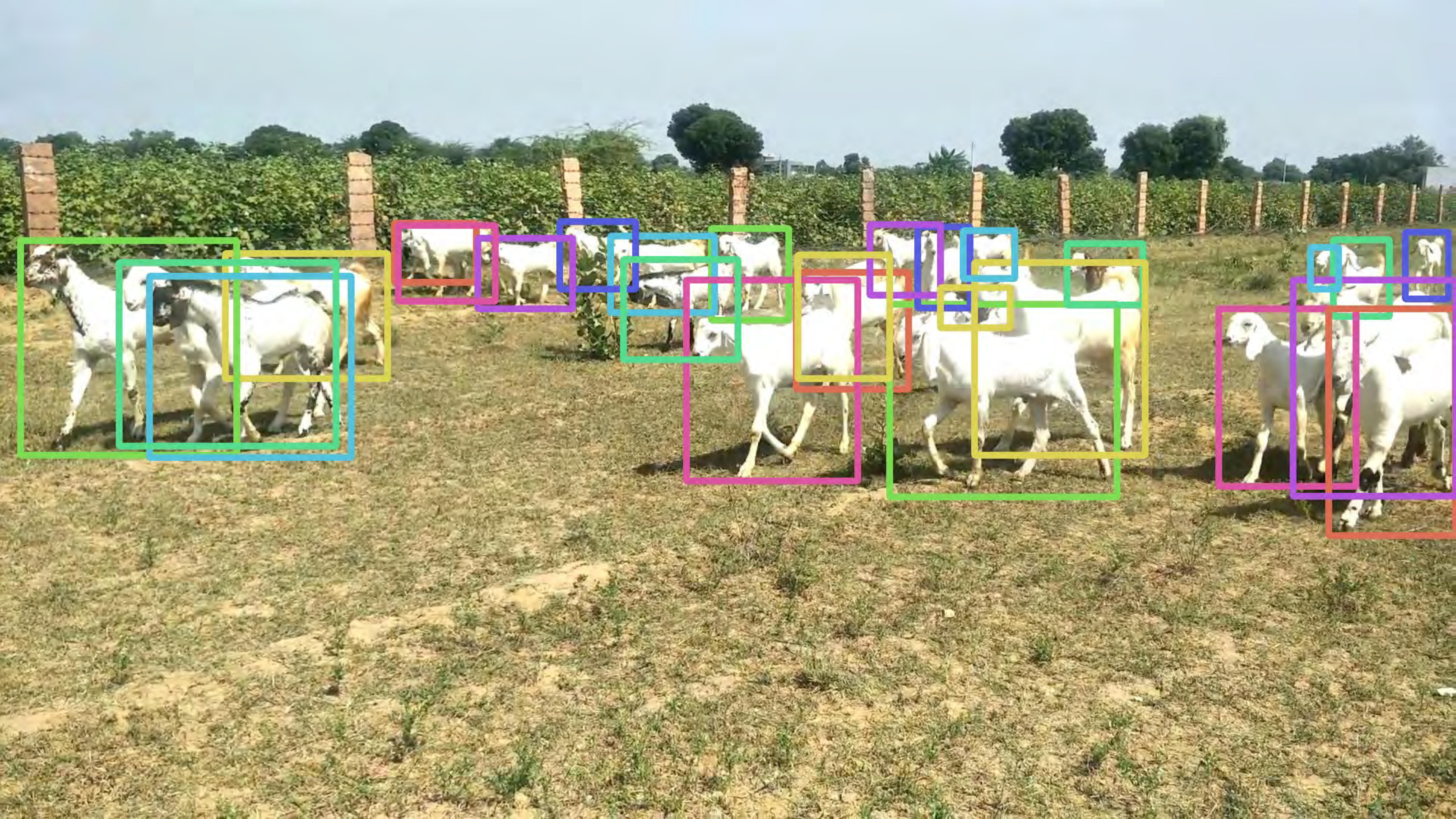}    &
		\includegraphics[width=0.195\linewidth,height=0.12\linewidth]{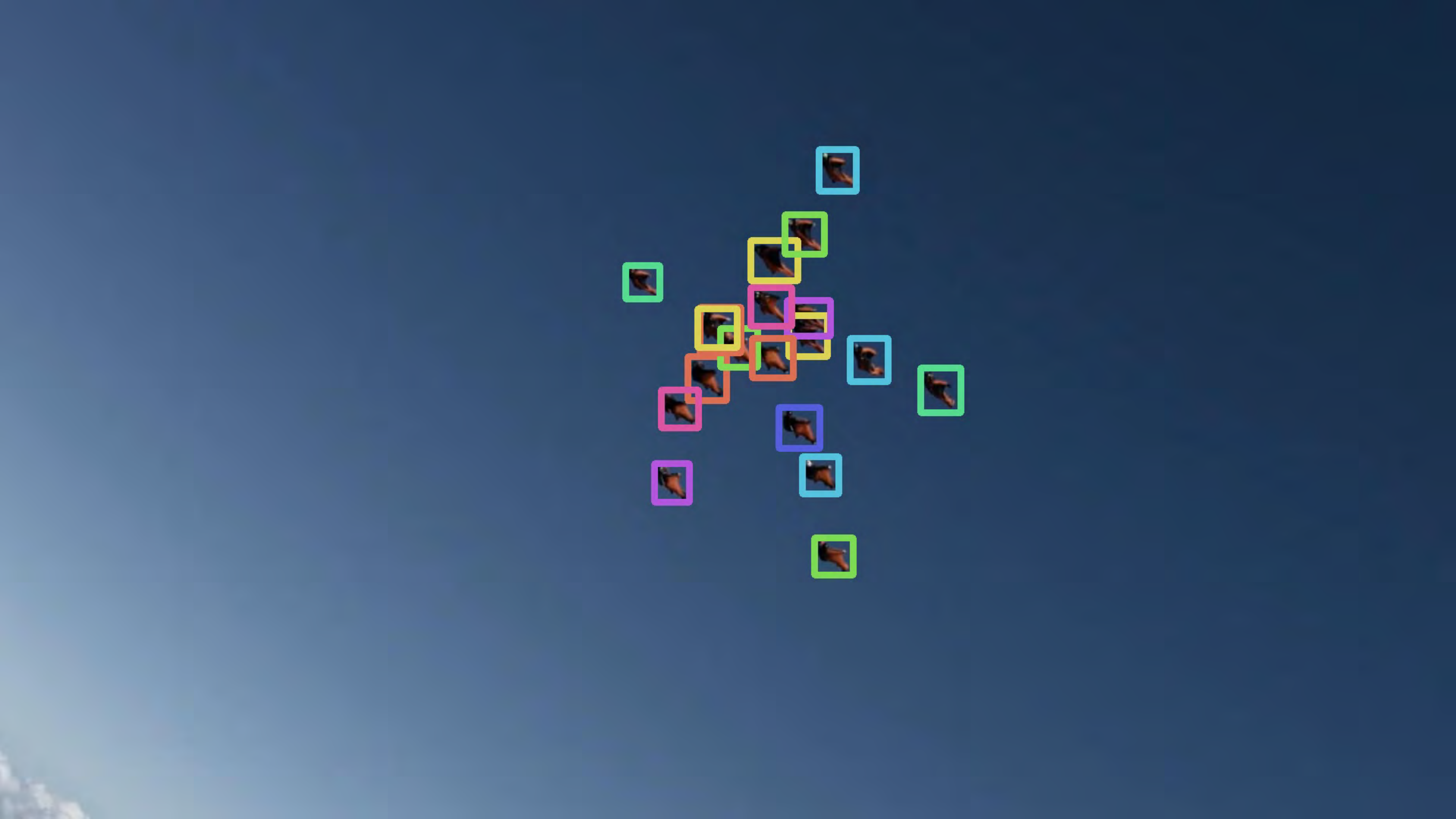} 
		\\
		\vspace{0.15mm}\\
		\\
		\includegraphics[width=0.195\linewidth,height=0.12\linewidth]{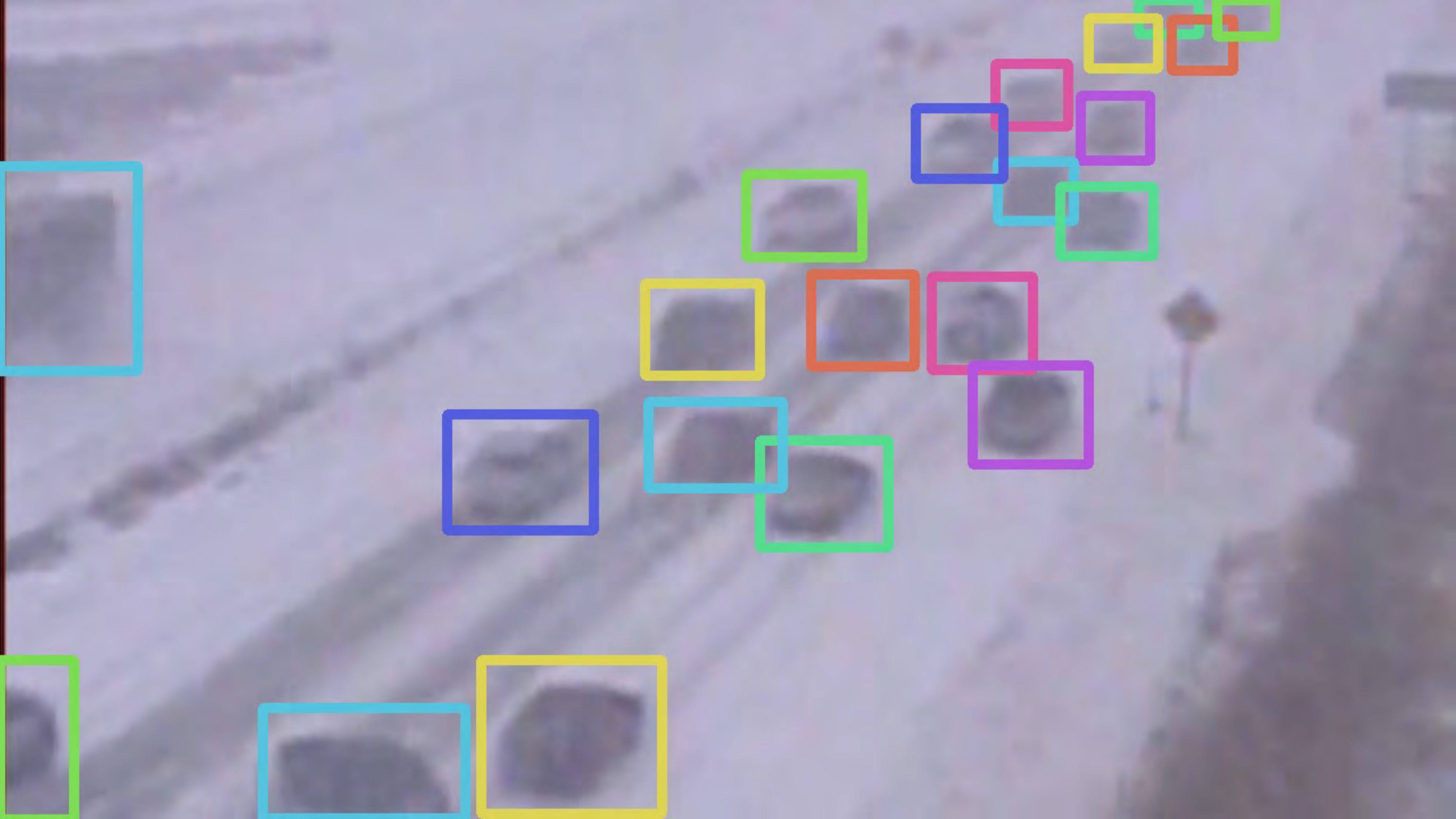}    & 
		\includegraphics[width=0.195\linewidth,height=0.12\linewidth]{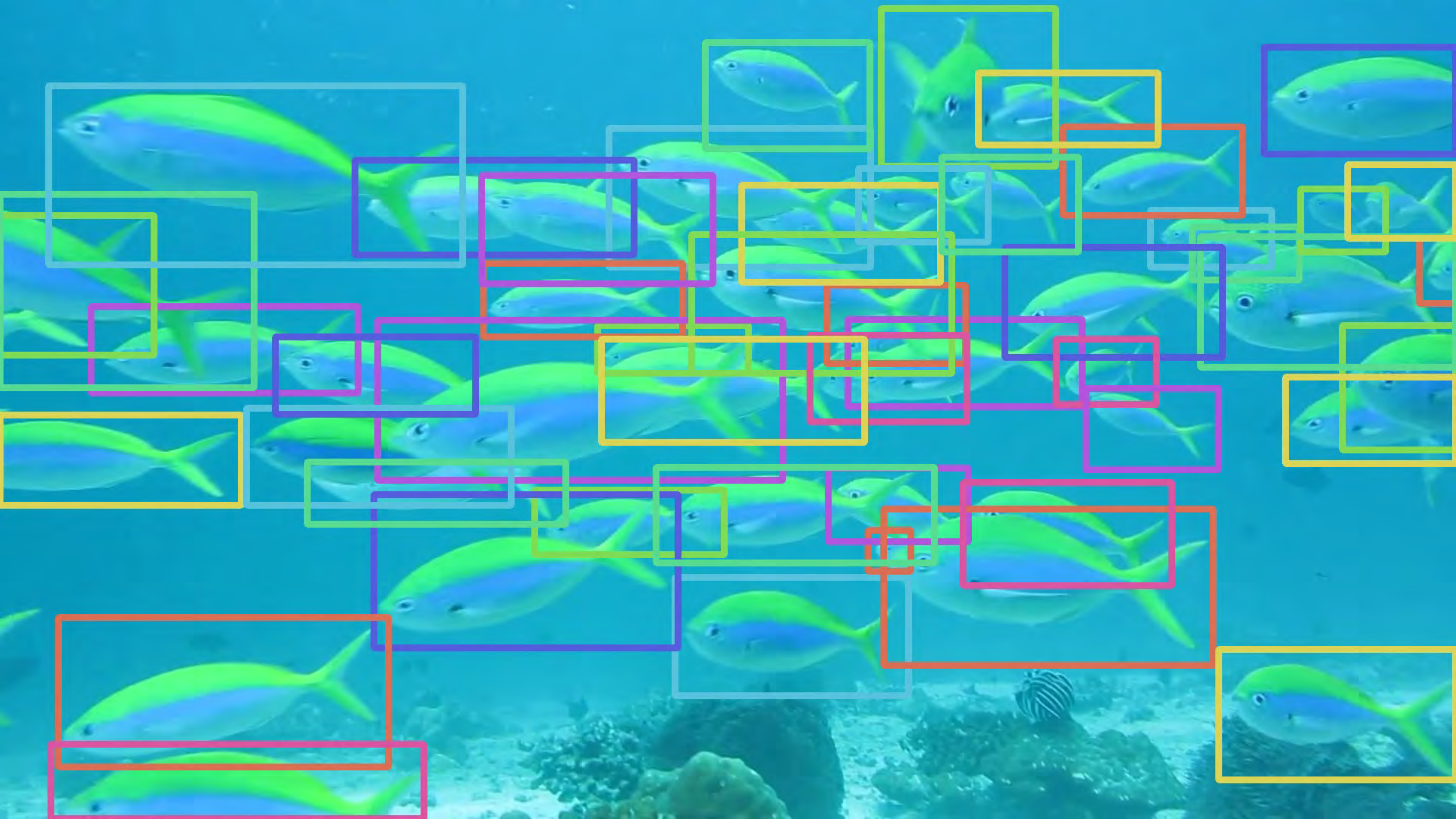}    & 
		\includegraphics[width=0.195\linewidth,height=0.12\linewidth]{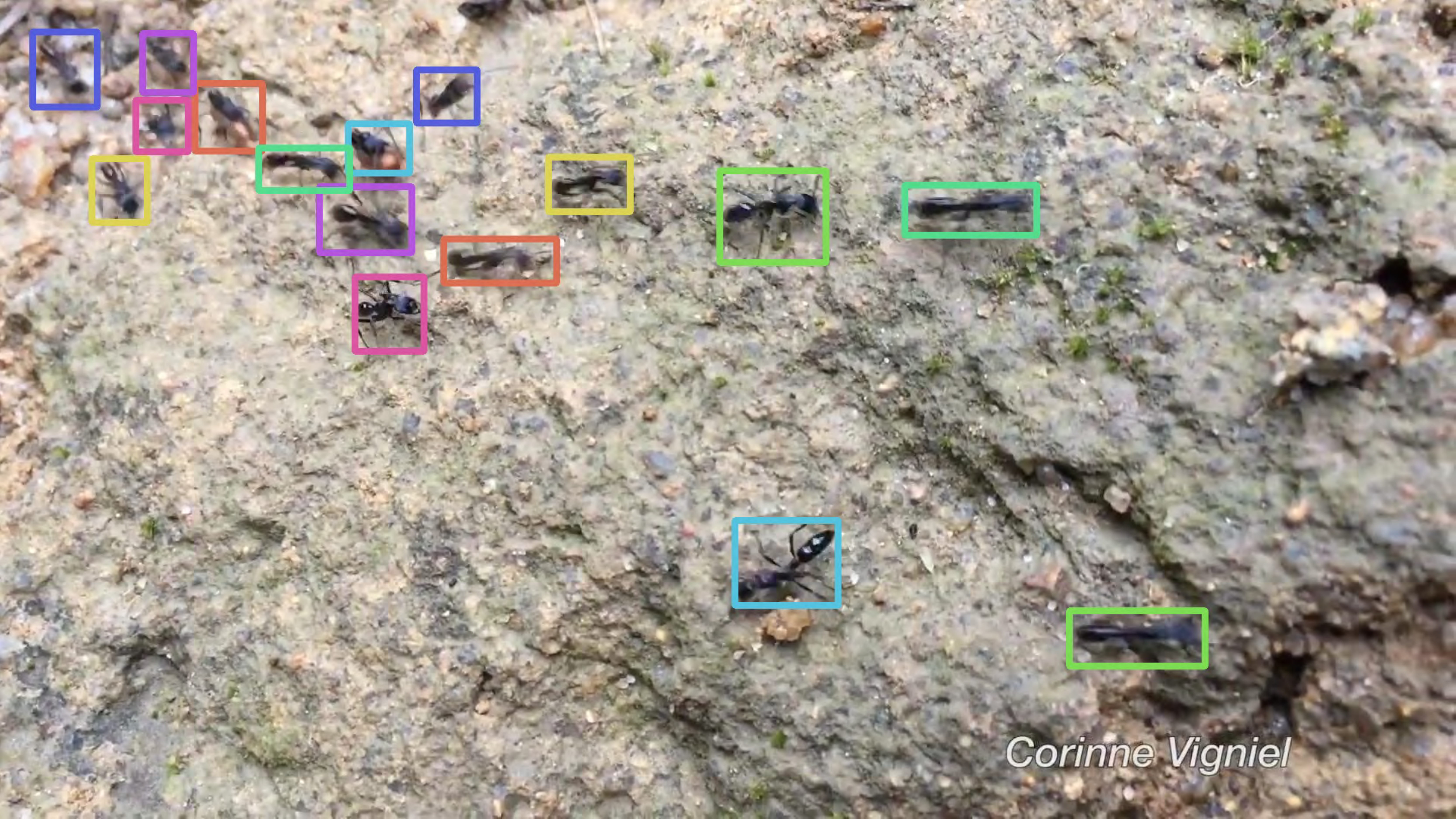}    &
		\includegraphics[width=0.195\linewidth,height=0.12\linewidth]{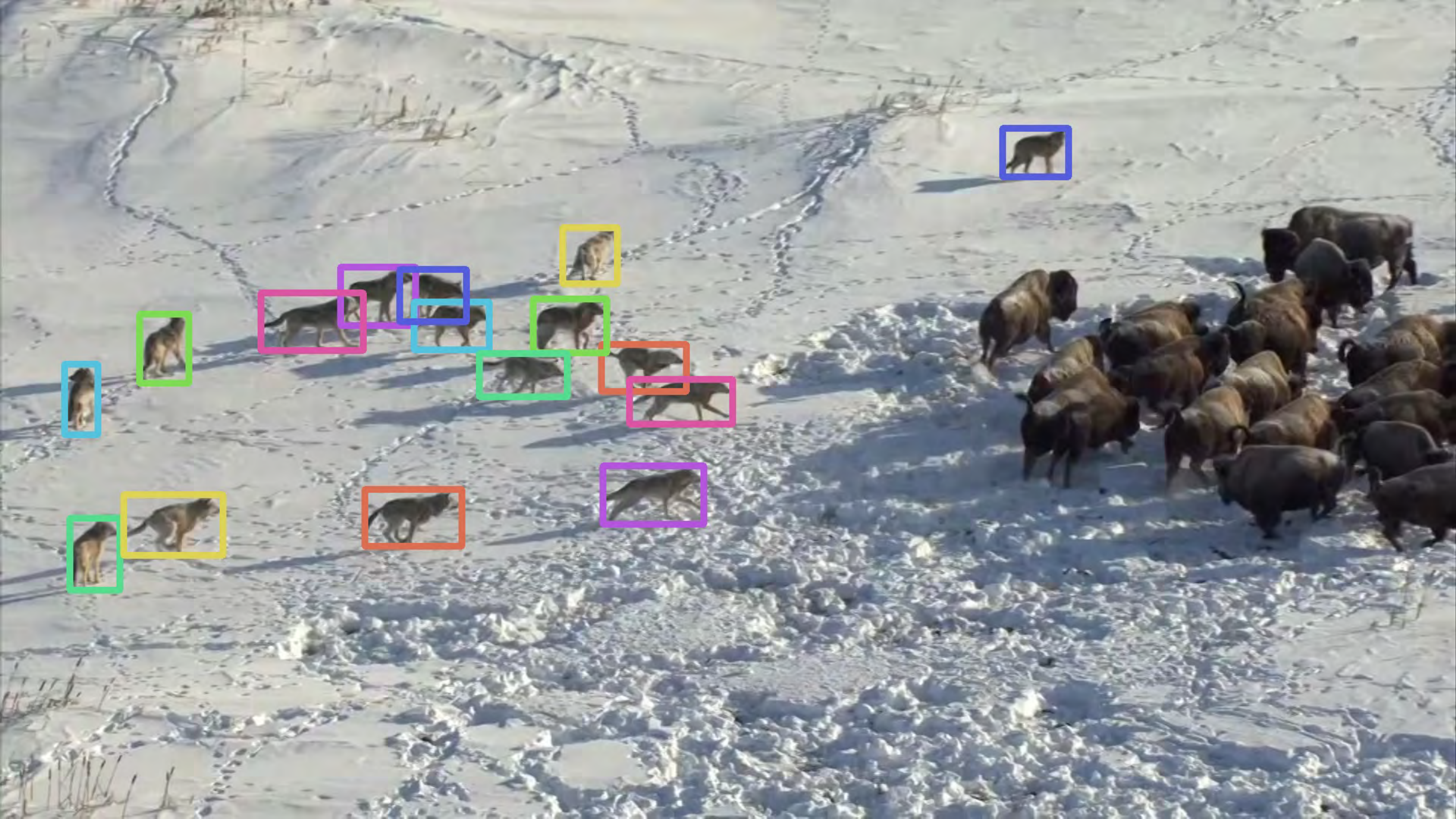}    &
		\includegraphics[width=0.195\linewidth,height=0.12\linewidth]{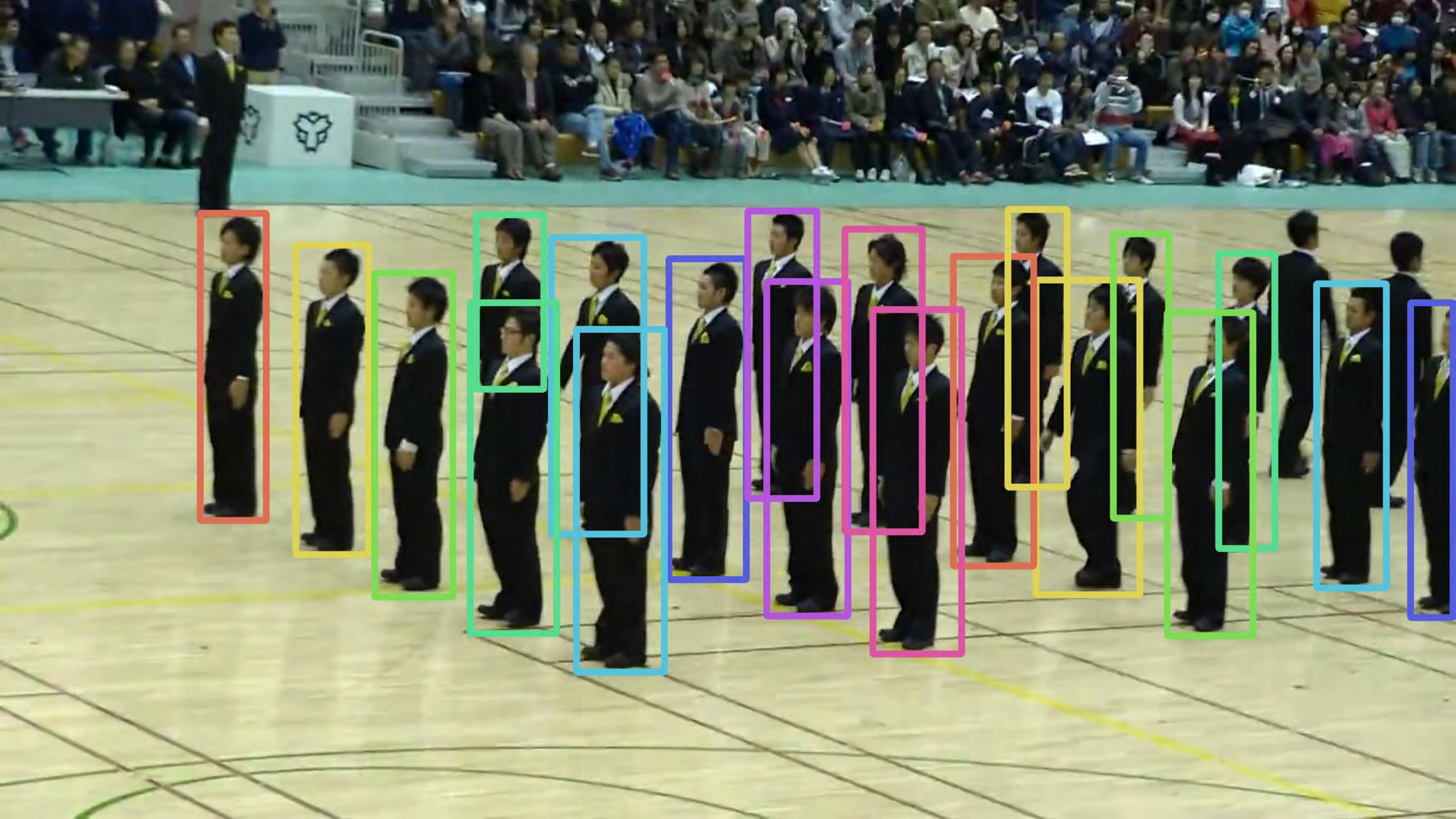}  \\
		\rule{0pt}{0.75ex}\\
		car & fish & insect & livestock & person\\
		
	\end{tabular}
	\endgroup
	\vspace{2mm}
	\caption{ Samples from each category of GMOT-40.}
	\label{fig:classes}
\end{figure*}

In this section, we will present the GMOT-40 dataset and the associated evaluation protocol.
As described in the related work, a serious GMOT dataset/benchmark is in great need for advancing the study of GMOT. By investigating the data issues in previous papers and borrowing ideas from recently popularized tracking benchmarks, we aim to construct a high-quality dataset in the following aspects:
\begin{itemize}
    \item \textit{Diversity in target category.} To address the generalization concern in previous MOT studies, GMOT-40 is designed to contain 40 sequences from 10 different categories, which is larger than most of previously studied datasets (typically less than 3 categories). The four sequences in each category are designed with further diversity. For example, the ``person" category in GMOT-40 covers both normal ``person" as in PASCAL-VOC~\cite{everingham2010pascal} and an unseen type ``wingsuit"; the ``insect" category covers ``ant" and ``bee", both of which are unseen in MS-COCO~\cite{lin2014microsoft} or PASCAL-VOCC~\cite{everingham2010pascal}. Some sample frames in GMOT-40 are shown in Figure~\ref{fig:classes}. 
    \item \textit{Real world challenges.} During sequence selection, we pay special attention to include sequences with various real-world challenges such as occlusion, target enter/exiting, fast motion, blur, \etc\ Moreover, the target density ranges from 3 to 100 targets per frame, with the average around 26. All these properties make GMOT-40 cover a wide range of scenarios. 
    \item \textit{High-quality annotation.} For high quality annotation, each frame in the sequence should be annotated by hand to ensure precise annotation. Besides, the initial annotation will be followed by careful validation and revision.
\end{itemize}
It is worth noting that, while more sequences would likely further improve the data usability, the additional non-trivial efforts in manual annotation may postpone the timely release of the dataset. In fact, as shown in Table~\ref{tab: model-free trackers}, GMOT-40 brings comprehensive improvements over previously used GMOT data, and is thus expected to facilitate the GMOT research in the future.
\subsection{Data Collection}
With the guidance mentioned above, we start by deciding 10 categories of objects that are highly possible to be dense and crowded. When selecting video sequences, we request that at least $80\%$ of the frames in a sequence to have more than 10 targets. Most targets of same category have similar appearance, while part of them differs on appearance, which is more close to reality. The minimum length of the sequence is set to 100 frames. 

After classes and requirements are determined, we started searching the YouTube with possible candidate videos. About 1000 sequences are initially picked as candidates. After scrutiny, we select 40 sequences out of them for better quality and more challenging task. Yet it does not mean that these 40 sequences are ready for annotation. Some of the sequences contain a large part that is irrelevant to our task. For example, in ``balloon" category, there are starting and ending sections focusing on the stage or the crowd of the celebration in the festival, which should be removed. In such a way, we carefully edit the video and select the best clips with a minimum of 100 frames.

Finally, GMOT-40 contains 50.65 trajectories per sequence on average. The whole dataset includes 9,643 frames in total, and each sequence has an average length of 240 frames. $85.28\%$ of the frames have more than 10 targets. The FPS ranges from 24 to 30 while resolution ranges from 480p to 1080p. 

The statistics of GMOT-40 in comparison with other densely annotated data used in GMOT studies are summarized in Table~\ref{tab: model-free trackers}. Note that we use the category definition of GMOT-40 here, since categories in other benchmarks are not general enough. As an example, both ``sky diving" and ``basketball" classes in \cite{Ionreid2020} belong to the ``person" class of GMOT-40. 

\subsection{Annotation}
The annotation format follows that of MOT15~\cite{MOTChallenge2015} where the detailed description is in the Supplementary Material. The only difference is that there is no out-of-view value and hence all bounding box in the groundtruth file should be considered in evaluation protocols.

Furthermore, only targets in the same category are annotated. For example, only the wolf in the ``stock" category is annotated as shown in Figure~\ref{fig:classes} since the initial bounding box indicates that only the wolf is the object of interest. Besides, the targets in the same categories are treated indiscriminately such as the red and white balloons in Figure~\ref{fig:classes}. 


The most important parts for building a high-quality GMOT dataset are manual labeling, double-checking, and error-correction. To ensure this, a group of experts such as Ph.D. students are included in the annotation team. For each video, it is first sent to the labeler to decide the group of interest. Then an expert will review the target group to see whether it reaches our requirement. After approval by experts, the labeler will start working on the annotation. The completed annotation will again be sent to experts for review and possible revision. 
\subsection{Video Attributes}
\begin{figure}[t]
\begin{center}
\includegraphics[width=1\linewidth]{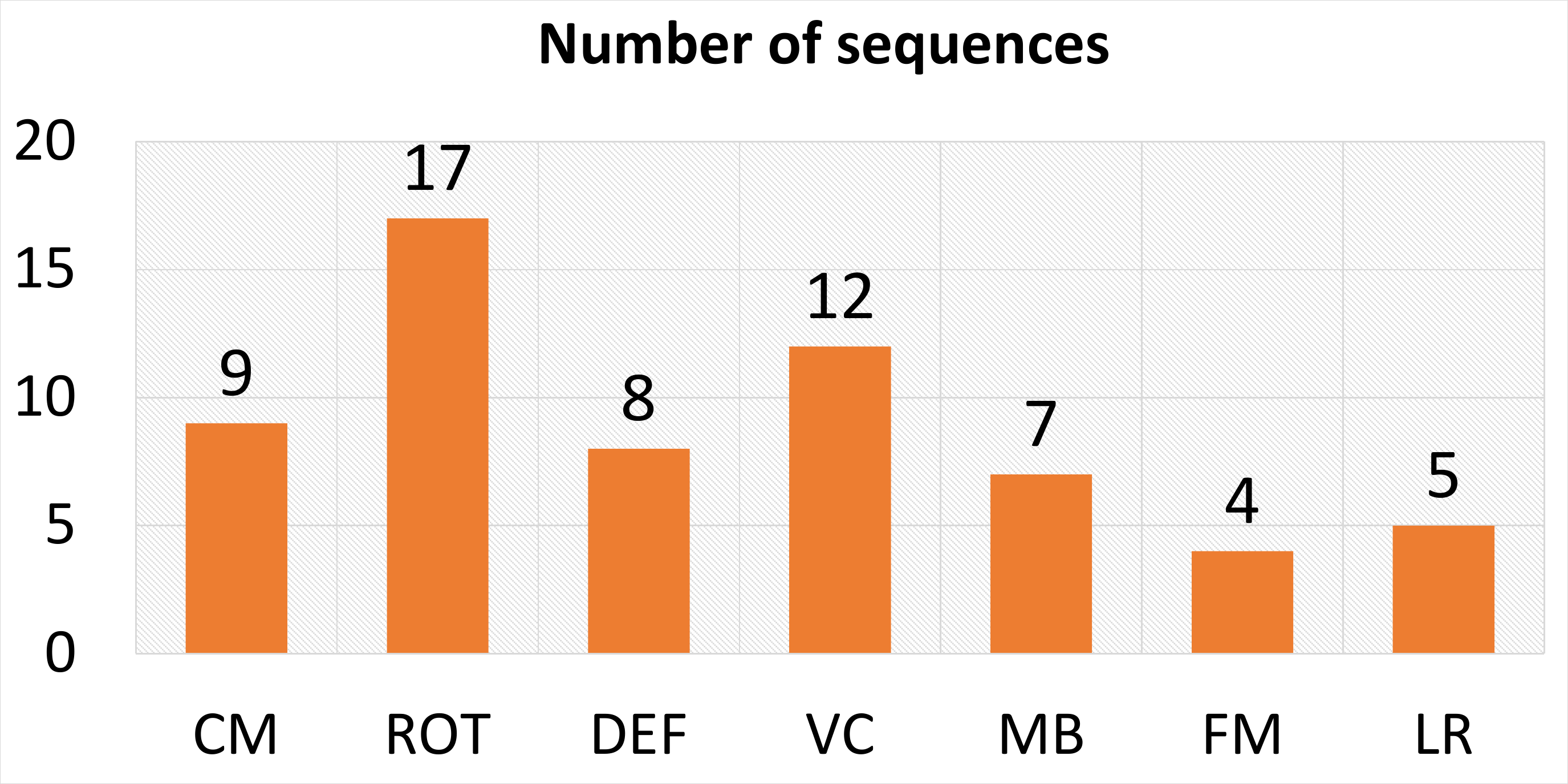}
\end{center}
\caption{Number of sequences for different attributes in our GMOT-40.}
\label{fig:Class_attr}
\end{figure}
As shown in the Figure~\ref{fig:classes}, diverse scenarios and hence more comprehensive attributes are included in GMOT-40 compared with other data used in previous GMOT papers. As an example, all of the ``person",  ``ball" and ``insect" classes have the properties of motion-blur and fast motion. Besides, the viewpoint significantly affects the appearance in ``boat" category. Furthermore, low resolution and camera motion appear in ``ball" and ``livestock" respectively. 

A detailed histogram on various attributes are presented in Figure~\ref{fig:Class_attr}. The abbreviation of attributes have the following meaning: \textbf{CM} -- camera motion; \textbf{ROT} -- target rotation; \textbf{DEF} -- target deforms in the tracking; \textbf{VC} -- significant viewpoint change that affects the appearance of target; \textbf{MB} -- target is blurred due to camera or target motion; \textbf{FM} -- fast motion of the targets with displacements larger than the bounding box; \textbf{LR} -- target bounding box is smaller than 1024 pixel for at least $30\%$ of the targets in the whole sequences. 

Although some of the attributes above are present in previous studies of GMOT~\cite{Ionreid2020,luo2013generic, luo2014bi, zhang2013preserving, zhu2016model}, yet GMOT-40 is the most comprehensive one, since it is collected from various natural scenes. These miscellaneous attributes of GMOT-40 can help the community to evaluate their trackers from multiple aspects. 

\section{GMOT Protocols and Tracking Baselines}
\subsection{Protocol}
Associated with the GMOT-40 dataset, we design a dedicated one-shot evaluation protocol for GMOT, adapting the settings from previous works such  as in~\cite{luo2014bi}. To facilitate the developing of GMOT trackers, an ablation study is also implemented to evaluate the association ability of a tracker.


The protocol aims to comprehensively evaluate the GMOT trackers in real-world application settings. As claimed in~\cite{luo2014bi}, a practical generic tracker is model-free thus is able to track multiple generic objects knowing only one template of targets. By adopting this Protocol, only one bounding box in the first frame of each video is provided to indicate the objects of interest. Trackers are supposed to use the object in that bounding box as a template and leverage the information of that object to detect and track all the targets in the video of same category.
All sequences in GMOT-40 are used to test the tracker for their performance on unseen category for the one-shot GMOT protocol. For comparison, we also design several new baselines (see Section~\ref{basline}) to generate the public detection for the whole sequence, using the only one sample given in the first frame. Trackers can be trained at any other benchmarks except GMOT-40. 

To choose the initial target of one sequence, we randomly sample some targets in the first frame that are not occluded. Then we carefully pick the best one out of them by hand to ensure it is representative and robust as the one-shot sample.


\subsection{Baselines for One-shot GMOT} \label{basline}
For one-shot GMOT protocol, we propose a series of two-stage baselines by adapting existing tracking algorithms. Each baseline consists of a one-shot detection stage, which obtains detection results for all frames in sequence, and a target association stage, which associates detected targets and gets the final tracking results. 

\subsubsection{One-Shot Detection Stage}
In our implementation, we adopt a recently proposed SOT method, GlobalTrack~\cite{Huang_Zhao_Huang_2020}, to create a one-shot detection method. GlobalTrack searches the whole image in following frames (search frames) while most SOT trackers only search a predefined neighborhood of the target position in the previous frame. The model is pretrained on other datasets~\cite{lin2014microsoft,huang2019got,fan2020lasot}.
We then split the modified model to two modules, a target-guided region proposal module, and a target-guided matching module. The target-guided region proposal module extracts features for the labeled target on the initial frame, and return regions that may contain targets on the search frame. Then target-guided matching module extracts features from these regions, computes similarity scores between these potential targets, and produces multiple search results with the refined position. Furthermore, those targets with similarity scores lower than the threshold (0.1) are filtered out.

In the one-shot detection process, the initial frame is always the first frame and the search frames include all frames in the sequence, including the first frame itself. The detection process is repeated to get results for all these frames. The whole process is shown in Algorithm~\ref{alg-gt}.

\subsubsection{Target Association Stage}
With these detection results, we now transform the one-shot GMOT task to a traditional MOT task with public detection. Most existing MOT algorithms can be adapted here to get association. The MOT algorithms used in evaluation are stated in Section~\ref{eval_tracker}.

Combining the one-shot detection method with different target association methods, we get a series of baselines for the one-shot GMOT task. We evaluate their tracking performances comprehensively in Section~\ref{p2_eval}.

\SetKwInput{KwModel}{Model}
\begin{algorithm}[t]
	\LinesNumbered
	\KwData{\\
		~~$\{I_1,...,I_m\}$: images in a sequence; \\
		~~$x_{gt}$: initial detection (groundtruth box) in $I_1$;\\
		~~$s_{th}$: threshold for detection similarity score.}
	\KwModel{\\
		~~$\phi_R$: target-guided region proposal module; \\
		~~$\phi_M$: target-guided matching module.}
	\KwOut{\\
		~~$\{x^k_i\}_{i=1}^{n_k}$: $n_k$ detected targets for $I_k$, $1\leq k \leq m$. 
	}
	\vspace{5pt}
	Extract features for the initial target\;
	\qquad $F_{gt}=\phi_R(I_1,x_{gt})$\;
	\For{$k=1,...,m$}{
		Use $F_{gt}$, $\phi_R$ to produce ${r_k}$ regions $R$ that may contain targets on image $I_k$\;
		\vspace{5pt}
		\qquad $R=\{x_1^k,...,x_{r_k}^k\}=\phi_R(F_{gt},I_k)$\;
		\vspace{5pt}
		Use $\phi_M$ to extract features $F_R$ from $R$\;
		\vspace{5pt}
		\qquad $F_R=\{f_1^k,...,f_{r_k}^k\}=\phi_M(R)$\;
		\vspace{5pt}
		Compute similarity scores $S$ between $F_R$ and $F_{gt}$, and produce targets $T$ with refined positions\;
		\vspace{5pt}
		\qquad $S=\{s_1^k,...,s_{r_k}^k\}=\phi_M(F_{gt},F_R)$\;
		\qquad $T=\{\tilde{x}_1^k,...,\tilde{x}_{r_k}^k\}=\phi_M(F_{gt},F_R)$\;				\vspace{5pt}
		~~Filter $T$ by comparing $S$ with $s_{th}$, and then get the final $n_k$ targets $T^k$\;
		\vspace{5pt}
		\qquad $T^k=\{x_1^k,...,x_{n_k}^k\}=C(T,S,s_{th})$\;
		\vspace{5pt}
		~~~~where $C$ denotes the comparison process\;
	}
	\caption{One-shot Detection Process.}
	\label{alg-gt}
\end{algorithm}

\section{Experiment}

\begin{figure*}[!h]
 	\small
	\centering
	\begingroup
	\tabcolsep=0.15mm
    \def\arraystretch{0.08}
	\begin{tabular}{p{10pt} ccccc}%
		  & MDP \cite{xiang2015learning} 
		  & Deep SORT \cite{wojke2017simple} 
		  & IOU tracker \cite{bochinski2017high} 
		  & FAMNet \cite{chu2019famnet} \\
		\\
		\\
		\\
		\\
		\\
		\\
		\\
		\rotatebox[y=30pt]{90}{Bird}  & 
		\includegraphics[width=0.24\linewidth,height=0.15\linewidth]{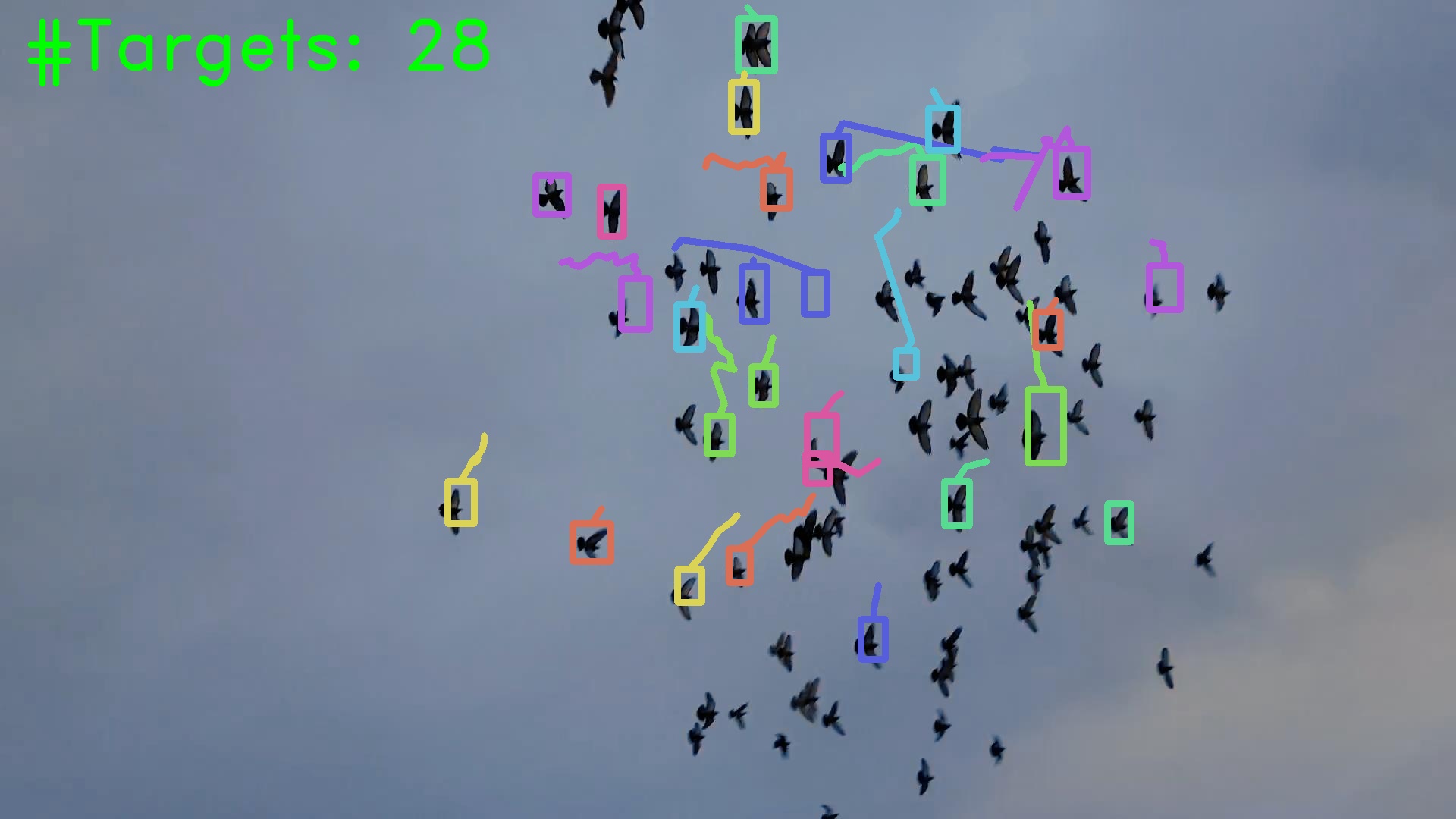}    & 
		\includegraphics[width=0.24\linewidth,height=0.15\linewidth]{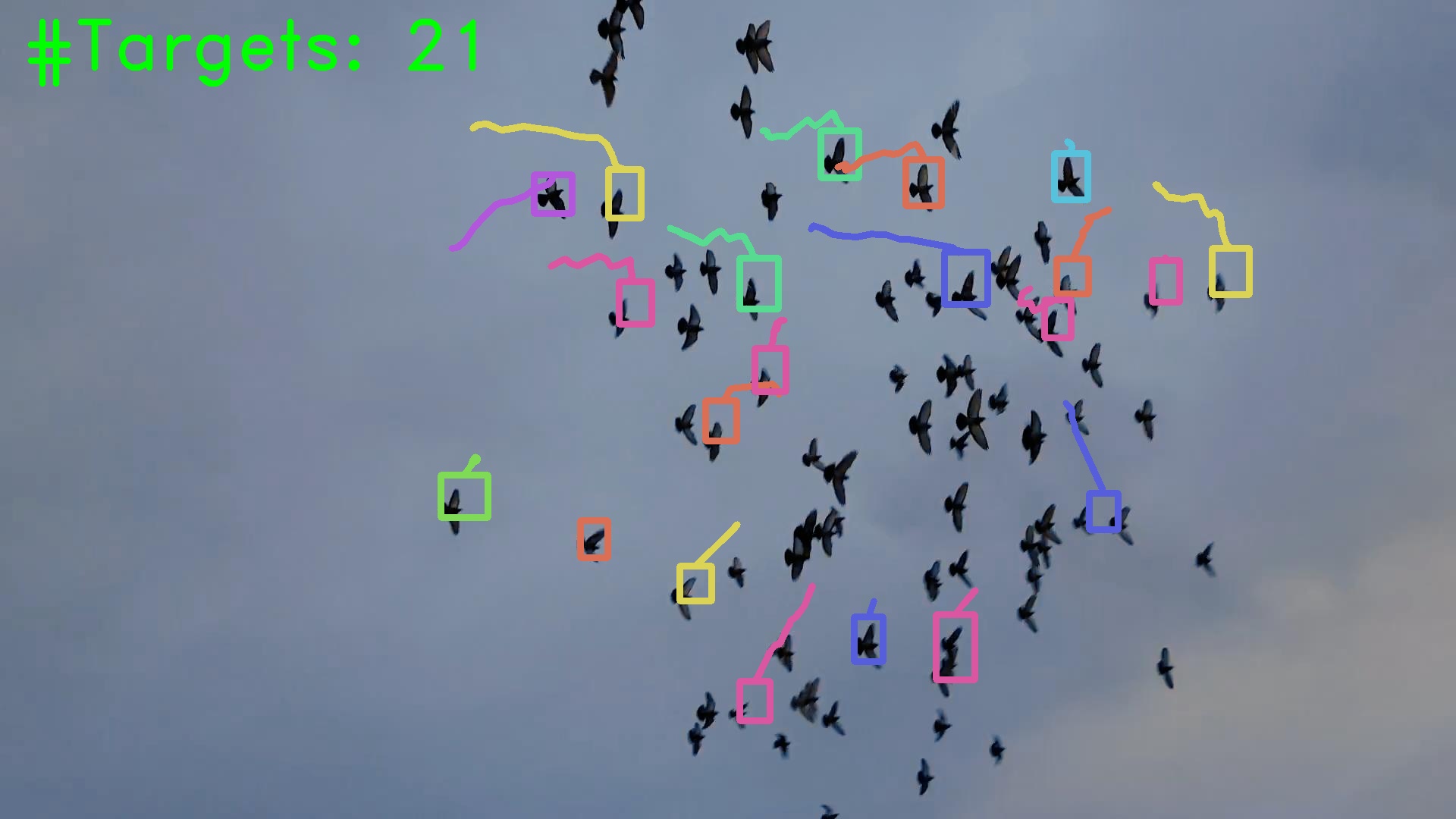}    & 
		\includegraphics[width=0.24\linewidth,height=0.15\linewidth]{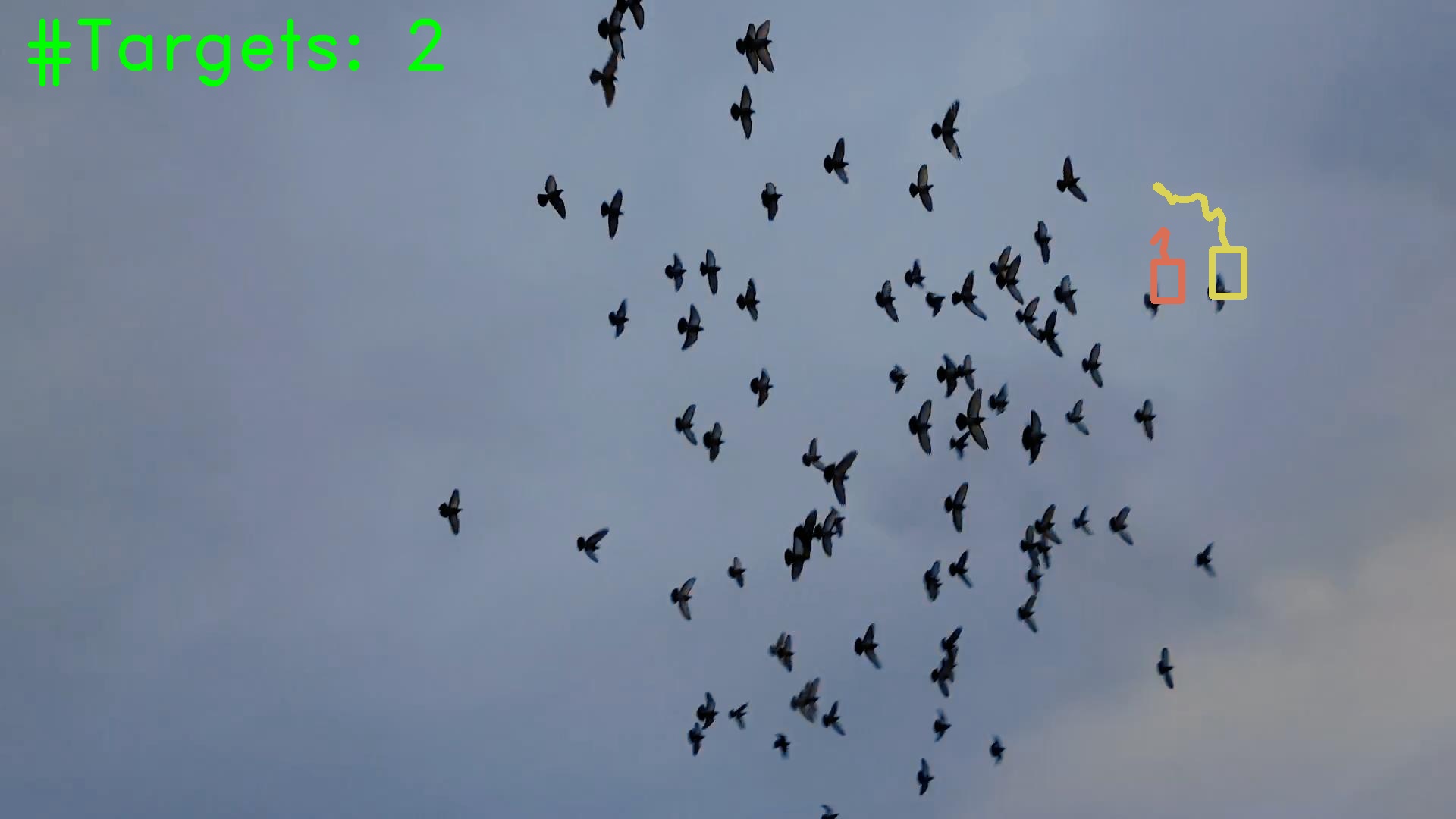}    &
		\includegraphics[width=0.24\linewidth,height=0.15\linewidth]{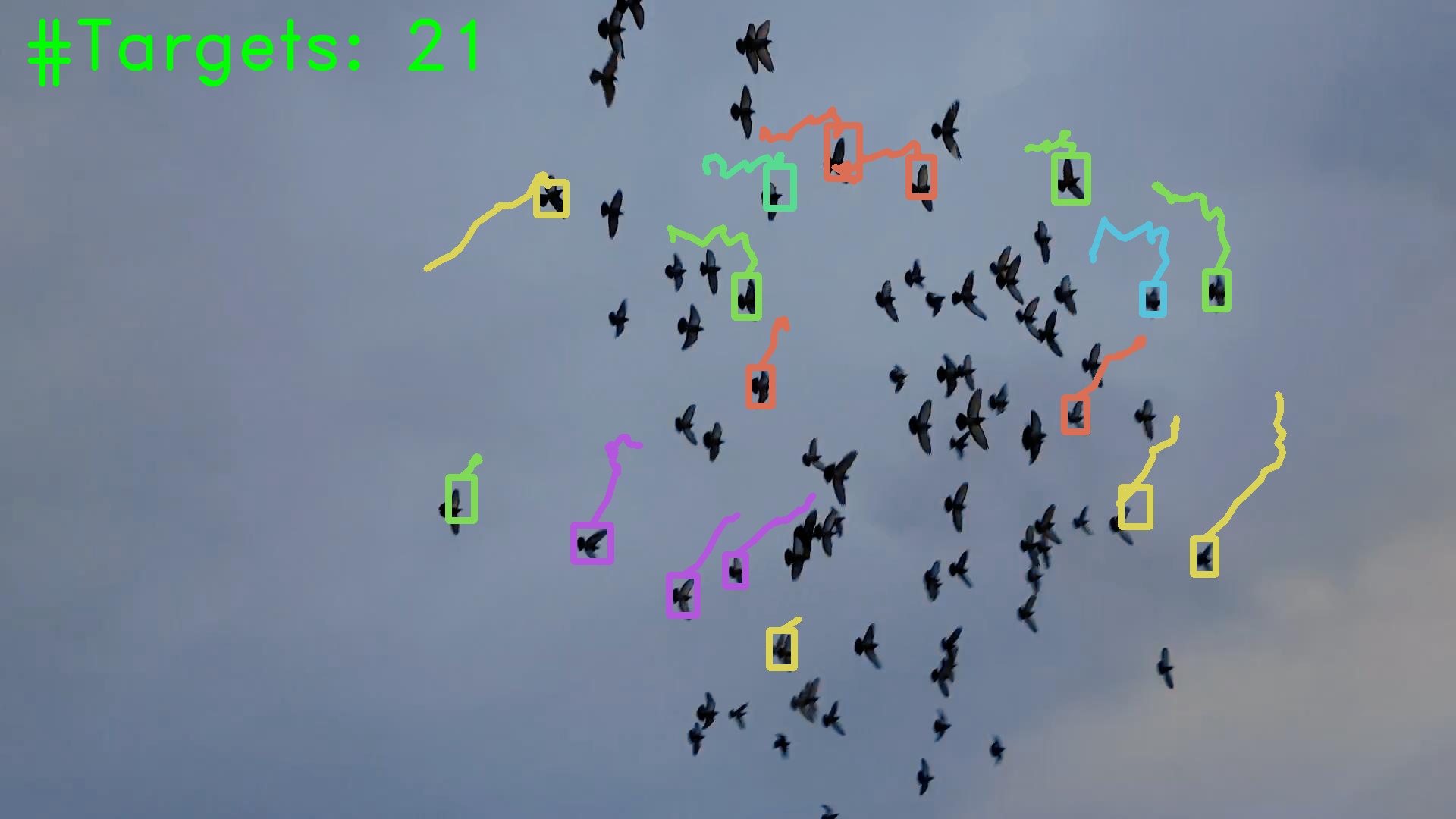}    \\
        \\
        \vspace{.1mm}\\
		\rotatebox[y=30pt]{90}{Fish}  & 
		\includegraphics[width=0.24\linewidth,height=0.15\linewidth]{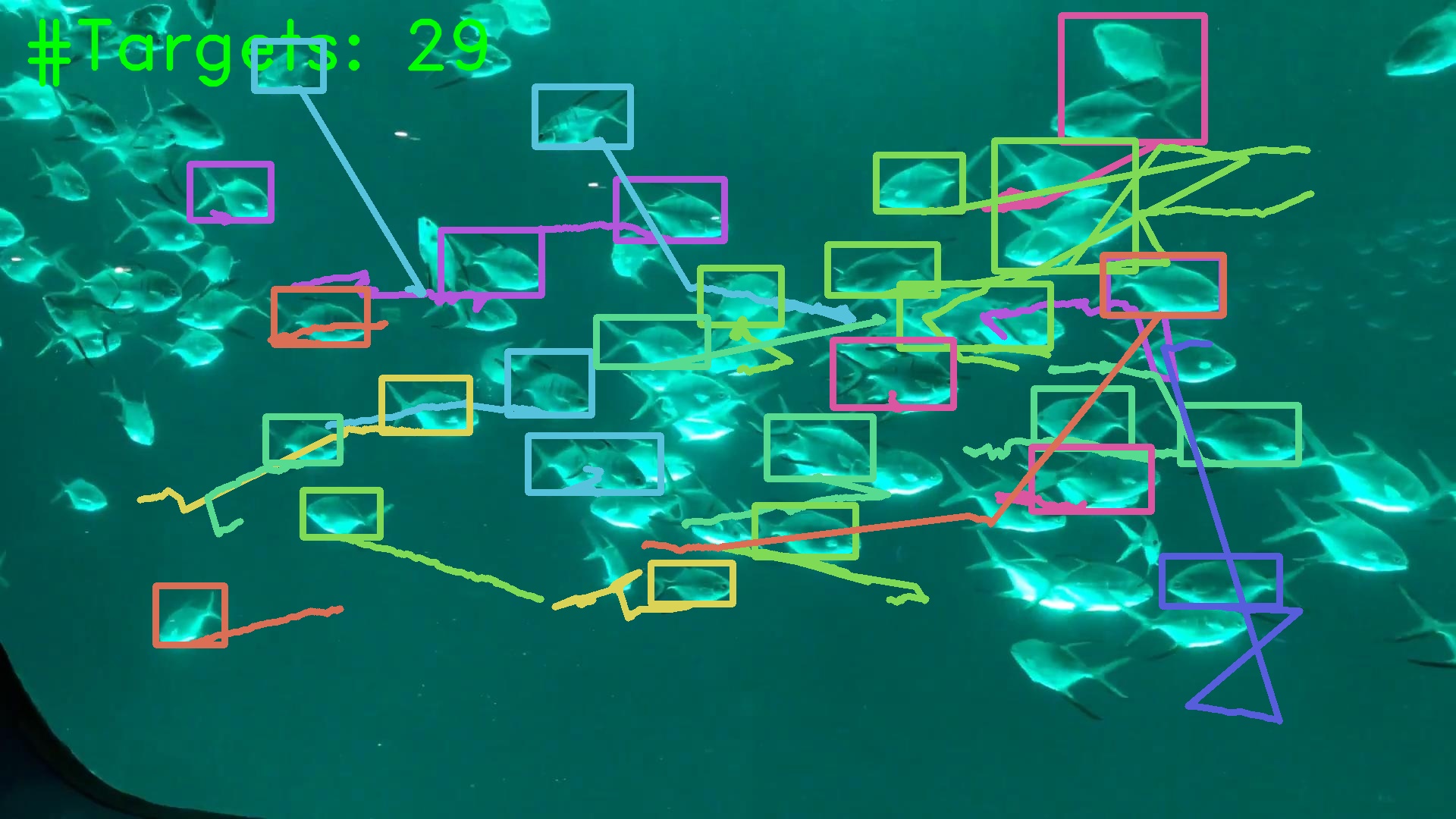}    & 
		\includegraphics[width=0.24\linewidth,height=0.15\linewidth]{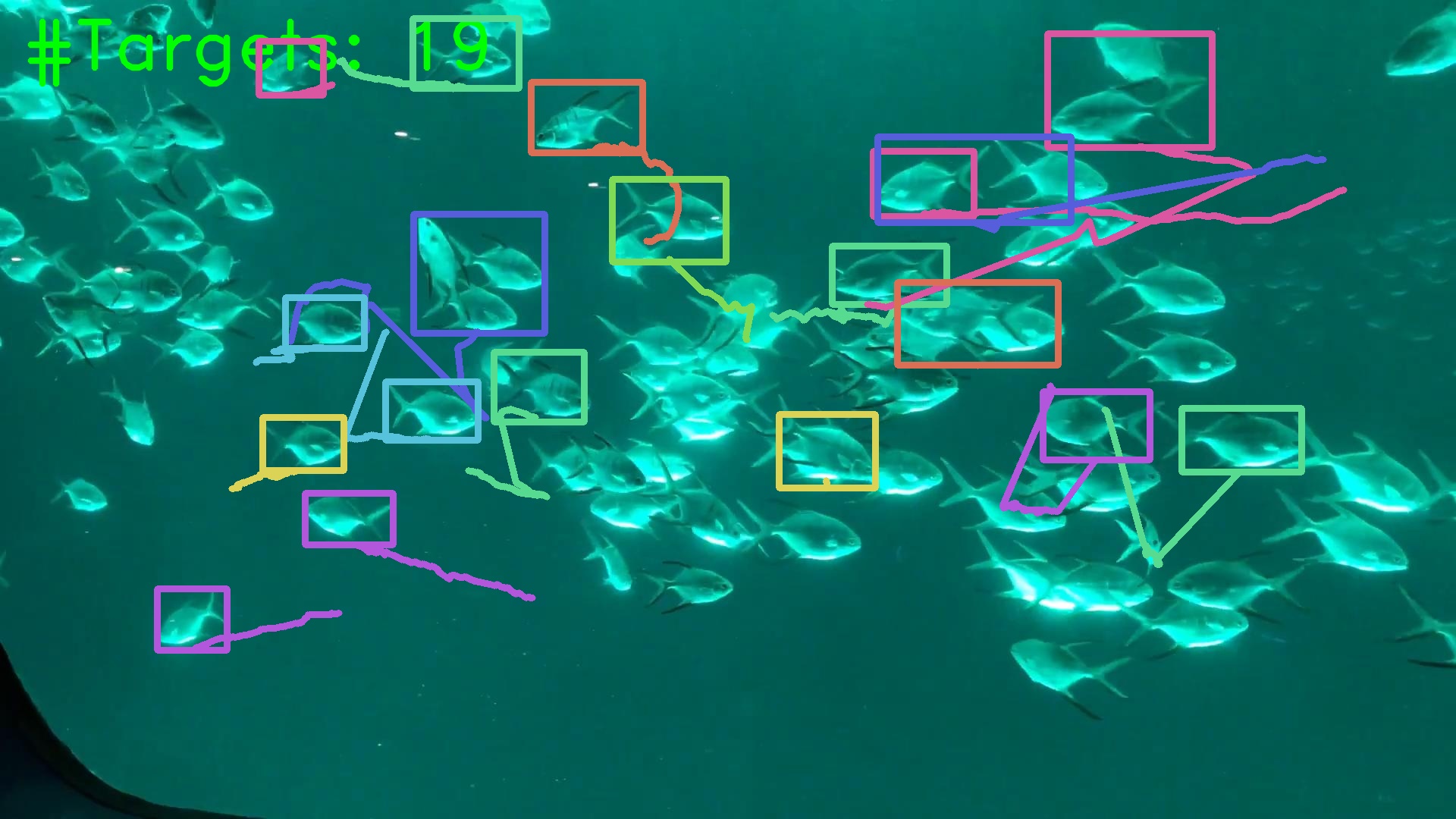}    & 
		\includegraphics[width=0.24\linewidth,height=0.15\linewidth]{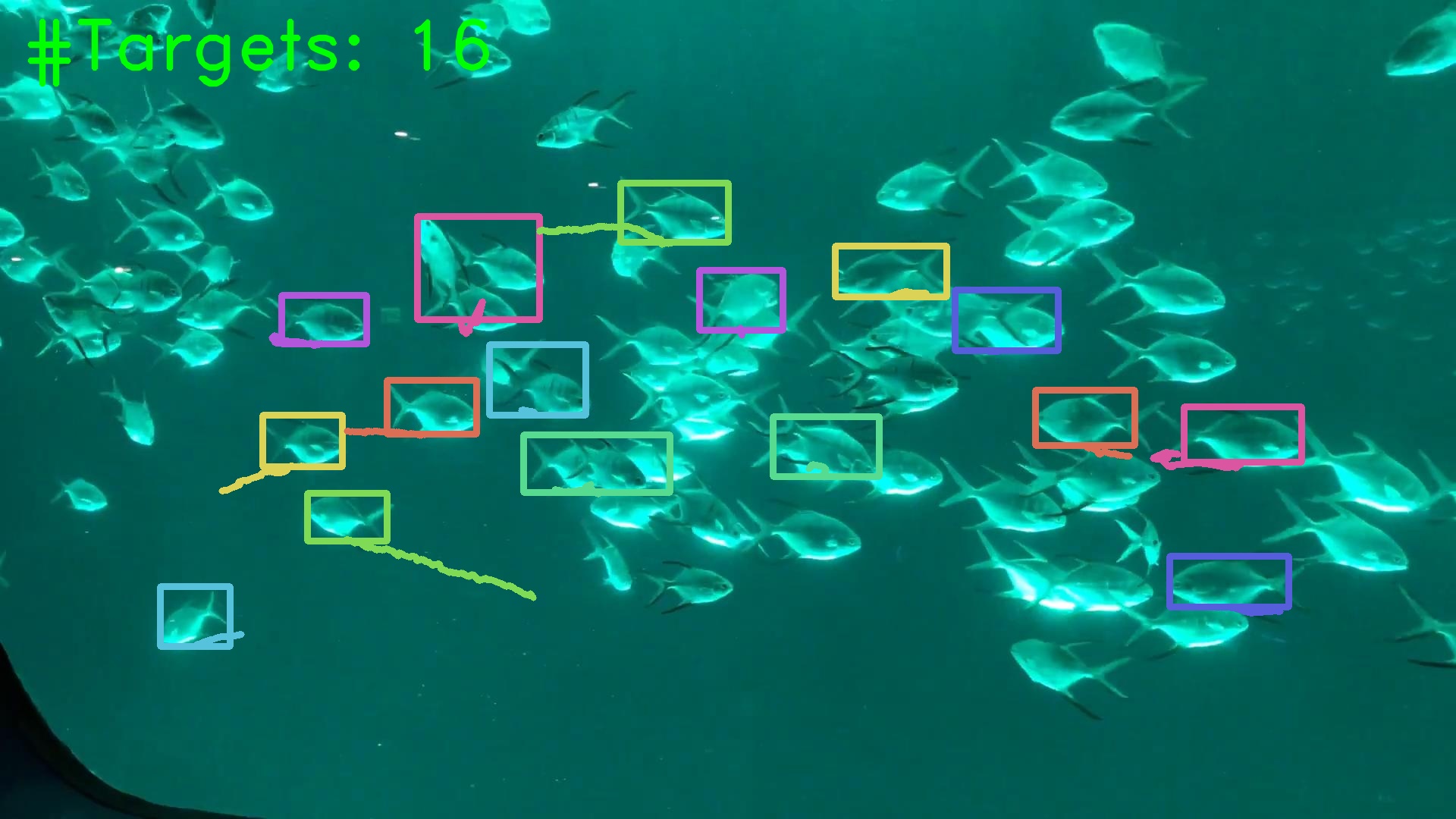}    &
		\includegraphics[width=0.24\linewidth,height=0.15\linewidth]{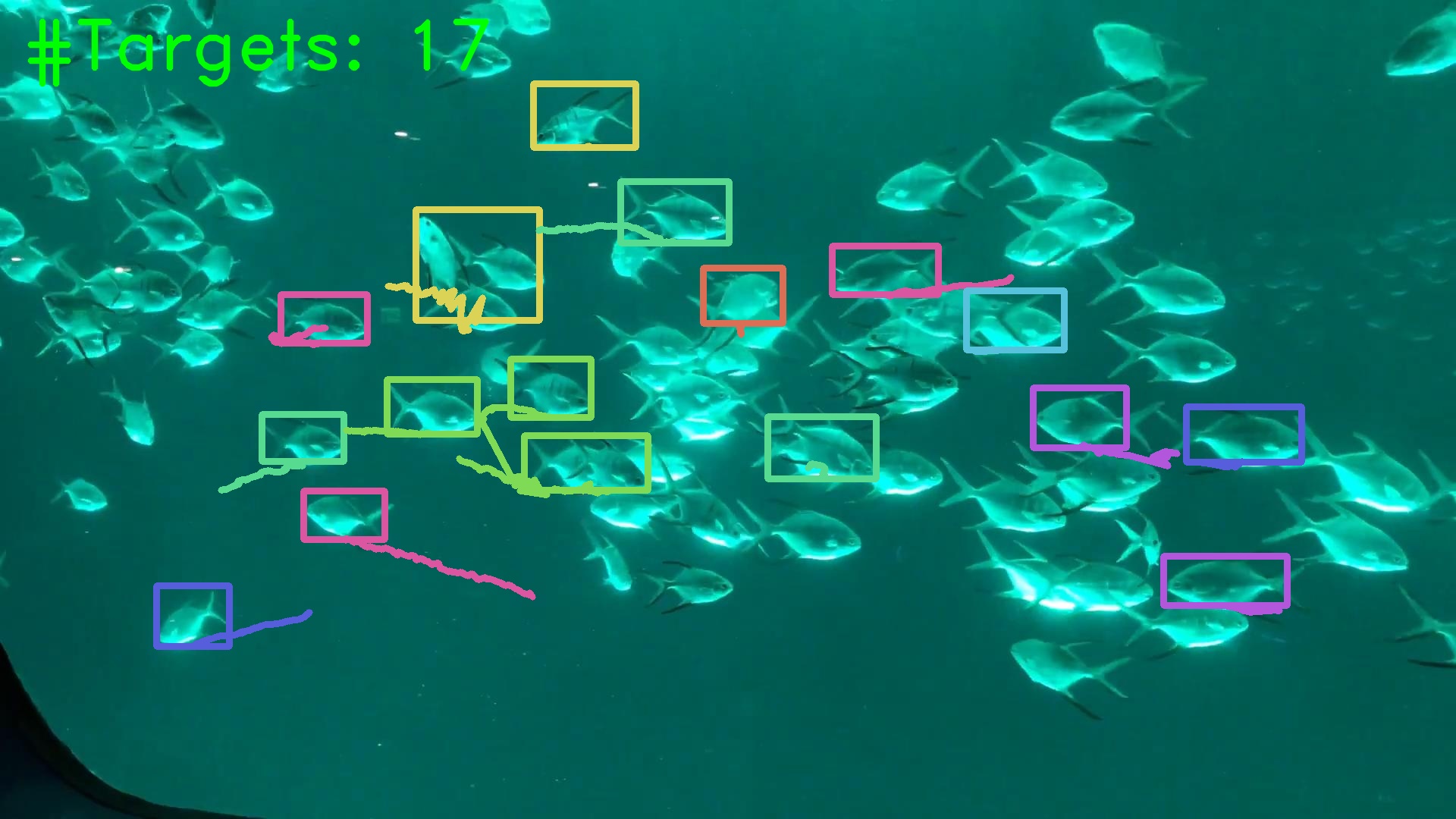}    \\
        \\
        \vspace{.1mm}\\
		\rotatebox[y=30pt]{90}{Ball}  & 
		\includegraphics[width=0.24\linewidth,height=0.15\linewidth]{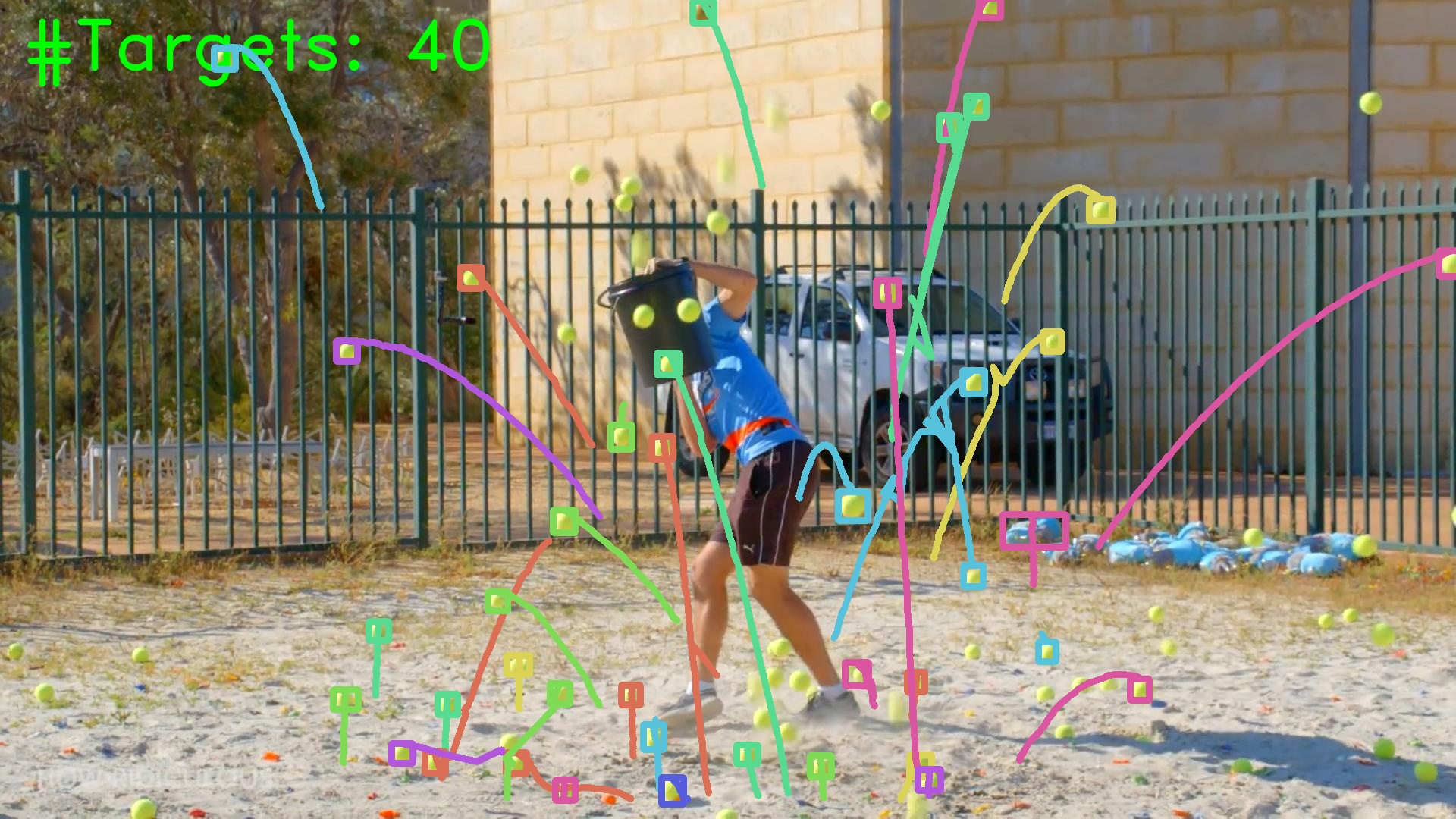}    & 
		\includegraphics[width=0.24\linewidth,height=0.15\linewidth]{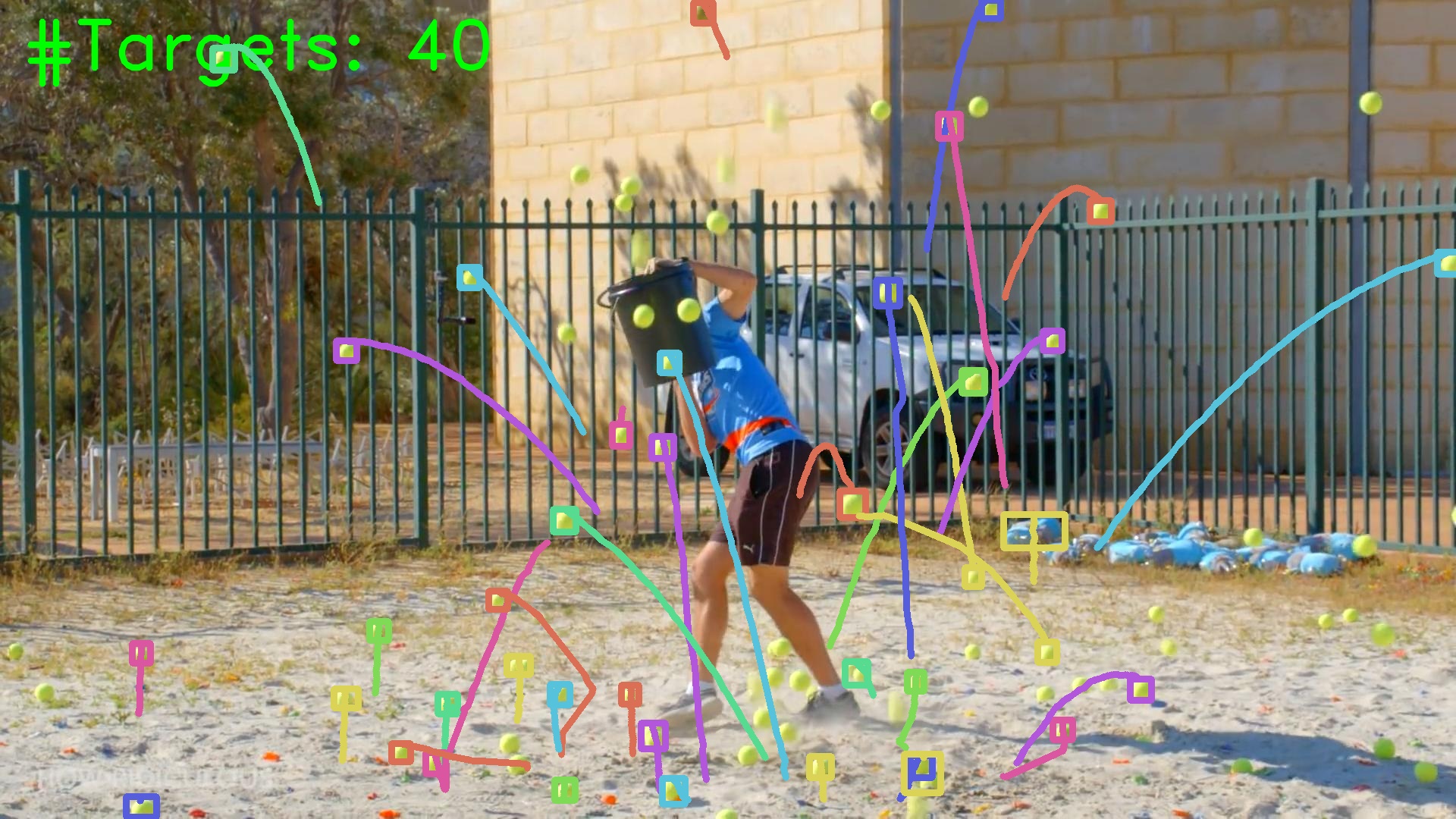}    & 
		\includegraphics[width=0.24\linewidth,height=0.15\linewidth]{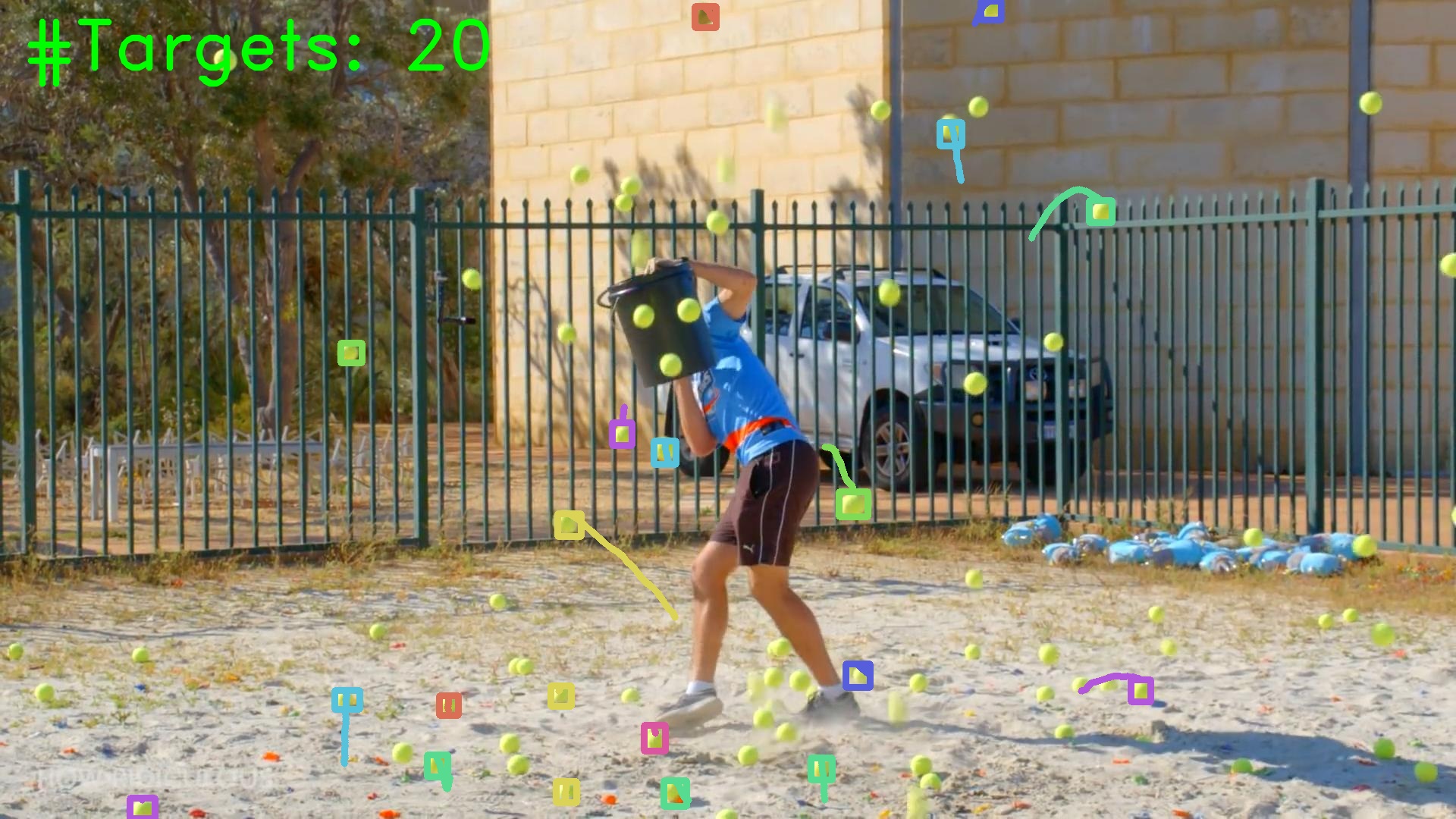}    &
		\includegraphics[width=0.24\linewidth,height=0.15\linewidth]{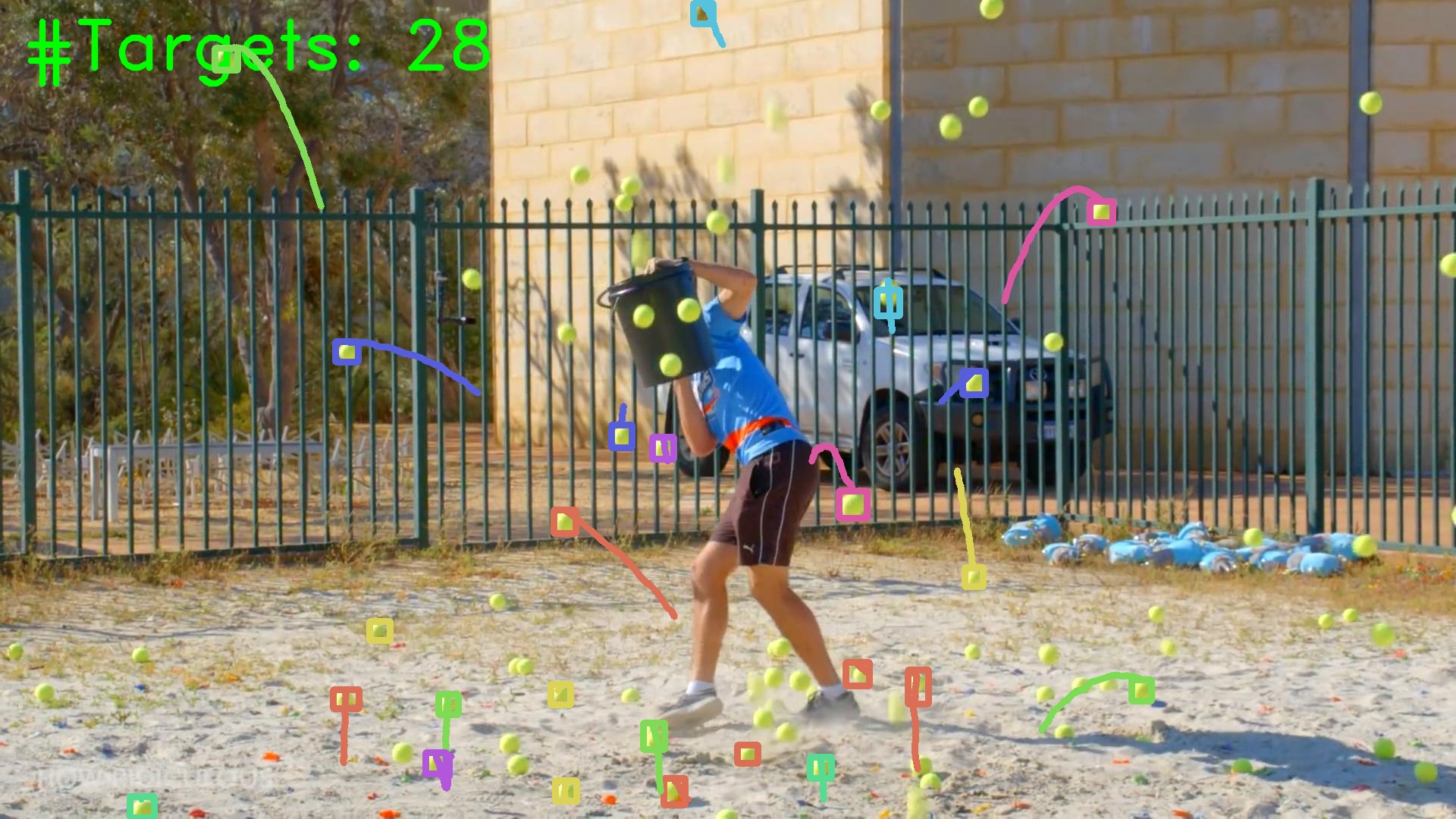}    \\
	\end{tabular}
	\endgroup
	\vspace{2mm}
	\caption{Results visualization of four trackers on sequences.}
	\label{fig:result}
\end{figure*}

\subsection{Evaluation Metrics}
A group of metrics on MOT has been proposed to fairly compare the tracker and reveal the performance. Among them the most widely used ones are CLEAR MOT metrics \cite{bernardin2008evaluating} and ID metrics \cite{ristani2016performance}. The former stresses the number of incorrect predictions while the latter focus on the longest time of following targets. Combining them will provide a comprehensive evaluation of the performance in GMOT-40.  
\subsection{Evaluated Trackers}\label{eval_tracker}
We focus on the trackers that are built on public detection and have publicly available code. Both classical and more recent trackers are included to provide a comprehensive review. Among them, there are FAMNet~\cite{chu2019famnet}, Deep SORT~\cite{wojke2017simple}, MDP~\cite{xiang2015learning}, IOU tracker~\cite{bochinski2017high}. 

\begin{table*}[t]
\resizebox{\textwidth}{!}{%
\begin{tabular}{l|ccccccccccccc}
\hline\thickhline

            & MOTA   & IDF1    & IDP     & IDR     & Rcll    & Prcn    & MT  & PT  & ML$\downarrow$   & FP$\downarrow$    & FN$\downarrow$   & IDs$\downarrow$  & FM$\downarrow$      \\
\hline
MDP \cite{xiang2015learning} & \textbf{19.80\%} & \textbf{31.30\%} & 61.80\% & \textbf{21.00\%} & \textbf{27.20\%} & 80.20\% & 142 & \textbf{621} & \textbf{1161} & 17260 & \textbf{186580} & 1779 & 2748\\
DeepSORT \cite{wojke2017simple} & 14.50\% & 24.40\% & \textbf{67.50\%} & 14.90\% & 18.50\% & \textbf{84.10\%} & 72 & 509 & 1363 & 9000 & 208818 & 1315 & 2233\\
IOU \cite{bochinski2017high} & 11.80\% & 20.30\% & 64.60\% & 12.00\% & 15.40\% & 82.60\% & 56  & 397 & 1491 & \textbf{8299} & 216921 & \textbf{754} & \textbf{1668}\\
FAMNet \cite{chu2019famnet} & 18.00\% & 28.30\% & 54.80\% & 19.10\% & 26.80\% & 76.80\% & \textbf{166} & 581 & 1197 & 20741 & 187730 & 1660 & 1878\\
\hline\thickhline
\end{tabular}%
}
\caption{Comparison of trackers with one-shot GMOT protocol.}
\label{tab: CompP2}
\end{table*}

\begin{table}[]
\centering
\small
\begin{tabular}{p{55px}|p{40px}|p{40px}|p{40px}}
\hline\thickhline
Methods     &   MOTA            &   MOTP            &   IDF1            \\
\hline
MDP\cite{xiang2015learning}         &$19.92\%\pm1.84\%$&$24.16\%\pm0.27\%$&$31.84\%\pm2.23\% $ \\
DeepSORT\cite{wojke2017simple}    &$14.98\%\pm1.47\%$&$23.66\%\pm0.53\%$&$25.38\%\pm2.32\%$  \\
IOU\cite{bochinski2017high}  &$12.36\%\pm1.60\%$&$25.34\%\pm0.36\%$&$20.90\%\pm2.73\%$  \\
FAMNet\cite{chu2019famnet}
&$17.60\%\pm0.85\%$&$22.56\%\pm0.23\%$&$27.76\%\pm1.16\%$  \\
\hline\thickhline
\end{tabular}
\caption{Average of five runs initiated by randomly picked one-shot templates.}
\label{randomfive}
\end{table}

\begin{table*}[t]
\resizebox{\textwidth}{!}{%
\begin{tabular}{l|ccccccccccccc}
\hline\thickhline
                            & MOTA   & IDF1    & IDP     & IDR     & Rcll    & Prcn     & MT   & PT  & ML$\downarrow$   & FP$\downarrow$    & FN$\downarrow$     & IDs$\downarrow$  & FM$\downarrow$       \\
\hline
MDP \cite{xiang2015learning}                & 75.00\%                & 72.50\% & 79.50\% & 66.70\% & 80.70\% & 96.20\% & 1105  & \textbf{703} & \textbf{136} & 8234 & 49448 & 4103 & 4758\\
DeepSORT \cite{wojke2017simple} & \textbf{80.60\%} & \textbf{79.30\%} & 85.30\% & \textbf{74.00\%} & \textbf{84.50\%} & 97.30\% & \textbf{1344} & 344 & 256 & 5944 & \textbf{39648} & 4074 & \textbf{2937} \\
IOU \cite{bochinski2017high} & 75.90\% & 79.00\% & 85.80\% & 73.20\% & 80.40\% & 94.20\% & 1237 & 260 & 447 & 12704 & 50232 & \textbf{1225} & 3767  \\
FAMNet \cite{chu2019famnet} & 67.40\% & 70.50\% & \textbf{86.30\%} & 60.50\% & 70.10\% & \textbf{97.60\%} & 1302 & 319 & 323 & \textbf{4505} & 76706 & 2454 & 6229\\
\hline\thickhline
\end{tabular}%
}
\caption{Comparison of trackers with the protocol in ablation study.}
\label{tab: Comp1.1}
\end{table*}

\subsection{Protocol Evaluation}\label{p2_eval}

We first evaluate the quality of the proposed target candidates that are generated by our baseline algorithm. Since in one-shot generic setting, the difference between categories is inconsequential. Thus we directly use AP (Average Precision) as our metric to report the ``detection'' solely performance. We have $AP_{50}$ of $15.65\%$ and $AP_{75}$ of $15.51\%$ while setting the IOU threshold at $0.5$ and $0.75$ respectively. Note that our baseline target candidate proposal is not trained on GMOT-40. In qualitative analysis, the baseline is found out to behave badly with deformation, rotation out-of-plane, motion blur and low resolution. The reason may be that the matching module of our modified GlobalTrack produced too many false negatives while ranking the confidence in the final stage. 

The detection results generated by our baseline algorithm serve as public detection in the following experiments. We test the trackers on all 40 sequences in its initial setting with the pre-trained model without any further modification.  The results as well as MOTA and IDF1 are listed in the Table~\ref{tab: CompP2}. With the inclusion of the one-shot detector, MDP becomes the best among them all. Yet its IDF1 is just $31.30\%$ and MOTA is just $19.80\%$ . Deep SORT and FAMNet here behave slightly worse than MDP with the IOU tracker after them. In other words, there is correlation between their processing of detection and their performance. A sample of results is presented Figure~\ref{fig:result} with each color standing for a different trajectory.

\begin{figure}[t]
\begin{center}
\includegraphics[width=1\linewidth,height=0.55\linewidth]{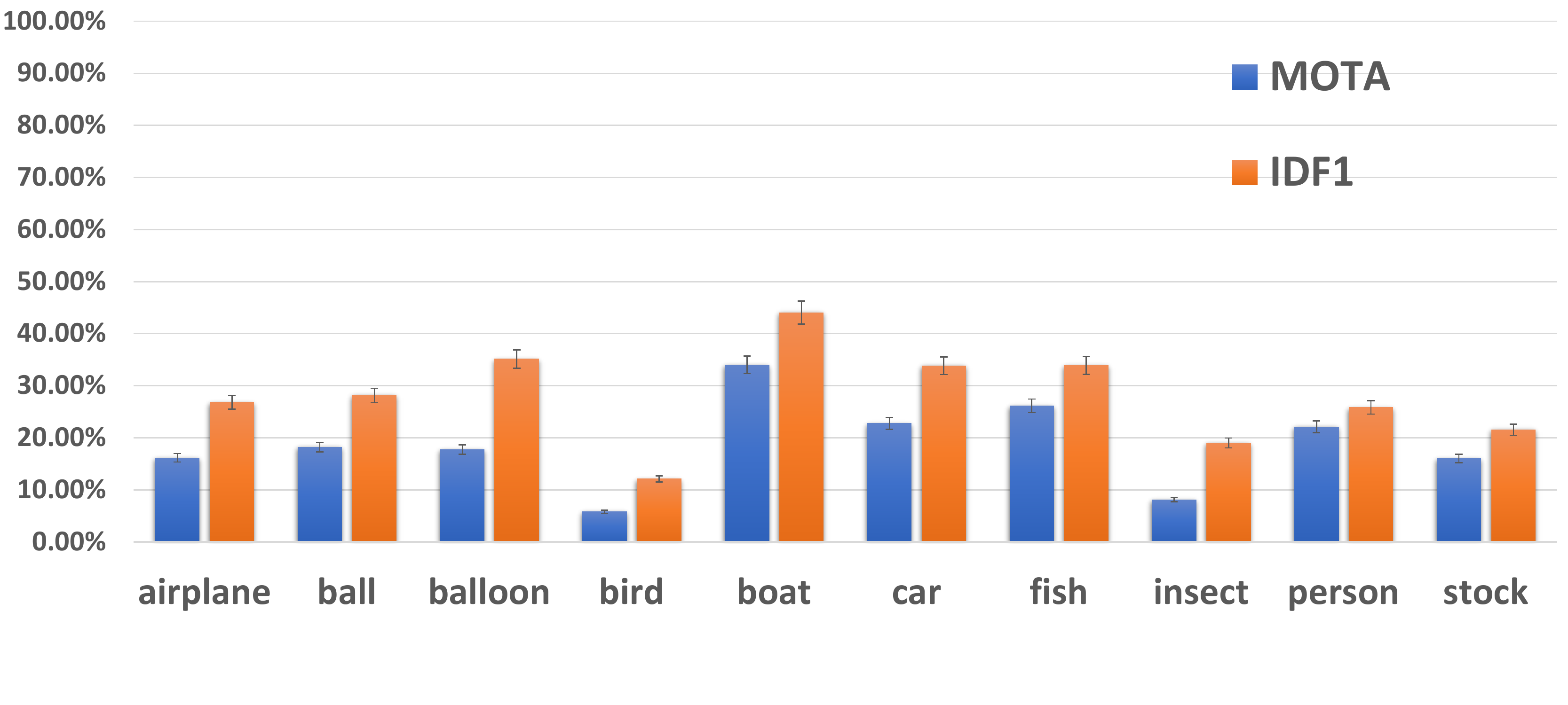}
\end{center}
\caption{Average scores of all trackers for different classes in one-shot GMOT Protocol.}
\label{fig:ClassComp2}
\end{figure}

Besides, we include Figure~\ref{fig:ClassComp2} to compare the performance in different classes. Each bar represents the mean of all 5 trackers. Specifically, the ``bird" and ``insect" classes poses a challenge for all the trackers. This again proves the necessity of diversity and hence the release of GMOT-40. A more detailed version is included in Supplementary Material. 

Finally, to make sure the results in experiment is unbiased from the initial results picked by user. We randomly sample the one target in the 1st frame for protocol and repeat this procedure for 5 times. Then we report the mean and standard deviation of the results over these 5 experiments. The results are shown in Table~\ref{randomfive}. As we can see, the fluctuations are very low, implying that the choice of the initial bounding box does not affect the result significantly. 

\begin{figure}[t]
\begin{center}
\includegraphics[width=1\linewidth,height=0.55\linewidth]{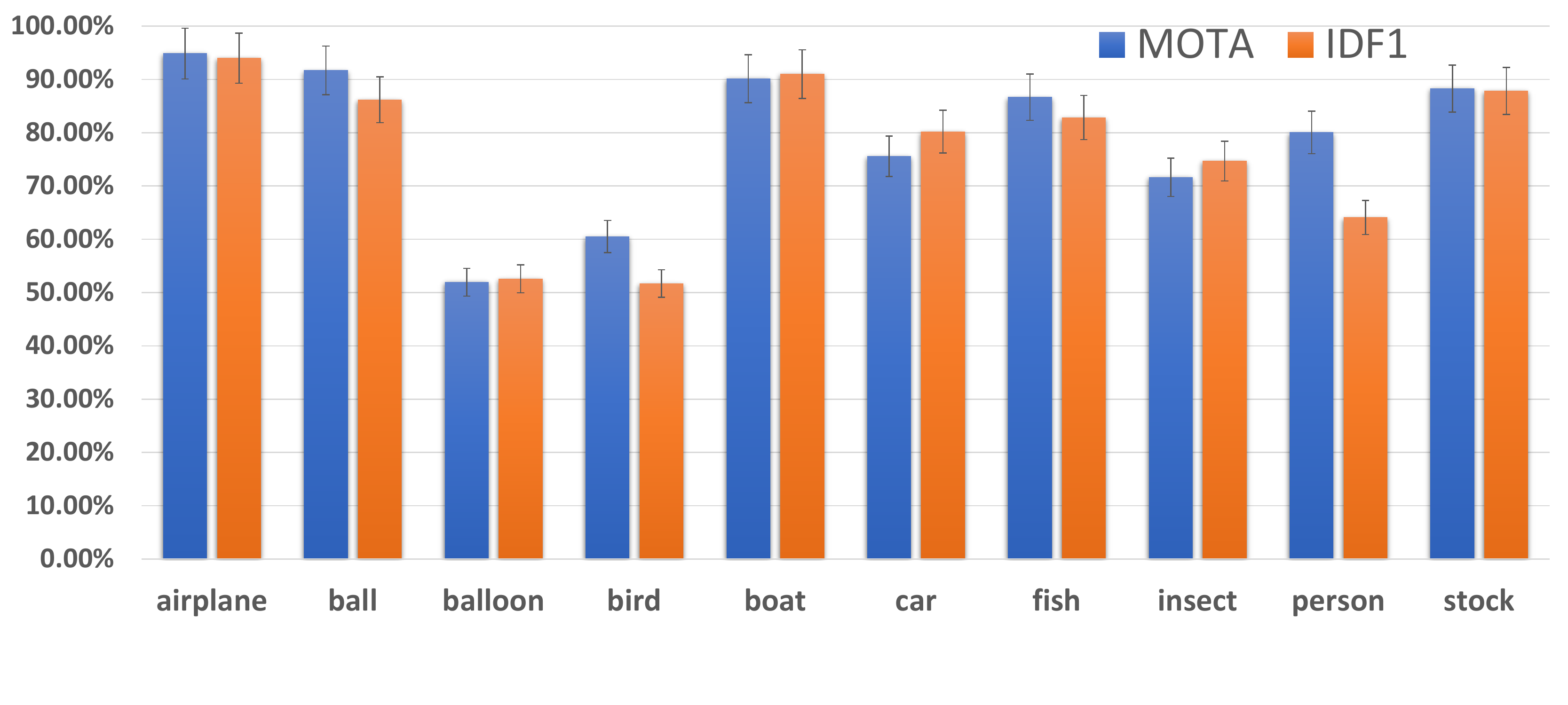}
\end{center}
\caption{Average scores of all trackers for different classes with the protocol in ablation study.}
\label{fig:ClassComp1.1}
\end{figure}

\subsection{Ablation Study}
In ablation study, the groundtruth detection are provided for the tracker while all other experiment conditions are the same. The result of this protocol is presented in Table~\ref{tab: Comp1.1}, where we can see nearly all trackers' performances improve significantly compared with Table~\ref{tab: CompP2}. Note that our benchmark contains many categories that are unseen for the tracker during their training. Hence the benchmark would favor the association based on Intersection Over Union (IOU) of targets across frames rather than appearance features. As a result, the simple IOU tracker has the 2nd best IDF1 and MOTA of $79.00\%$ and $75.90\%$, respectively. While using both motion and appearance information, Deep SORT has the best MOTA and IDF1 score by maintaining a reasonable balance between them. For MDP, its performance is not as good as Deep SORT and IOU tracker. The reason may be its superfluous processing on detection since we directly provide groundtruth detection here. For FAMNet~\cite{chu2019famnet}, its mediocre performance is mainly due to processing on detection noise. Although groundtruth detection are provided here, FAMNet drops too many detection and hence causes many false negatives. 


Furthermore, we include Figure~\ref{fig:ClassComp1.1} to compare the performance under different categories. Generally speaking, the trackers perform much better in ablation study. The difference in performance among categories emphasizes the importance of releasing a GMOT benchmark to evaluate trackers more comprehensively.


\section{Conclusion}

In this paper, we proposed the first, to the best of our knowledge, publicly available densely annotated generic multiple object tracking (GMOT) benchmark named GMOT-40. By thoroughly considering major MOT factors and carefully annotating all tracking objects, GMOT-40 contains 40 sequences evenly distributed among 10 object categories. Associated with the GMOT-40 dataset is the one-shot evaluation protocol for GMOT. Several new baseline algorithms dedicated to one-shot GMOT are developed as well, and evaluated together with relevant MOT trackers to provide references for future study. The evaluation shows that there is still large room to improve for GMOT and further studies are desired. Overall, we expect the benchmark, along with the initial studies, to largely facilitate future research on GMOT, which is an important yet under-explored problem in computer vision.

\paragraph{Acknowledgements.}
We thank the anonymous reviewers for their insightful suggestions that largely help improve the work. This work was supported in part by US National Science Foundation (No.~1814745 and No.~2006665).

{\small
\bibliographystyle{ieee_fullname}
\bibliography{egbib}
}

\end{document}


\title{GMOT-40: A Benchmark for Generic Multiple Object Tracking\\---Supplementary Material---}


\maketitle
\pagestyle{empty}  
\thispagestyle{empty} 

\section{Label Format for GMOT-40}
The label format for proposed GMOT-40 is shown in the Table~\ref{tab: format}. We mainly follow the format of the widely-used MOT15 dataset~\cite{MOTChallenge2015}. The only difference is, MOT15 does not take some challenging targets, like small ones, into consideration for evaluation. It uses an extra flag to indicate that these labeled targets should be ignored. On the contrary, GMOT-40 includes all of them in evaluation, no matter how challenging the targets could be, and the flag is not used here. This is consistent with our motivation, i.e., trackers need to deal with these \textit{real world challenges}.

\begin{table*}[htb]
\resizebox{\textwidth}{!}{%
\begin{tabular}{c|l|l}
\hline\thickhline
Position & Name & Description \\
\hline
1 & Frame number        & Starts from 0, indicates which frame the target belongs to\\
2 & Identity number     & Each trajectory is identified as an unique ID. For detection, it is set to be -1. \\
3 & Bounding box left   & Coordinates of the top-left corner of the bounding box\\
4 & Bounding box top    & Coordinates of the top-left corner of the bounding box\\
5 & Bounding box width  & Width of bounding box in pixels \\
6 & Bounding box height & Height of bounding box in pixels \\
7 & Confidence score    & Predicted probability of the detection being foreground. For groundtruth, it is set to be 1. \\
8-10 & -1      &  Padding to fit MOTChallenge format \\
\hline\thickhline
\end{tabular}%
}
\caption{Annotation format in GMOT-40 dataset.}
\label{tab: format}
\end{table*}

\section{Qualitative Analysis}
\subsection{One-shot GMOT Protocol}
The one-shot GMOT protocol visualization results copied from body part are shown in the top three rows of Figure~\ref{fig:result} (Protocol A). Each bounding box with a polygon line in color represents a tracked target and its trajectory. Compared with the protocol used in ablation study where groundtruth detections are available, IOU tracker performance drops a lot. We think this is because IOU tracker can not handle errors, like false positive/negative detection results, induced by imperfect detectors. By contrast, MDP, Deep SORT and FAMNet have extra mechanisms to refine these faulty detection results during tracking process. Furthermore, unlike Deep SORT and FAMNet, MDP did not use any pre-trained CNN, which makes itself more robust against during generalization to unseen categories. 

\subsection{Protocol of Ablation Study}
\begin{figure*}[!h]
	\small
	\centering
	\begingroup
	\tabcolsep=0.15mm
	\def\arraystretch{0.08}
	\begin{tabular}{p{10pt} ccccc}%
		& MDP \cite{xiang2015learning} 
		& Deep SORT \cite{wojke2017simple} 
		& IOU tracker \cite{bochinski2017high} 
		& FAMnet \cite{chu2019famnet} \\
				\rotatebox[y=30pt]{90}{Protocol A}  & 
		\includegraphics[width=0.23\linewidth]{fig/result/2_MDP.jpg}    & 
		\includegraphics[width=0.23\linewidth]{fig/result/2_DeepSORT.jpg}    & 
		\includegraphics[width=0.23\linewidth]{fig/result/2_IOUTracker.jpg}    &
		\includegraphics[width=0.23\linewidth]{fig/result/2_FAMNet.jpg}    \\
		\\
		\vspace{.1mm}\\
		\rotatebox[y=30pt]{90}{Protocol A}  & 
		\includegraphics[width=0.23\linewidth]{fig/result/32_MDP.jpg}    & 
		\includegraphics[width=0.23\linewidth]{fig/result/32_DeepSORT.jpg}    & 
		\includegraphics[width=0.23\linewidth]{fig/result/32_IOUTracker.jpg}    &
		\includegraphics[width=0.23\linewidth]{fig/result/32_FAMNet.jpg}    \\
		\\
		\vspace{.1mm}\\
		\rotatebox[y=30pt]{90}{Protocol A}  & 
		\includegraphics[width=0.23\linewidth]{fig/result/22_MDP.jpg}    & 
		\includegraphics[width=0.23\linewidth]{fig/result/22_DeepSORT.jpg}    & 
		\includegraphics[width=0.23\linewidth]{fig/result/22_IOUTracker.jpg}    &
		\includegraphics[width=0.23\linewidth]{fig/result/22_FAMNet.jpg} 
		\\
		\\
		\\
		\\
		\\
		\\
		\\
		\\
		\rotatebox[y=30pt]{90}{Protocol B} &
		\includegraphics[width=0.23\linewidth]{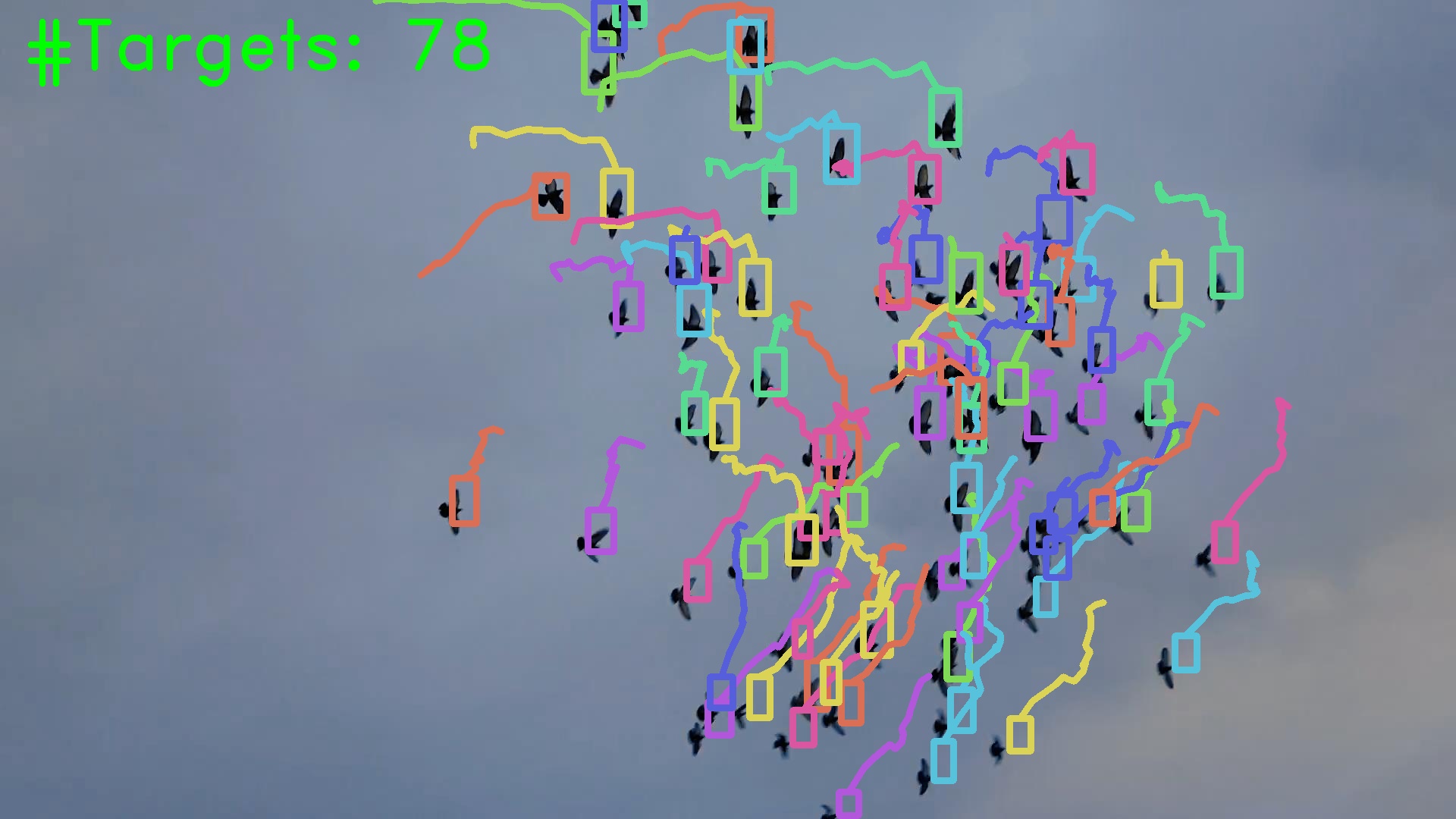}    & 
		\includegraphics[width=0.23\linewidth]{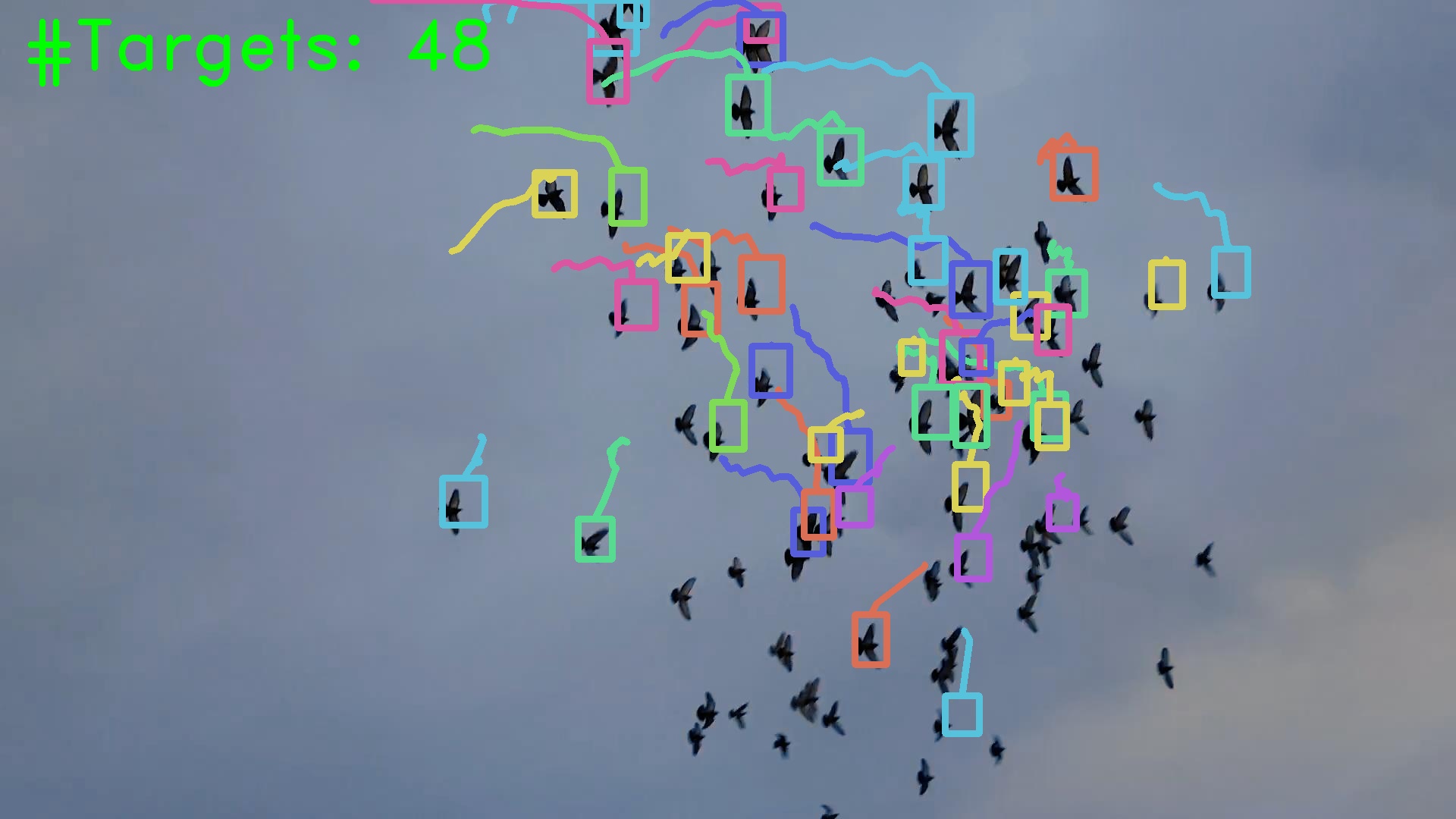}    & 
		\includegraphics[width=0.23\linewidth]{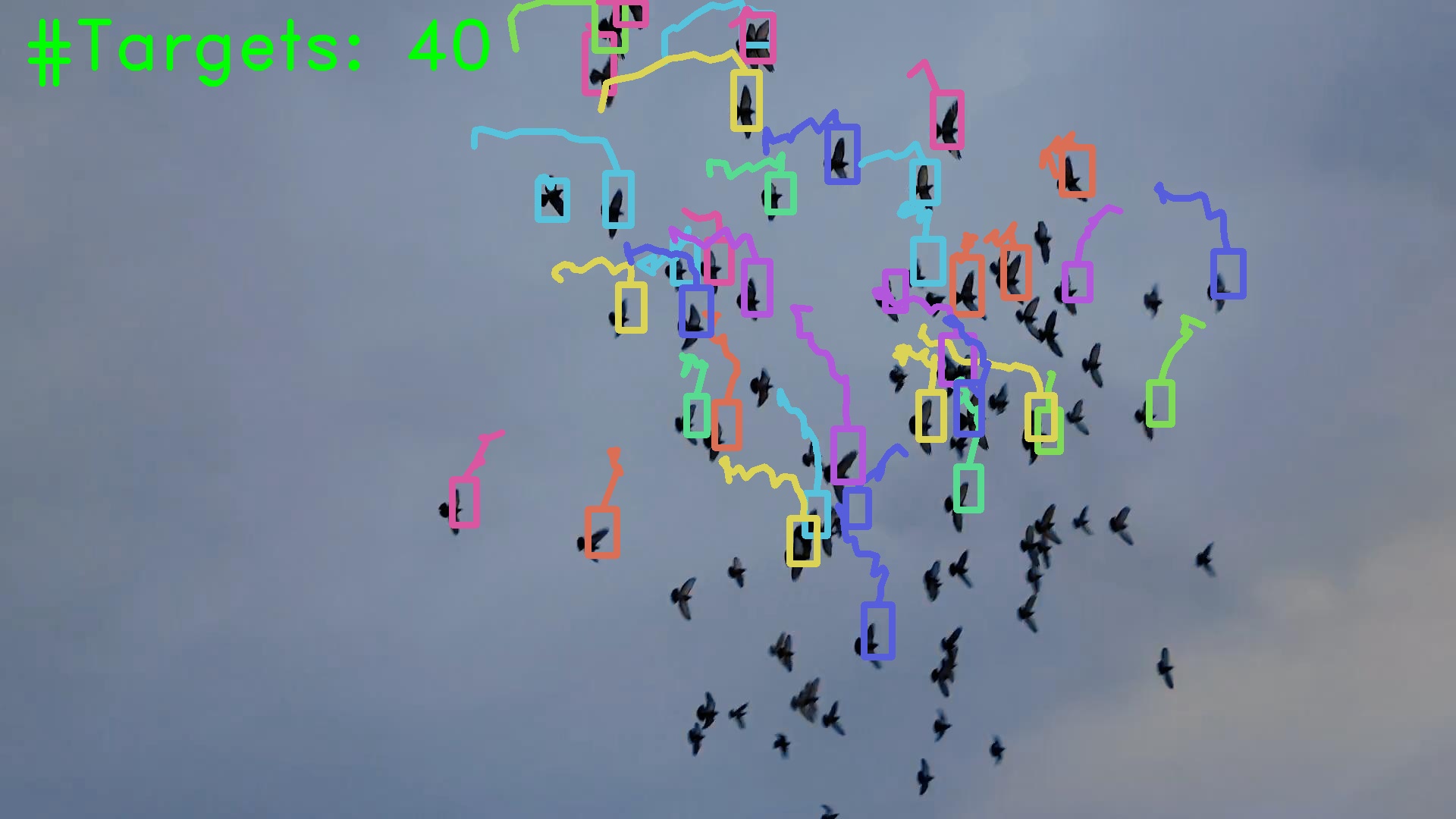}    &
		\includegraphics[width=0.23\linewidth]{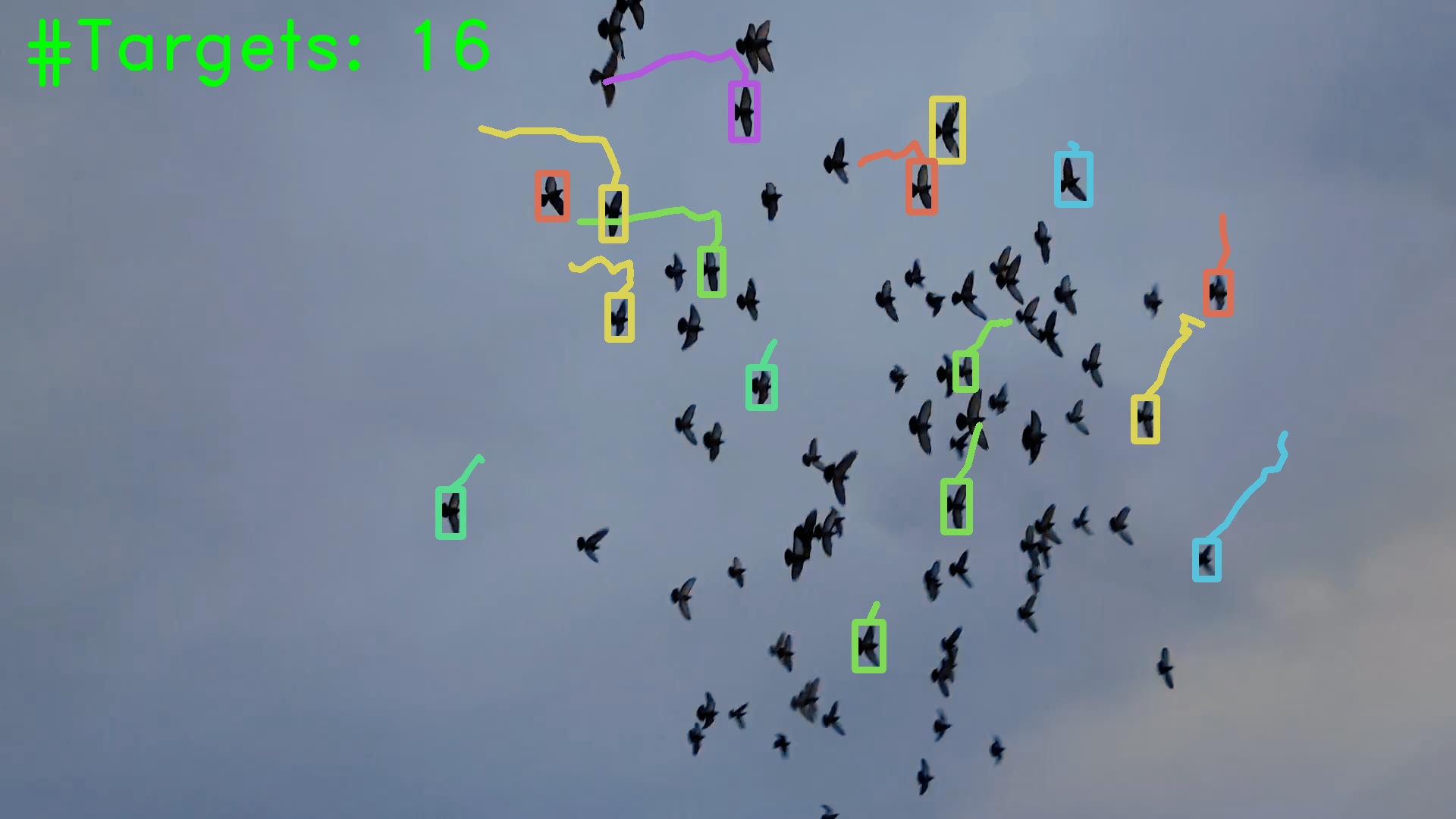}    \\
		\vspace{.1mm}\\
		\\
		\rotatebox[y=30pt]{90}{Protocol B} &
		\includegraphics[width=0.23\linewidth]{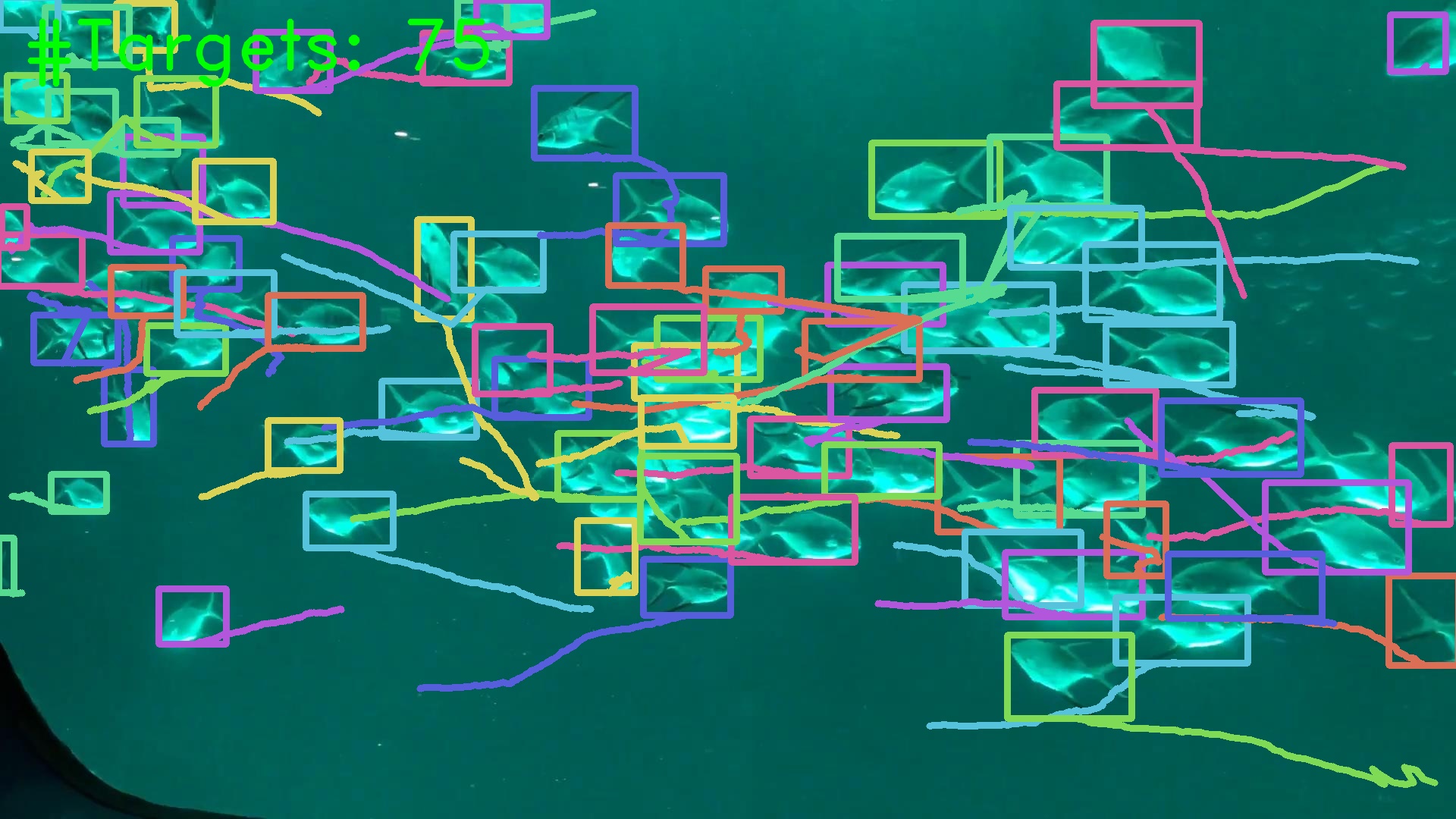}    & 
		\includegraphics[width=0.23\linewidth]{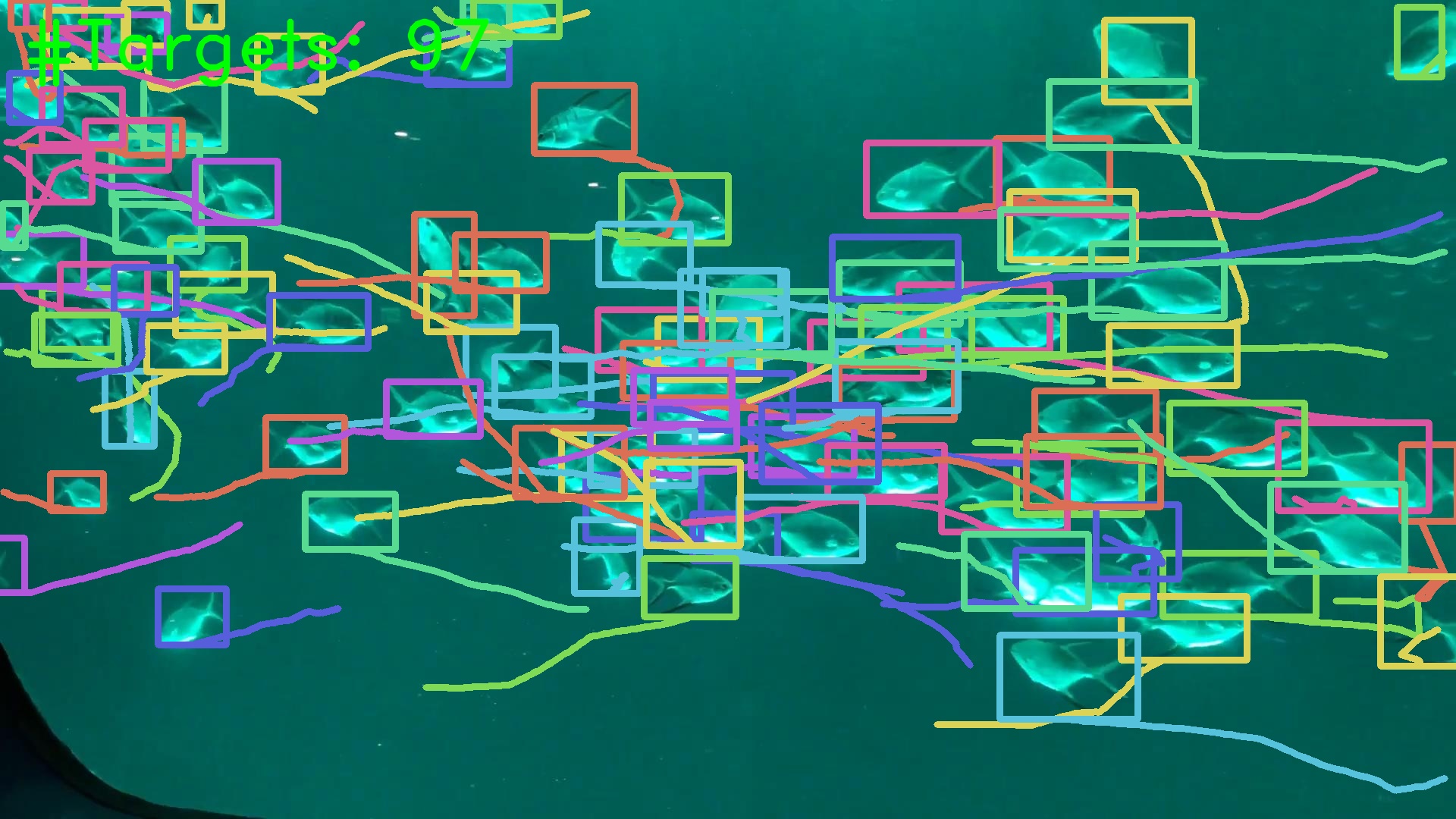}    & 
		\includegraphics[width=0.23\linewidth]{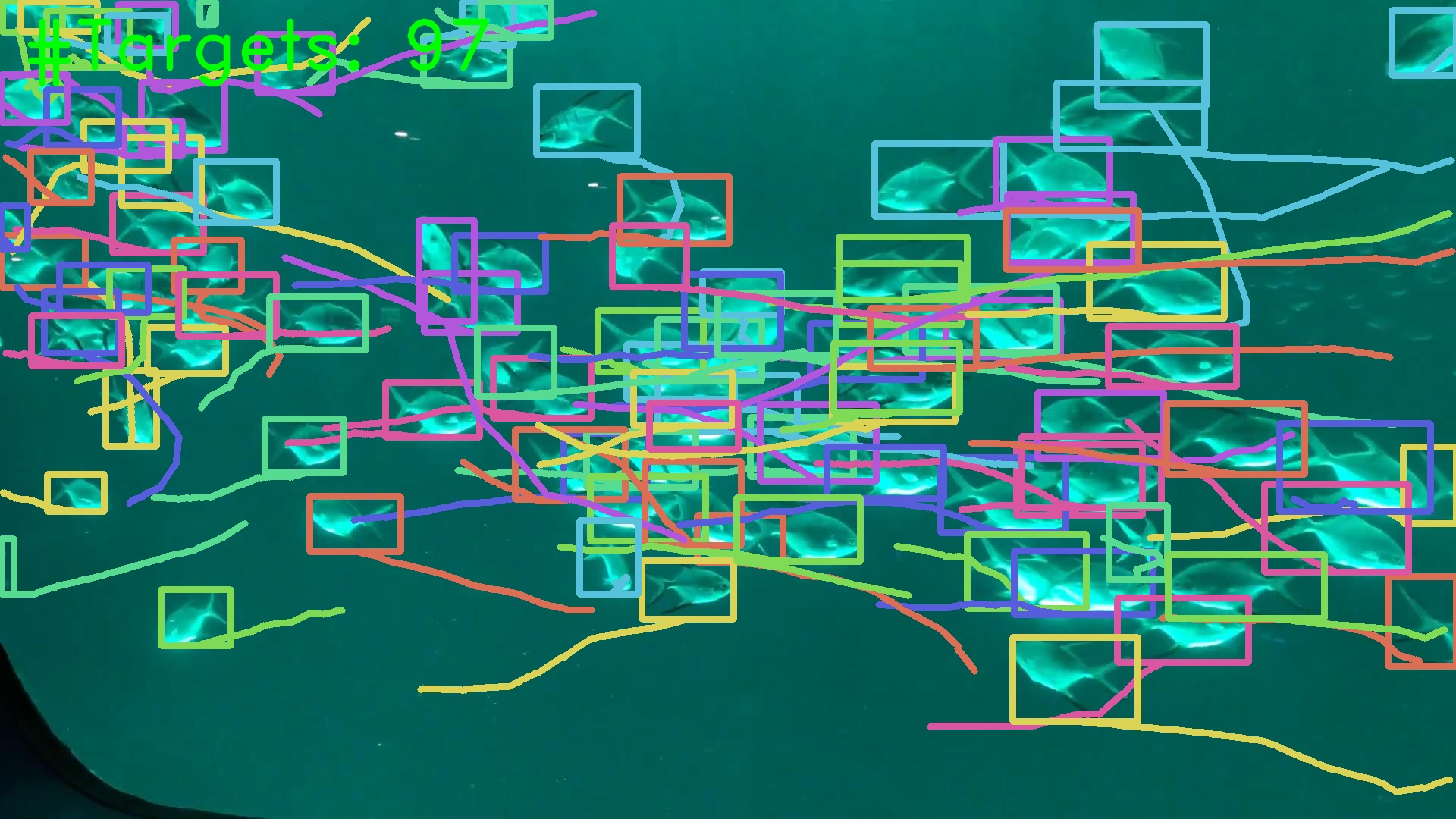}    &
		\includegraphics[width=0.23\linewidth]{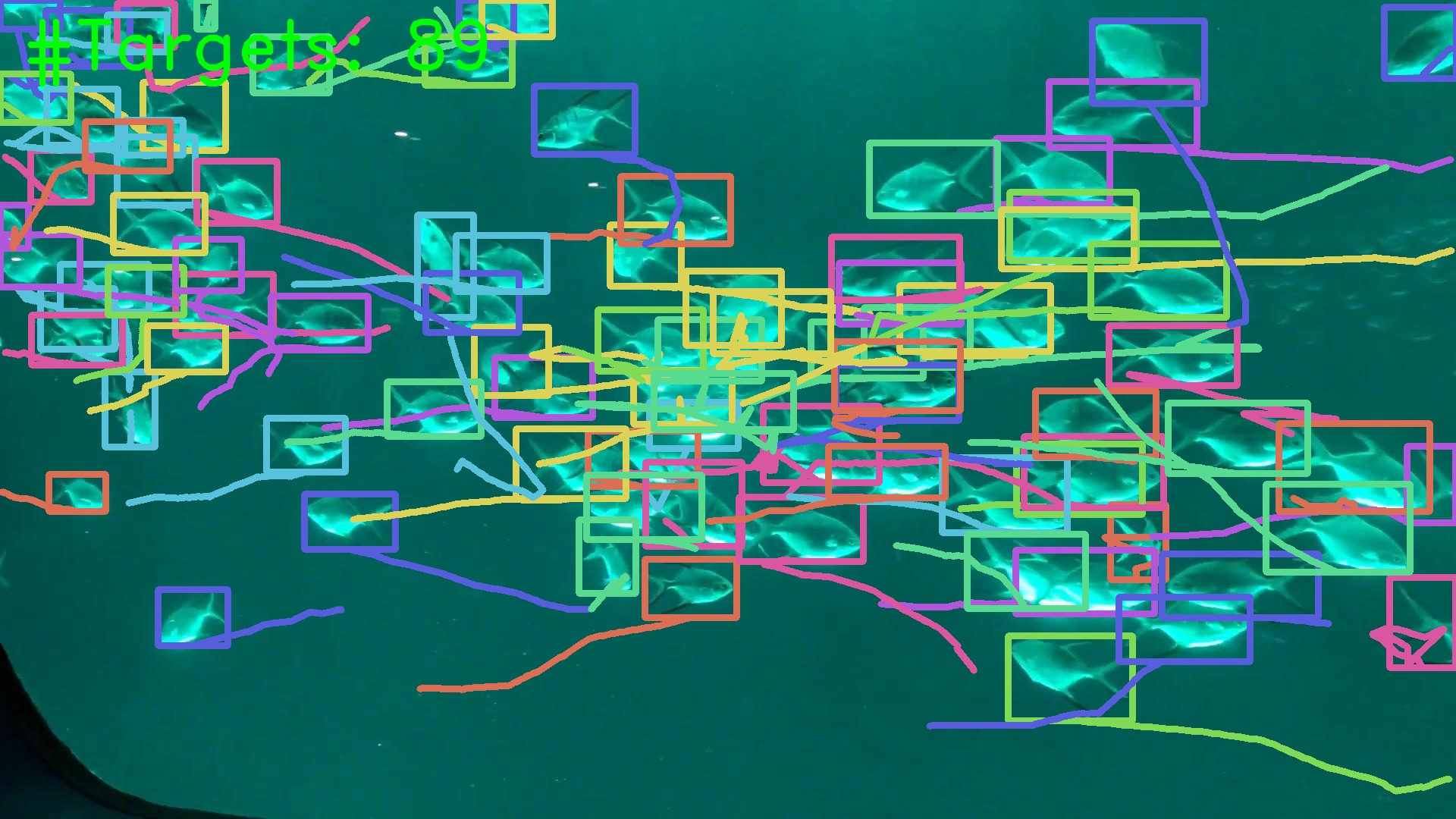}    \\
		\\
		\vspace{.1mm}\\
		\rotatebox[y=30pt]{90}{Protocol B} &
		\includegraphics[width=0.23\linewidth]{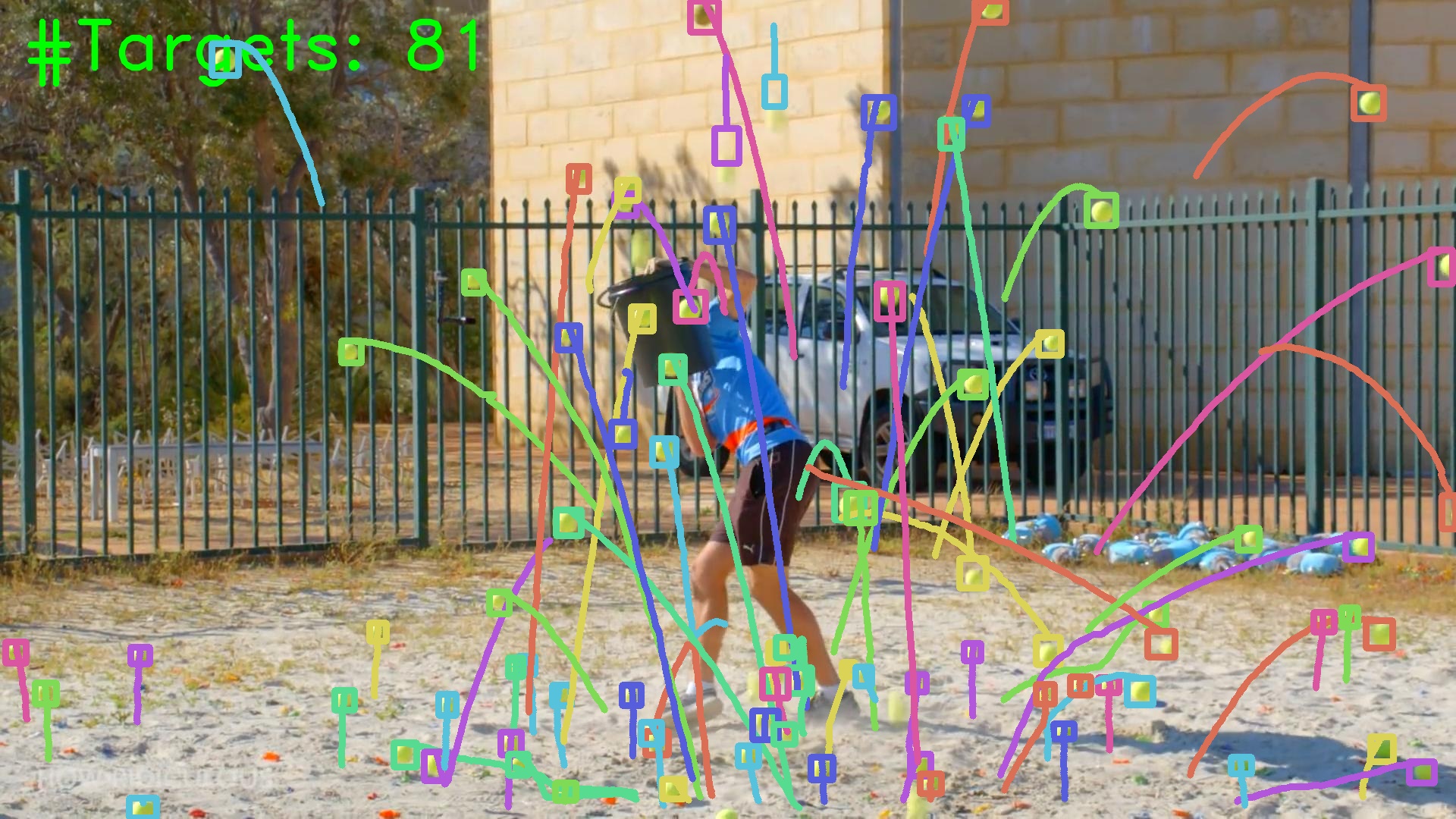}    & 
		\includegraphics[width=0.23\linewidth]{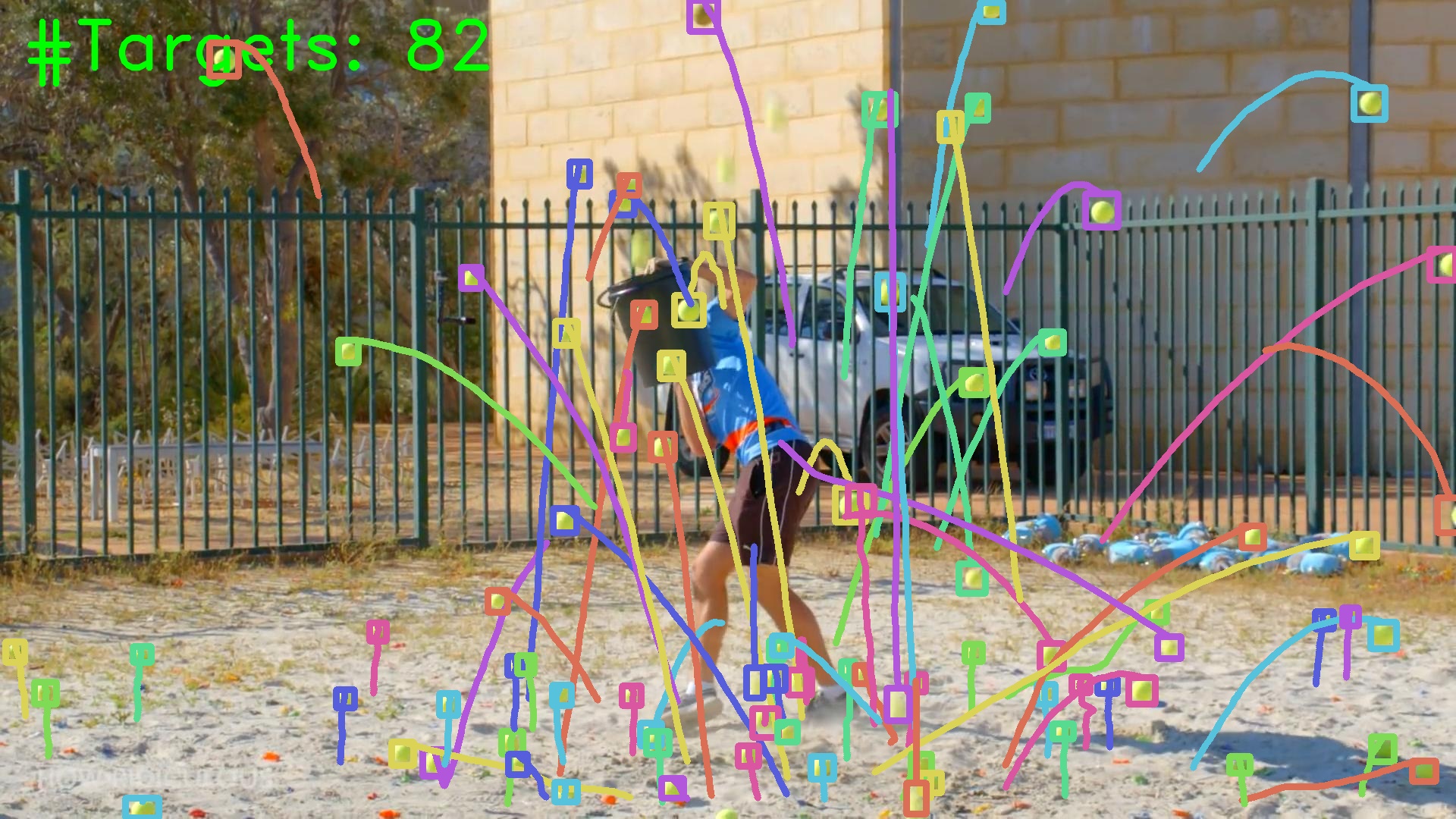}    & 
		\includegraphics[width=0.23\linewidth]{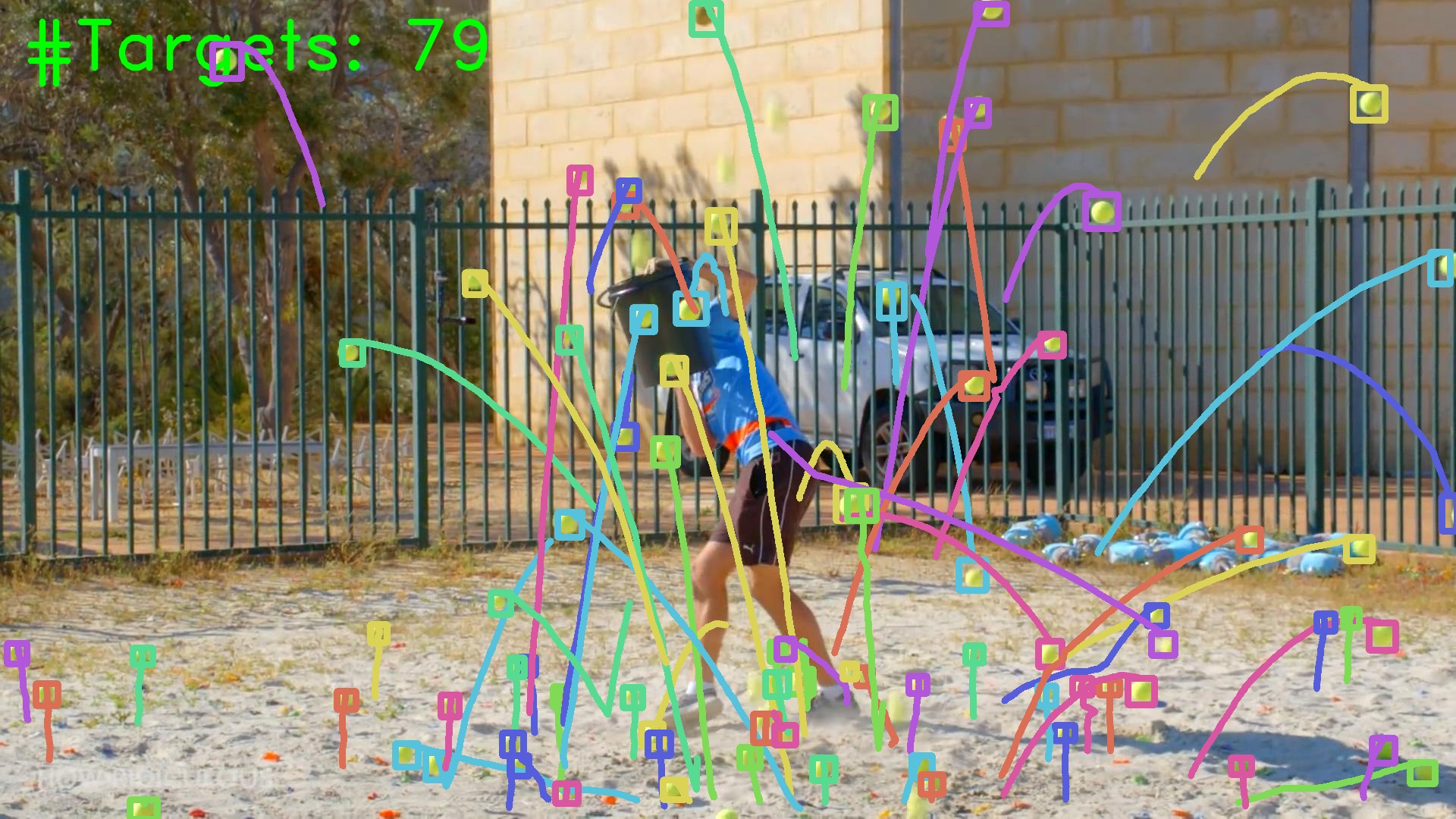}    &
		\includegraphics[width=0.23\linewidth]{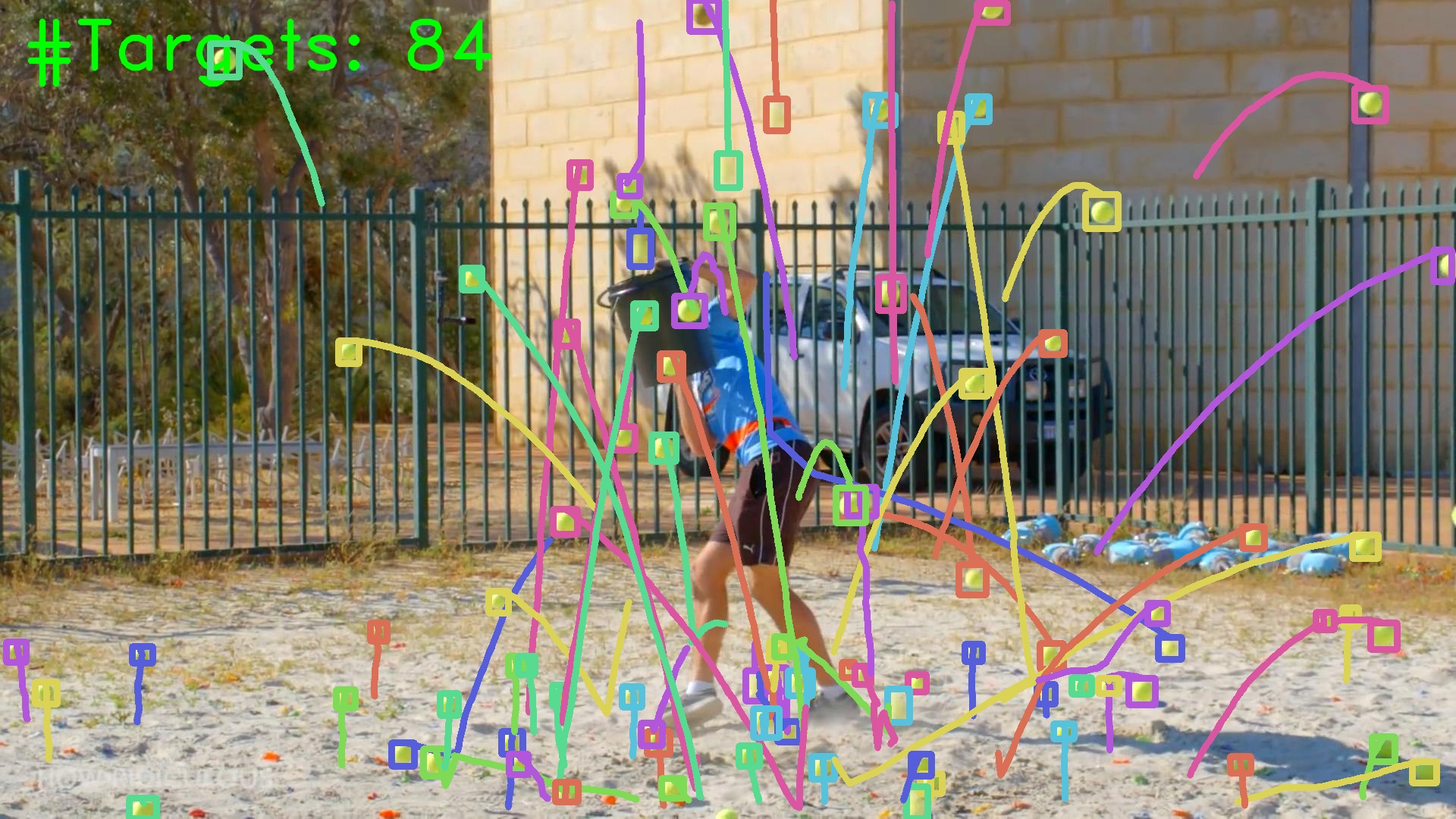}    
   \\
	\end{tabular}
	\endgroup
	\vspace{2mm}
	\caption{Results visualization of four trackers on several sequences using different protocols.}
	\label{fig:result}
\end{figure*}

The protocol of ablation study visualization results are shown in the bottom three rows of Figure~\ref{fig:result} (Protocol B). For the 2nd row of fish sequence in Protocol B result, Deep SORT~\cite{wojke2017simple} and IOU tracker~\cite{bochinski2017high} have tracked more targets than MDP~\cite{xiang2015learning} and FAMNet~\cite{chu2019famnet}. The reason might be Deep SORT and IOU tracker mainly adopt IOU-based tracking paradigm, and they could perform well with all ground truth detection results available. But MDP and FAMNet have superfluous pre-processing on detection which may be harmful under this protocol. Note the visualization result may not be consistent with the quantitative result in the main body for each sequence, due to the averaging process of computing metrics.




\begin{figure*}[!h]
 	\small
	\centering
	\begingroup
	\tabcolsep=0.5mm
    \def\arraystretch{0.10}
	\includegraphics[width=0.9\linewidth]{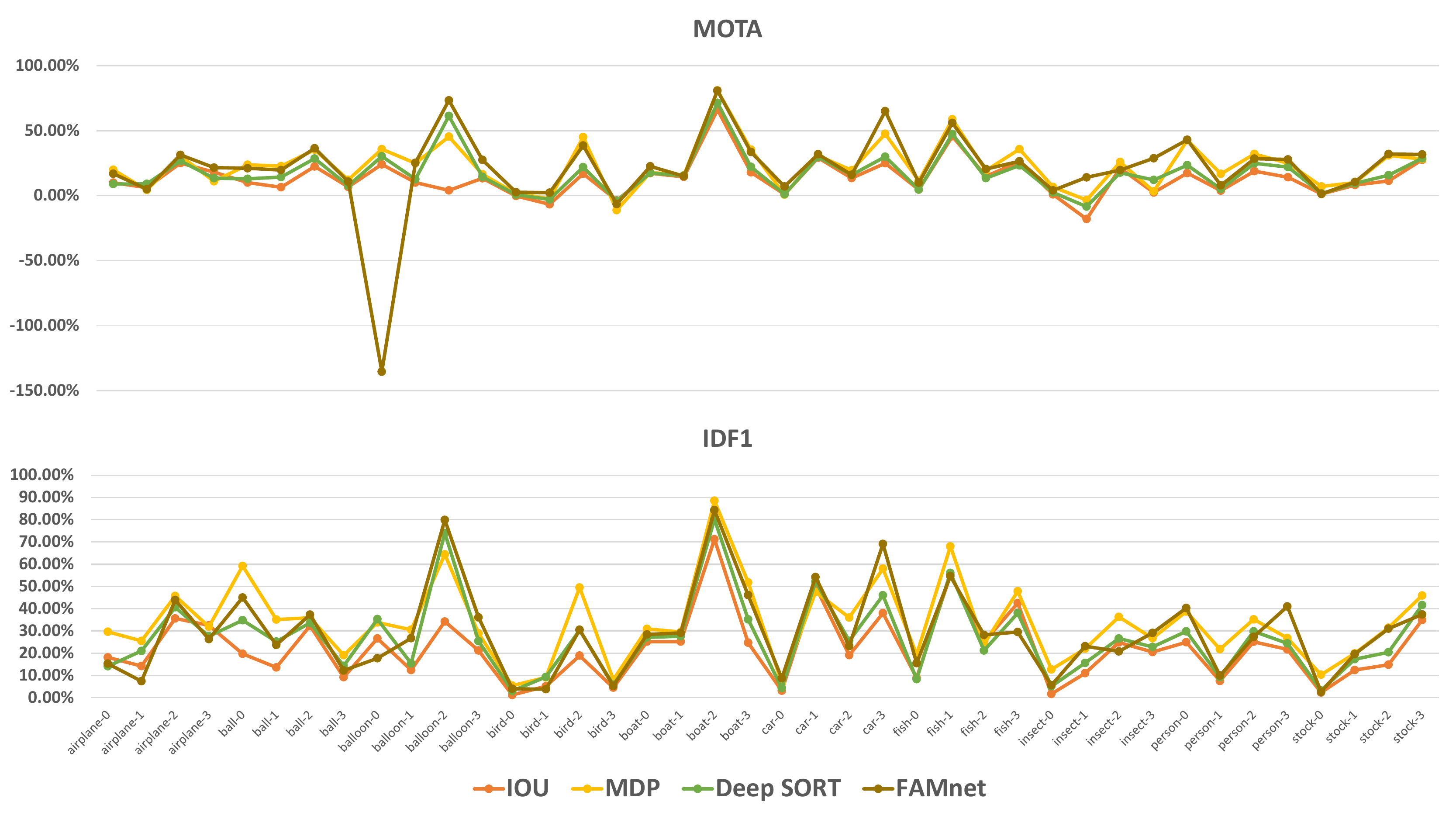}
	\endgroup
	\vspace{2mm}
	\caption{Scores in one-shot GMOT protocol.}
	\label{fig:protocol_2_score}
\end{figure*}

\begin{figure*}[!h]
 	\small
	\centering
	\begingroup
	\tabcolsep=0.5mm
    \def\arraystretch{0.10}
	\includegraphics[width=0.9\linewidth]{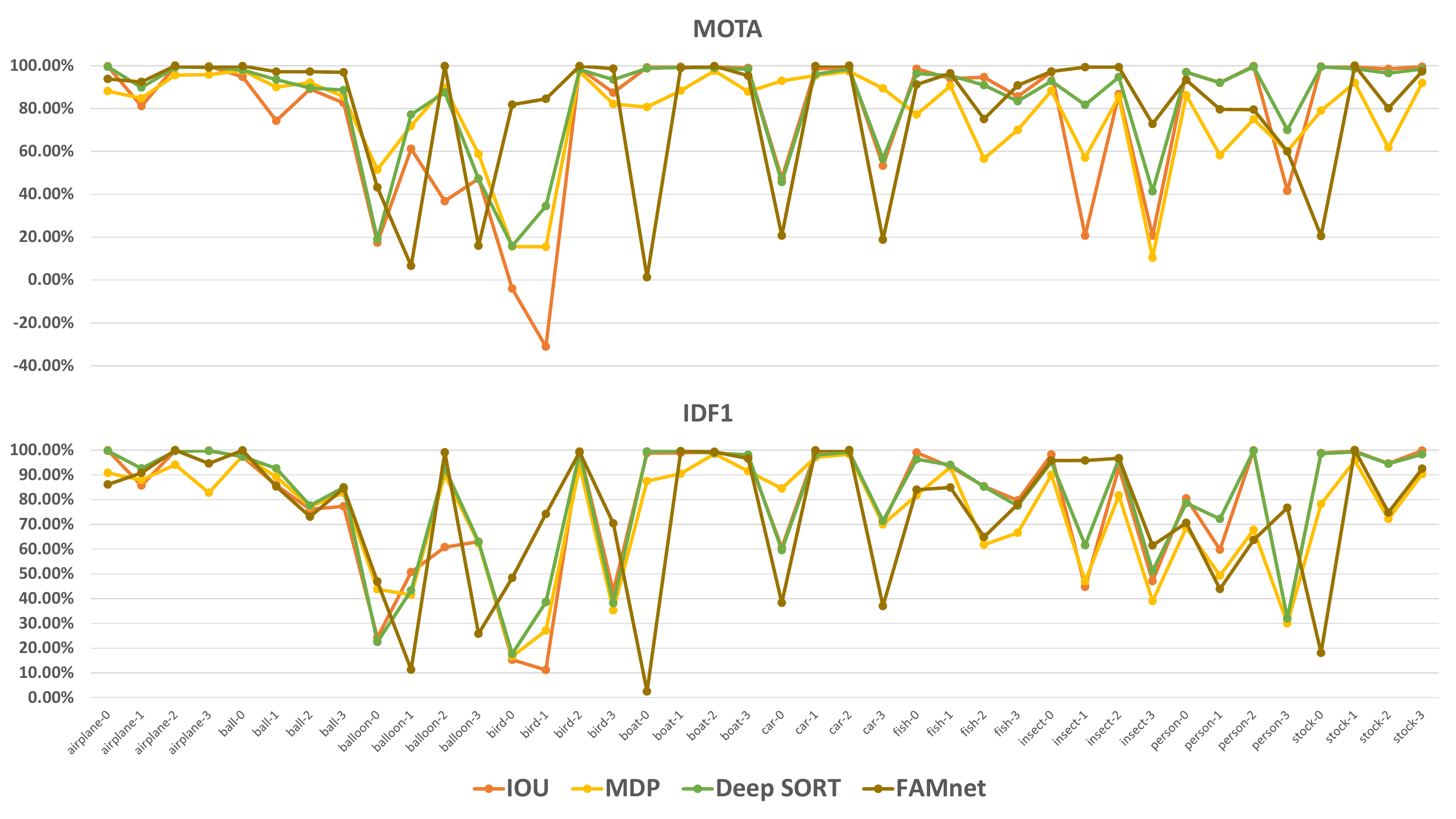}
	\endgroup
	\vspace{2mm}
	\caption{Scores in the protocol of ablation study.}
	\label{fig:protocol_1_score}
\end{figure*}

\section{Scores for All Sequences}

Figure~\ref{fig:protocol_2_score} and Figure~\ref{fig:protocol_1_score} present the scores for both protocols (one-shot GMOT protocol and the protocol used in ablation study) and all sequences. We can see that the sequences that are easy to handle in the protocol of ablation study may be challenging in one-shot GMOT protocol. Yet the challenging sequences in the protocol of ablation study are still difficult in one-shot GMOT protocol. Such difference and similarity again stress the importance and necessity of a one-shot framework in Generic MOT.

{\small
\bibliographystyle{ieee_fullname}
\bibliography{egbib}
}